\documentclass{article}

\usepackage[preprint,nonatbib]{neurips_data_2024}

\usepackage[utf8]{inputenc} 
\usepackage[T1]{fontenc}    
\usepackage{hyperref}       
\usepackage{url}            
\usepackage{booktabs}       
\usepackage{amsfonts}       
\usepackage{nicefrac}       
\usepackage{microtype}      
\usepackage{xcolor}         
\usepackage{array}
\usepackage{amsmath} 
\usepackage{cleveref}
\usepackage{tabularx}  
\usepackage{graphics}
\usepackage{multirow}
\usepackage{bm}
\usepackage{tablefootnote}
\usepackage{manyfoot}
\usepackage{multicol}
\usepackage[para]{footmisc}
\usepackage{hyperref}
\usepackage[backend=bibtex,style=numeric]{biblatex} 
\title{HEST-1k: A Dataset for Spatial Transcriptomics and Histology Image Analysis}

\author{
Guillaume Jaume$^{1,2,*}$ $\quad$
Paul Doucet$^{1,3,*}$ $\quad$
Andrew H. Song$^{1,2}$
$\quad$ Ming Y. Lu$^{1,2,4}$ \\
\textbf{
Cristina Almagro-Pérez$^{1,2}$ $\quad$
Sophia J. Wagner$^{1,6,7}$ $\quad$
Anurag J. Vaidya$^{1,2,5}$
}\\
\textbf{
Richard J. Chen$^{1,2}$ $\quad$
Drew F.K. Williamson$^{8}$ $\quad$
Ahrong Kim$^{1,9}$ $\quad$
Faisal Mahmood$^{1,2}$
}\\
  $^1$Mass General Brigham, Boston, USA $\quad$
  $^2$Harvard Medical School, Boston, USA\\
  $^3$ETH Zurich, Switzerland $\ \ $
  $^4$EECS MIT, Cambridge, USA $\ \ $\\
  $^5$HST MIT, Cambridge, USA $\ \ $
  $^6$TUM, Munich, Germany $\ \ $\\
  $^7$Helmholtz Munich, Munich, Germany
  $^8$Emory School of Medicine, Atlanta, USA $\quad$\\
  $^9$Pusan National University, South Korea\\
  \texttt{gjaume@bwh.harvard.edu} $\quad$ \texttt{faisalmahmood@bwh.harvard.edu}
}

\begin{document}

\maketitle
\def\thefootnote{*}\footnotetext{Equal contribution}\def\thefootnote{\arabic{footnote}}

\begin{abstract}
Spatial transcriptomics enables interrogating the molecular composition of tissue with ever-increasing resolution and sensitivity. However, costs, rapidly evolving technology, and lack of standards have constrained computational methods in ST to narrow tasks and small cohorts. In addition, the underlying tissue morphology, as reflected by H\&E-stained whole slide images (WSIs), encodes rich information often overlooked in ST studies. Here, we introduce HEST-1k, a collection of 1,229 spatial transcriptomic profiles, each linked to a WSI and extensive metadata. HEST-1k was assembled from 153 public and internal cohorts encompassing 26 organs, two species (\textit{Homo Sapiens} and \textit{Mus Musculus}), and 367 cancer samples from 25 cancer types. HEST-1k processing enabled the identification of 2.1 million expression--morphology pairs and over 76 million nuclei. To support its development, we additionally introduce the HEST-Library, a Python package designed to perform a range of actions with HEST samples. We test HEST-1k and Library on three use cases: (1) benchmarking foundation models for pathology (HEST-Benchmark), (2) biomarker exploration, and (3) multimodal representation learning. HEST-1k, HEST-Library, and HEST-Benchmark can be freely accessed at \href{https://github.com/mahmoodlab/hest}{\texttt{https://github.com/mahmoodlab/hest}}.
\end{abstract}

\addtocontents{toc}{\protect\setcounter{tocdepth}{0}}

\section{Introduction}

Advances in molecular profiling enable spatially-resolved gene expression analysis with increasingly large gene panels, enhanced spatial resolution, and greater sensitivity~\cite{asp2020spatially,rao2021exploring}. From the early days of bulk RNA sequencing constrained by its coarse resolution and limited gene panels, spatially-resolved technologies have progressed to achieve whole-transcriptome sequencing at sub-cellular resolution~\cite{li2021from}. In cancer research, spatial transcriptomics (ST) holds particular promise for characterizing the tumor microenvironment, a key element in understanding disease progression and treatment response~\cite{staahl2016visualization, marx2021method, devisser2023evolving,xiao2021tumor}. With the large amount of transcriptomics data generated by a single ST sample (e.g., $>$10 million transcripts are detected in a typical 10x Genomics Xenium assay), computational methods are often used to uncover promising biomarkers, such as employing clustering methods for cell phenotyping~\cite{rao2021exploring}.

However, high costs and rapidly evolving technology have constrained computational methods to narrow tasks and data cohorts of only a few patients~\cite{meylan2022tertiary,janesick2023high,valdeolivas2023charting}. Consequently, we observe a lack of standardized resources and unified formats for handling ST, which limits the development of deep learning models on a large scale~\cite{moses2022museum}. In addition, the underlying tissue morphology, traditionally visualized in hematoxylin and eosin (H\&E)-stained tissue sections (whole-slide images, WSIs), is often overlooked in ST studies, despite encoding valuable information. In particular, pairs of ST and WSI enable analyzing expression changes in their morphological context, which may facilitate the identification of morphological biomarkers (e.g., changes in nuclear shape) that correspond to gene regulation patterns. Alternatively, pairs of ST and WSI can enable multimodal tissue representation learning for joint modeling of the morphomolecular signature of tissue at a scale and resolution beyond bulk RNA sequencing~\cite{chen2022pan}. Finally, the development of ``foundation models'' for encoding histopathology images~\cite{wang2021transpath,filiot2023scaling,chen2024towards,huang2023visual,lu2024towards} has increased the need for new, diverse, and challenging benchmarks beyond diagnostic tasks. Using ST, new tasks can be defined to predict gene expression changes from histology.

\begin{figure}[t]
   \centering
   \resizebox{\textwidth}{!}{
       \includegraphics{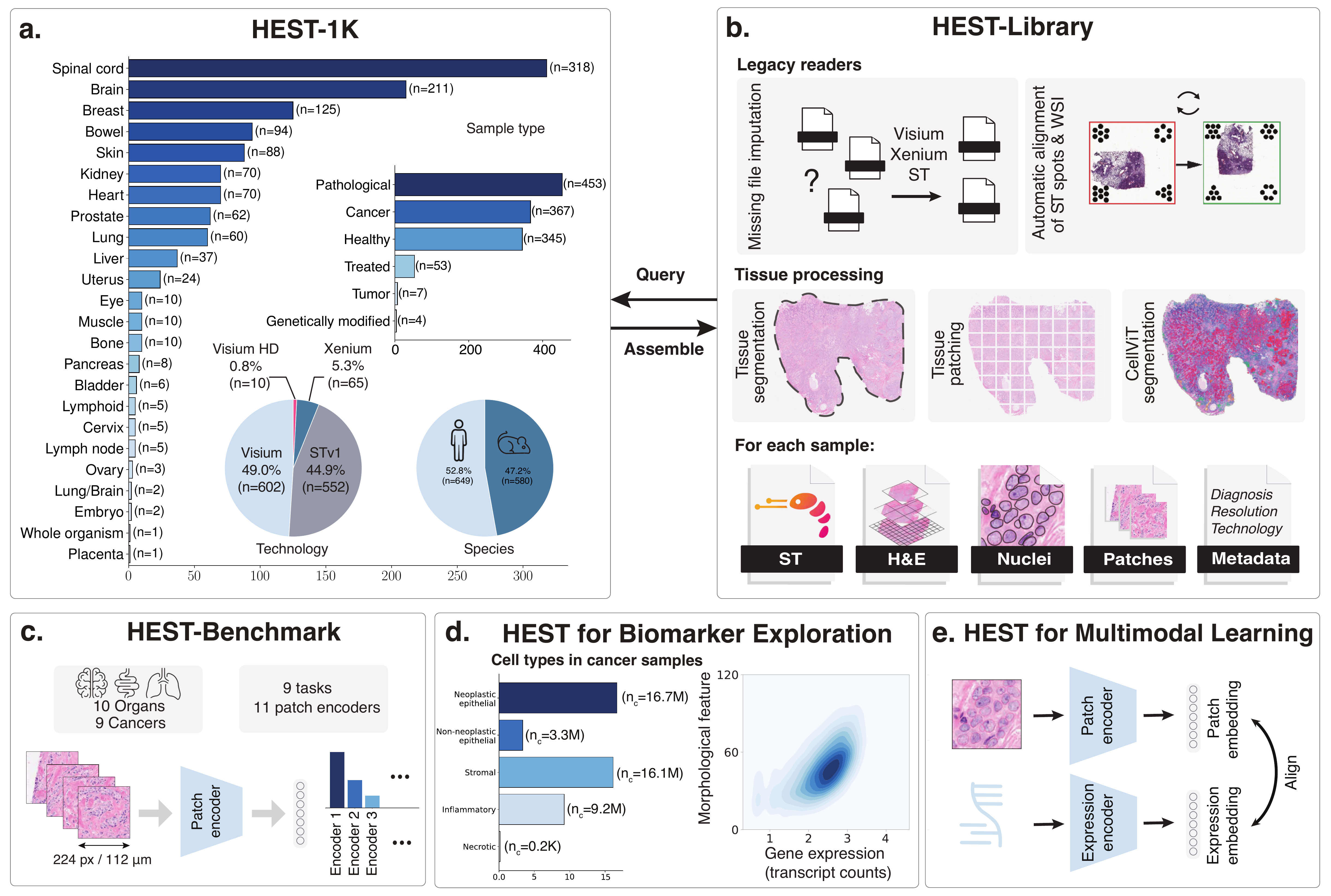}
   }
   \caption{
       \textbf{The HEST environment.}
       \textbf{a.} Overview of HEST-1k, a dataset of $n$=1,229 paired spatial transcriptomics, H\&E-stained whole-slide images and metadata. ``Pathological'' cases refer to non-tumor/non-cancer samples; ``Tumor'' refers to non-cancer samples.
       \textbf{b.} Overview of HEST-Library functionalities.
       \textbf{c., d., e.} Applications of HEST-1k include benchmarking foundation models for histology (\textbf{c.}), biomarker exploration (\textbf{d.}) and multimodal representation learning (\textbf{e.}).
   }
   \label{fig:overview}
\end{figure}

Here, we introduce HEST-1k, a collection of paired ST and H\&E-stained WSIs curated from public and internal cohorts ({\autoref{fig:overview}.a}). HEST-1k comprises 1,229 samples from 153 cohorts encompassing 26 organs, two species (\textit{Homo Sapiens} and \textit{Mus Musculus}), and 367 cancer samples from 25 different subtypes. Processing all samples in HEST-1k resulted in 2.1 million expression--morphology pairs and 76 million detected nuclei. With new cohorts frequently made public, we also introduce the HEST-Library, a Python package for interacting with HEST-1k data and assembling new samples as they become available ({\autoref{fig:overview}.b}). We highlight the potential of HEST-1k through three use cases: (i) benchmarking foundation models for histology using the HEST-Benchmark, a set of nine tasks (eight human cancer types and nine organs) for gene expression prediction from histology and evaluated on eleven state-of-the-art models ({\autoref{fig:overview}.c}), (ii) a proof-of-concept demonstrating the use of HEST-1k for biomarker characterization ({\autoref{fig:overview}.d}), and (iii) a proof-of-concept for expression-guided fine-tuning of foundation models for histology ({\autoref{fig:overview}.e}).

\section{Related work}

\noindent\textbf{Libraries for ST analysis.} Libraries to process, visualize, and analyze ST have been built around two core pipelines: \textit{Scanpy}~\cite{wolf2018scanpy} (and the {\small\textsc{Anndata}} format) in Python and \textit{Seurat}~\cite{butler2018integrating} in R. \textit{Scanpy} has served as the foundation for several subsequent developments such as \textit{Squidpy}~\cite{palla2022squidpy} for spatial data exploration at cellular-, gene-, and morphological-level, \textit{SpatialData}~\cite{marconato2024spatialdata} for multi-technology integration and deep learning interfacing, \textit{STlearn}~\cite{pham2023robust} for cell-cell interactions and spatiotemporal trajectory analyses, and SOPA~\cite{blampey2024sopa} for designing multistep pipelines. In R, \textit{Seurat}~\cite{butler2018integrating} has been consolidated with packages such as \textit{BayeSpace}~\cite{zhao2021spatial} for clustering and spot super-resolution, and \textit{Giotto suites}~\cite{chen2023giotto} for preprocessing, data integration and visualization of multiple ST technologies. 10x Genomics also includes proprietary software analytics through the Xenium and Visium Explorer pipelines for multimodal visualization, nuclear segmentation, and cell deconvolution. However, none of these pipelines were designed to handle the diversity of legacy data, where datasets can suffer from missing or incorrect data, such as alignment mismatches, incorrect pixel resolution, inconsistent image file formats, etc. 

\noindent\textbf{Molecular profile prediction from H\&E.} Molecular profiling from histology images has been explored both at (1) \emph{slide-level} to predict bulk molecular status/changes from a WSI and at (2) \emph{patch-level} to predict local molecular status/changes from regions-of-interest. (1) Slide-level profiling has been explored to predict the gene mutations~\cite{saldanha2023self,wagner2023fully,kather2020pan,echle2021deep,fu2020pan,wang2019predicting,loeffler2022artificial}, microsatellite instability \cite{wagner2023fully,kather2019deep}, and gene expression changes \cite{schmauch2020deep,el2024regression,he2020integrating,alsaafin2023learning}, among others. The motivation is two-fold: First, patient screening to substitute or complement costly clinical molecular assays, and second, to identify morphological correlates of molecular alterations for discovering novel biomarkers. Such studies can be conducted on large patient cohorts as they mainly rely on data generated by the routine clinical workflow (e.g., using TCGA cohorts with $>$11,000 cases from 33 cancer types). (2) With ST, several works have explored predicting expression changes from regions-of-interest~\cite{he2020integrating,xie2023spatially,pang2021leveraging,zhao2024hist2cell,rahaman2023breast,monjo2022efficient,mondol2023hist2rna,chung2024accurate,wang2024m2ort}. Due to limited cohort sizes (typically one to ten patients), transfer learning has become the norm using pretrained models based on ConvNets~\cite{he2020integrating} or Vision Transformers~\cite{xie2023spatially,pang2021leveraging}. Due to the inherent noise found in transcriptomic measurements, several methods have been developed for integrating context that can account for global and local information from surrounding ST spots~\cite{hu2021spagcn, chung2024accurate,zhao2024hist2cell,wang2024m2ort}. While recent technologies offer near-single-cell resolution (such as Visium HD and Xenium), legacy assays operate at a more coarse resolution, which can be upsampled using super-resolution techniques~\cite{zhao2021spatial,bergenstrahle2022super,zhang2024inferring}. The potential clinical and research implications of such methods are still being explored, with HEST-1k potentially catalyzing their large-scale development. 

\noindent\textbf{Foundation models in pathology.} A fundamental task in computational pathology is to extract \textit{general-purpose} embeddings of image patches (typically 256$\times$256 to 512$\times$512-pixel regions) that can then be used for downstream tasks, such as diagnosis or prognosis prediction. To achieve this, self-supervised learning (SSL) has been extensively applied~\cite{koohbanani2021self,filiot2023scaling,vorontsov2024foundation,wang2021transpath,kang2023benchmarking,ciga2022self,wang2022transformer,azizi2023robust,chen2024towards,lu2024towards,jaume2024transcriptomics,jaume2024multistain}, such as based on the DINOv2 framework~\cite{oquab2023dinov2}. General-purpose patch encoders are trained on increasingly large and diverse patient cohorts (e.g., UNI~\cite{chen2024towards} uses a ViT-Large trained on 100k WSIs, Virchow~\cite{vorontsov2024foundation} uses a ViT-Huge trained on more than 1.5M WSIs). Recently, vision-language encoders designed for pathology have also been proposed~\cite{gamper2021multiple,lu2023visual,huang2023visual,lu2024towards,lu2024multimodal} and rely on large-scale paired data scraped from social media, textbooks, or publications. As the number of such models rapidly increases, new, diverse, and challenging benchmarks are needed to replace or complement well-established tasks where performance has saturated. HEST-Benchmark aims to address this by offering a set of nine patch-level tasks for gene expression prediction from histology. 

\noindent\textbf{Patch-level benchmarks in histopathology.} Early task and dataset contributions in computational pathology revolved around classifying small regions of interest. Over the years, a variety of benchmarks have been established: In prostate cancer, Gleason grading at pixel- and patch-level has been widely explored, with public resources such as AGGC~\cite{huo2022comprehensive}, DiagSet~\cite{koziarski2024diagset}, and SICAPv2~\cite{silvarodriguez2020going}. In colorectal cancer, datasets have been proposed for tissue classification, such as HunCRC~\cite{pataki2022huncrc}, UniToPatho~\cite{barbano2021unitopatho}, MHIST~\cite{wei2021petri}, and CRC-100k~\cite{kather2019predicting}. In breast cancer, morphological subtyping has been vastly explored (e.g., for atypical ductal hyperplasia detection), such as BACH~\cite{aresta2019bach}, BRACS~\cite{brancati2021bracs}, and BreakHis~\cite{spanhol2015dataset}, and for lymph node metastasis detection with Patch CAMELYON (pCAM)~\cite{veeling2018rotation}, respectively. However, the performance on many of these datasets has saturated; for instance, Gleason scoring reaches similar or better performance than pathologists~\cite{bulten2022artificial}, which limits objective comparisons of new methods and hinders well-informed model selection for developing better features. Instead, HEST-Benchmark provides a collection of diverse and challenging tasks that enable assessing the predictive capabilities of foundation models for histology.

\section{HEST-1k Dataset}

We present HEST-1k, a dataset of paired ST, H\&E-stained WSIs, and metadata (\autoref{fig:overview}.a). To this end, we extracted all publicly available cohorts that provide ST with H\&E-stained whole-slide images. Specifically, we harvested data from 10x Genomics public datasets (TENX)\footnote{https://www.10xgenomics.com/datasets}, Mendeley (MEND)\footnote{https://data.mendeley.com/}, Spatial-Research (SPA)\footnote{https://www.spatialresearch.org/}, Zenodo (ZEN)\footnote{https://zenodo.org/}, the National Center for Biotechnology Information (NCBI)\footnote{https://www.ncbi.nlm.nih.gov/gds/}, GitHub\footnote{https://github.com/}, the Human Cell Atlas\footnote{https://data.humancellatlas.org/}, BioStudies\footnote{https://www.ebi.ac.uk/biostudies/}, HTAN\footnote{https://humantumoratlas.org/}, and internal data cohorts. A summary of all sources is provided in Appendix~\autoref{tab:hest1k} with specifics in Appendix~\autoref{tab:tenx}, \ref{tab:ncbi}, \ref{tab:spa},\ref{tab:mend},\ref{tab:misc},\ref{tab:internal}, and \ref{tab:zenodo}.

\subsection{Metadata}

As spatial transcriptomics experiments were not intended for large-scale computational research, they are provided in various formats (e.g., images can be in JPG or TIFF format, with or without cross-modal alignment files) and resolutions. We unified all data with comprehensive metadata with \textit{generic}-, \textit{histology}-, and \textit{expression}-related descriptors for all samples. \textbf{Generic:} We provide the reference to the original publication, download link, year of publication, license, and sample species. Each sample is then categorized as either healthy, cancer, tumor (non-cancer), treated (which refers to a post-compound administration), genetically modified (mostly knock-out mouse samples), or pathological (i.e., non-tumorous with extra specification). All cancer samples were unified using the OncoTree code, a taxonomy of cancer types provided by the Memorial Sloan Kettering Cancer Center\footnote{https://oncotree.mskcc.org}. Finally, we provide the organ using the highest level of the OncoTree taxonomy as a reference. \textbf{Expression:} We report the number of genes and spots per sample, the spot resolution and spacing, the total number of reads, and the mean number of reads per spot. We additionally provide the transcriptomic technology (ST, Visum, Visium HD, or Xenium). \textbf{Histology:} We provide the image resolution (in $\mu$m/pixel) and magnification as 10$\times$ (1.15 to 0.8 $\mu$m/px), 20$\times$ (0.8 to 0.4 $\mu$m/px) and 40$\times$ (0.4 to 0.1 $\mu$m/px). All images with a pixel size higher than 1.15 $\mu$m/px were discarded to ensure an acceptable image quality. In addition, we provide the image size at the highest resolution and the tissue preparation protocol (frozen or formalin-fixed paraffin-embedded, FFPE). 

\subsection{Histology}

All tissue sections were normalized and transformed into a generic TIFF object, a pyramidal image that can easily be integrated into computational frameworks using \textsc{Openslide} or viewers such as \textsc{QuPath}~\cite{bankhead2017qupath}. In addition, we provide a contour object that delineates all the tissue regions identified in the image. We developed a robust tissue \emph{vs.} background detection method where we fine-tuned a DeepLabV3~\cite{chen2017rethinking} model with an ImageNet-pretrained ResNet50 backbone on a set of annotated segmentation regions (including pen marks, fiducials, multiple stains, artifacts, etc.). From the tissue segmentation, we extracted 224$\times$224-pixel patches at 20$\times$ magnification around each spot. For Xenium samples, we generated ``pseudo-Visium'' spots by pooling transcripts on 55 $\times$ 55-$\mu$m patches without spacing. This yielded 2.1 million valid patches for which a corresponding expression profile was derived. Such patching can readily be used for various downstream tasks, such as employed in the \textbf{HEST-Benchmark} or multimodal fine-tuning of foundation models for histology (\Cref{sec:benchmark} and~\ref{sec:dl}). 

\subsection{Nuclear segmentation and classification}

In addition to patching, we include nuclear segmentation that delineates each nucleus identified in all slides from HEST-1k. We used CellViT~\cite{horst2024cellvit}, a state-of-the-art nuclear segmentation model that was trained on the PanNuke dataset~\cite{gamper2019pannuke,gamper2020pannuke}. CellViT enables joint instance segmentation and classification of each nucleus into five classes: neoplastic epithelial, non-neoplastic epithelial, inflammatory, stromal, and necrotic. On average, we identified 62.1k nuclei per slide, for a total of 76.4 million nuclei identified across all samples. Among those, 17.6 million are classified as neoplastic, 21.5 million as stromal, 4.9 million as normal epithelial, 15.4 million as inflammatory, and 76 thousand as necrotic. The resulting nuclear segmentation and classification can easily be visualized using \textsc{QuPath} (using geojson) or loaded as Python/R objects (using JSON). For all Xenium samples, we additionally provide the nuclear and cell segmentation derived from the DAPI staining finely aligned with the H\&E slide. 

\subsection{Gene expression}

All expression data were unified in a {\small \textsc{Anndata}} object that can be loaded with \textit{scanpy}. {\small \textsc{Anndata}} encodes the gene names (as \textit{var}) and number of spots (as \textit{obs}). Each entry represents the raw transcript counts of a gene in a given spot. No additional normalization was conducted, and we let users explore various normalization strategies based on needs, e.g., using total count normalization, log-normalization, etc. In addition, we include metadata to specify the number of genes, the gene panel, and the tissue site. For all Xenium samples, we also provide the list of all measured transcripts with their exact 2D position in the tissue (aligned with the H\&E slide).

To use the expression in tandem with the WSI, an alignment file describing the mapping between the image and the spots is needed. However, relying on publicly available alignment information brings three challenges: (1) most datasets report alignment with respect to a low-resolution version of the image, (2) they are not standardized, and (3) alignment quality can be low. To address these limitations, we re-aligned all samples under the same unified format between the WSI and the corresponding expression profile. For all Visium samples, we developed an automatic alignment pipeline based on fiducial detection (see \Cref{sec:hestlib}) and embedded the alignment in the \textit{scanpy} object. For all Xenium samples, we used the publicly available VALIS~\cite{gatenbee2023valis} pipeline for fine-grained image registration to align the DAPI image (aligned with the transcripts by design) and the H\&E slide. 

\section{HEST-Library} \label{sec:hestlib}

The HEST-Library is built around \textit{scanpy} and {\small \textsc{Anndata}}. At its core, the HEST-Library enables (1) assembling and querying HEST-1k, (2) visualizing and mitigating batch effects, and (3) running the HEST-Benchmark (\Cref{sec:benchmark}). We describe its core functionalities, particularly for unifying legacy data.

\noindent\textbf{Conversion to generic TIFF.} We integrate functions to convert a WSI from common formats found in public ST datasets (e.g., OME.TIF, JPG, BigTIFF, etc.) to a pyramidal generic TIFF format. Pyramidal formats offer seamless integration with \textsc{OpenSlide} (commonly used in computational pathology pipelines) and \textsc{QuPath} (open-access software for WSI visualization and annotation).

\noindent\textbf{Automatic alignment in Visium.} Spot alignment is crucial to ensure an accurate match between the ST spots and the WSI. While software such as LoupeBrowser enables manual alignment using fiducials (i.e., reference markers placed at the corners of the capture area), the process remains time-consuming when processing large batches of samples. Instead, we implemented an automatic fiducial detection algorithm based on YOLOv8~\cite{redmon2016you} for processing Visium samples (Appendix \autoref{fig:fig_autoalign}). Specifically, we manually annotated 119 fiducial regions that we further augmented using tissue and fiducial mixing. We then fine-tuned YOLOv8 pretrained on the COCO dataset. In early versions that do not provide corner fiducials (e.g., STv1), we realigned using the provided spot position files. In Xenium, we use VALIS~\cite{gatenbee2023valis} to register the DAPI staining (aligned with the transcripts) with the H\&E image. 

\noindent\textbf{Automatic detection of image resolution.} From the alignment and the spot resolution, we can infer the exact pixel size. To this end, we compute the distance in pixel between two neighboring spots and leverage the known inter-spot distance in $\mu$m to estimate the pixel width in $\mu$m/px. For Xenium samples, we use the H\&E alignment file provided as part of the assay, which provides an affine transformation from the DAPI-stained image (with known pixel size) to the H\&E image. We then compared the self-reported image resolution and our re-estimations to manually inspect and correct discrepancies. 

\noindent\textbf{Conversion to {\small \textsc{Anndata}}.} ST data is provided in multiple formats, such as CSV, MEX, TXT, h5, etc. We provide functions to unify a large set of existing formats into a {\small \textsc{Anndata}} object that stores the raw transcript counts as a matrix of genes by the number of spots, in addition to metadata about the samples (e.g., the (x,y) coordinates of each spot, the pixel resolution, etc.). 

\noindent\textbf{Tissue segmentation and patching.} We provide a tissue segmentation pipeline optimized for Visium/Xenium images. The segmentation can then automatically tessellate the tissue into fixed-size image patches at a predefined resolution (expressed in $\mu$m/px) around each spot. 

\noindent\textbf{Automatic HEST-1k download.} To facilitate downloading part or all of the HEST-1k dataset (over $>$1TB), we implemented an easy download option where the user can specify entries of the metadata, for instance, to query all human invasive breast cancer cases. 

\noindent\textbf{Batch effect visualization and mitigation.} We provide functions to help visualize batch effects using dimensionality reduction techniques with user-prompted stratification (e.g., tissue site, institution, disease, etc.). In addition, we provide a wrapper of well-established batch effect mitigation strategies (namely ComBat~\cite{zhang2020combat}, Harmony~\cite{korsunsky2019harmony} and matching mutual nearest neighbors~\cite{haghverdi2018mnn}), which can be applied to a list of HEST samples. 

\section{HEST-Benchmark} \label{sec:benchmark}

From HEST-1k, we curated the HEST-Benchmark, a set of nine tasks for gene expression prediction from histology in human cancer samples. The goal is two-fold: (i) benchmarking foundation models for histology under a diverse and challenging benchmark and (ii) understanding the predictive capabilities of state-of-the-art models in predicting expression from morphology. Compared to existing tasks (e.g., Camelyon16~\cite{bejnordi2017diagnostic}), the HEST-Benchmark brings increased morphological diversity and more complex challenges, particularly with the inherent difficulty of expression prediction.

\subsection{Task definition}

We define nine tasks with data from eight human cancers and nine organs (eight primary and one metastatic dataset), which include \textbf{invasive ductal carcinoma} (breast cancer, IDC, Task 1), \textbf{prostate adenocarcinoma} (prostate cancer, PRAD, Task 2), \textbf{pancreatic adenocarcinoma} (pancreatic cancer, PAAD, Task 3), \textbf{skin cutaneous melanoma} (skin cancer, SKCM, Task 4), \textbf{colonic adenocarcinoma} (colon cancer, COAD, Task 5), \textbf{rectal adenocarcinoma} (rectum cancer, READ, Task 6), \textbf{clear cell renal cell carcinoma} (kidney cancer, ccRCC, Task 7), \textbf{lung adenocarcinoma}  (lung cancer, LUAD, Task 8), and \textbf{axillary lymph nodes in IDC} (metastatic, LYMPH-IDC, Task 9). Additional information is provided in Appendix~\autoref{tab:hestbenchmark}. 

For each task, we predict the expression of the top 50 genes with the highest normalized variance across all samples from 112$\times$112 $\mu$m H\&E regions (equivalent to 224$\times$224-pixel patches at 20$\times$). To avoid train/test patient-level data leakage, we use patient-stratified splits, resulting in a $k$-fold cross-validation, where $k$ is the number of patients. In ccRCC, we use $k/2$-fold cross-validation due to the large number of patients.

\subsection{Evaluating foundation model for pathology}


We use the HEST-Benchmark to evaluate 11 foundation models for pathology. Namely,
\textbf{ResNet50 (IN)~\cite{lu2020data}} (ImageNet pretrained), 
\textbf{CTransPath~\cite{wang2022transformer}} (adapted MoCov3 pretrained on TCGA and PAIP), 
\textbf{Remedis~\cite{azizi2023robust}} (SimCLR~\cite{chen2020simple} pretrained on TCGA),
\textbf{Phikon~\cite{filiot2023scaling}} (iBOT pretrained on TCGA),
\textbf{UNI~\cite{chen2024towards}} (DINOv2 ViT-Large pretrained on internal hospital data and GTEx), 
\textbf{CONCH~\cite{lu2024towards}} (visual-language model using CoCa pretrained on captions from publications and educational resources),
\textbf{GigaPath}~\cite{xu2024wholeslide} (DINOv2 ViT-giant pretrained on proprietary data),
\textbf{Virchow}~\cite{vorontsov2024foundation} (DINOv2 ViT-Huge pretrained on proprietary data),
\textbf{Virchow 2}~\cite{Zimmermann2024Virchow2SS} (DINOv2 ViT-Huge pretrained on proprietary data), 
\textbf{H-Optimus-0} (DINOv2 ViT-giant pretrained on proprietary data), and 
\textbf{UNIv1.5} (DINOv2 ViT-giant pretrained on public and proprietary data). 
Additional information is provided in~\autoref{tab:vision_encoders} and~\Cref{app:hestbenchmark}. 

\begin{table}[t]
\centering
\caption{\textbf{Comparison of 11 patch encoders evaluated on the HEST-Benchmark.} Reported results are based on PCA with 256 factors followed by a ridge regression. Model performance measured with Pearson correlation. Standard deviation is reported across all folds (i.e., patients). Best is \textbf{bold}, second best is \underline{underlined}.}
\label{tab:pca_ridge_hestbench}
\small
\scalebox{0.8}{
\begin{tabular}{lcccccccccc}
\toprule
{} &               \textbf{IDC} &              \textbf{PRAD} &              \textbf{PAAD} &              \textbf{SKCM} &              \textbf{COAD} &              \textbf{READ} &             \textbf{ccRCC} &              \textbf{LUAD} &         \textbf{LYMPH IDC} &           \textbf{Average} \\
\midrule

\textbf{ResNet50  (IN) } &                     0.4741 &                     0.3075 &                     0.3889 &                     0.4822 &                     0.2528 &                     0.0812 &                     0.2231 &                     0.4917 &                     0.2322 &                      0.326 \\
            &   \scriptsize{$\pm$ 0.047} &  \scriptsize{$\pm$ 0.0309} &  \scriptsize{$\pm$ 0.0754} &  \scriptsize{$\pm$ 0.1141} &  \scriptsize{$\pm$ 0.0372} &  \scriptsize{$\pm$ 0.0517} &  \scriptsize{$\pm$ 0.0554} &  \scriptsize{$\pm$ 0.0119} &  \scriptsize{$\pm$ 0.0491} &  \scriptsize{} \\

\textbf{CTransPath     } &                      0.511 &                     0.3427 &                     0.4378 &                     0.5106 &                     0.2285 &                       0.11 &                     0.2279 &                     0.4985 &                     0.2353 &                     0.3447 \\
            &  \scriptsize{$\pm$ 0.0531} &  \scriptsize{$\pm$ 0.0458} &  \scriptsize{$\pm$ 0.0664} &  \scriptsize{$\pm$ 0.0827} &  \scriptsize{$\pm$ 0.0557} &  \scriptsize{$\pm$ 0.0764} &  \scriptsize{$\pm$ 0.0475} &  \scriptsize{$\pm$ 0.0414} &  \scriptsize{$\pm$ 0.0477} &  \scriptsize{} \\

\textbf{Phikon    } &                     0.5327 &                      0.342 &                     0.4432 &                     0.5355 &                     0.2585 &                     0.1517 &                     0.2423 &                     0.5468 &                     0.2373 &                     0.3656 \\
            &  \scriptsize{$\pm$ 0.0914} &  \scriptsize{$\pm$ 0.0767} &  \scriptsize{$\pm$ 0.0684} &  \scriptsize{$\pm$ 0.0549} &  \scriptsize{$\pm$ 0.0056} &  \scriptsize{$\pm$ 0.0822} &  \scriptsize{$\pm$ 0.0263} &  \scriptsize{$\pm$ 0.0045} &  \scriptsize{$\pm$ 0.0457} &  \scriptsize{} \\

\textbf{CONCH      } &                     0.5363 &                     0.3548 &                     0.4475 &                     0.5791 &                     0.2533 &                     0.1674 &                     0.2179 &                     0.5312 &                     0.2507 &                     0.3709 \\
            &  \scriptsize{$\pm$ 0.0842} &  \scriptsize{$\pm$ 0.0099} &  \scriptsize{$\pm$ 0.0729} &  \scriptsize{$\pm$ 0.0542} &  \scriptsize{$\pm$ 0.0075} &  \scriptsize{$\pm$ 0.0476} &  \scriptsize{$\pm$ 0.0353} &  \scriptsize{$\pm$ 0.0107} &   \scriptsize{$\pm$ 0.042} &  \scriptsize{} \\

\textbf{REMEDIS } &                      0.529 &                     0.3471 &                     0.4644 &                     0.5818 &                     0.2856 &                     0.1145 &                     0.2647 &                     0.5336 &                     0.2473 &                     0.3742 \\
            &   \scriptsize{$\pm$ 0.069} &  \scriptsize{$\pm$ 0.0074} &  \scriptsize{$\pm$ 0.0722} &  \scriptsize{$\pm$ 0.0421} &    \scriptsize{$\pm$ 0.02} &  \scriptsize{$\pm$ 0.0987} &  \scriptsize{$\pm$ 0.0539} &  \scriptsize{$\pm$ 0.0326} &  \scriptsize{$\pm$ 0.0585} &  \scriptsize{} \\

\textbf{GigaPath   } &                     0.5508 &         \underline{0.3708} &                     0.4768 &                     0.5538 &          \underline{0.301} &                      0.186 &                     0.2391 &                     0.5399 &                     0.2493 &                     0.3853 \\
            &  \scriptsize{$\pm$ 0.0726} &   \scriptsize{$\pm$ 0.021} &  \scriptsize{$\pm$ 0.0489} &  \scriptsize{$\pm$ 0.0586} &  \scriptsize{$\pm$ 0.0145} &  \scriptsize{$\pm$ 0.0704} &  \scriptsize{$\pm$ 0.0364} &  \scriptsize{$\pm$ 0.0369} &  \scriptsize{$\pm$ 0.0522} &  \scriptsize{} \\

\textbf{UNI   } &                     0.5702 &                      0.314 &                     0.4764 &         0.6254 &                      0.263 &                     0.1762 &                     0.2427 &                     0.5511 &                     0.2565 &                     0.3862 \\
            &  \scriptsize{$\pm$ 0.0833} &  \scriptsize{$\pm$ 0.0715} &  \scriptsize{$\pm$ 0.0687} &  \scriptsize{$\pm$ 0.0338} &  \scriptsize{$\pm$ 0.0311} &  \scriptsize{$\pm$ 0.0565} &  \scriptsize{$\pm$ 0.0368} &  \scriptsize{$\pm$ 0.0198} &  \scriptsize{$\pm$ 0.0436} &  \scriptsize{} \\

\textbf{Virchow    } &                     0.5702 &                     0.3309 &         0.4875 &                     0.6088 &             \textbf{0.311} &                     0.2019 &                     0.2637 &                     0.5459 &         0.2594 &                     0.3977 \\
            &  \scriptsize{$\pm$ 0.0939} &  \scriptsize{$\pm$ 0.0081} &  \scriptsize{$\pm$ 0.0412} &  \scriptsize{$\pm$ 0.0733} &  \scriptsize{$\pm$ 0.0083} &  \scriptsize{$\pm$ 0.0467} &   \scriptsize{$\pm$ 0.039} &  \scriptsize{$\pm$ 0.0262} &   \scriptsize{$\pm$ 0.043} &  \scriptsize{} \\

\textbf{Virchow2        } &         0.5922 &                     0.3465 &                     0.4661 &                     0.6174 &                     0.2578 &         0.2084 &            \textbf{0.2788} &            \textbf{0.5605} &                     0.2582 &         0.3984 \\
            &  \scriptsize{$\pm$ 0.0814} &   \scriptsize{$\pm$ 0.029} &  \scriptsize{$\pm$ 0.0676} &  \scriptsize{$\pm$ 0.0174} &  \scriptsize{$\pm$ 0.0189} &  \scriptsize{$\pm$ 0.0502} &  \scriptsize{$\pm$ 0.0516} &  \scriptsize{$\pm$ 0.0172} &  \scriptsize{$\pm$ 0.0296} &  \scriptsize{} \\

\textbf{UNIv1.5} & \textbf{0.5989} & 0.3645 & \underline{0.4902} & \underline{0.6401} & 0.2925 & \underline{0.2240} & 0.2522 & \underline{0.5586} & \textbf{0.2597} & \underline{0.4090} \\
            &  \scriptsize{$\pm$ 0.0842} &  \scriptsize{$\pm$ 0.0308} &  \scriptsize{$\pm$ 0.0502} &  \scriptsize{$\pm$ 0.041} &  \scriptsize{$\pm$ 0.0142} &  \scriptsize{$\pm$ 0.0378} &  \scriptsize{$\pm$ 0.04} &  \scriptsize{$\pm$ 0.026} &  \scriptsize{$\pm$ 0.0424} &   \scriptsize{} \\

\textbf{H-Optimus-0} &            \underline{0.5982} &             \textbf{0.385} &            \textbf{0.4932} &            \textbf{0.6432} &                     0.2991 &            \textbf{0.2292} &         \underline{0.2654} &         0.5582 &            \underline{0.2595} &            \textbf{0.4146} \\
            &  \scriptsize{$\pm$ 0.0843} &  \scriptsize{$\pm$ 0.0008} &  \scriptsize{$\pm$ 0.0443} &  \scriptsize{$\pm$ 0.0668} &  \scriptsize{$\pm$ 0.0007} &   \scriptsize{$\pm$ 0.041} &  \scriptsize{$\pm$ 0.0309} &  \scriptsize{$\pm$ 0.0324} &    \scriptsize{$\pm$ 0.04} &  \scriptsize{} \\

\bottomrule
\end{tabular}
}
\end{table}

We learn a regression model to map model-specific patch embeddings (512 to 2,048 dimensions) to the log1p-normalized expression of the top 50 highly variable genes. All tasks are evaluated using the Pearson correlation between the predicted and measured gene expression. We report mean and standard deviation across all folds (or patients). All experiments were run on a single NVIDIA 3090 GPU. We report performance using three downstream regression models: (i) PCA-reduced embeddings (with n=256 factors) followed by Ridge regression trained with adaptive regularization as shown in \autoref{tab:pca_ridge_hestbench}, (ii) Ridge regression model as shown in Appendix \autoref{tab:ridge_hestbench}, and (iii) an XGBoost regression model with 100 estimators and a maximum depth of 3 as shown in Appendix~\autoref{tab:xgboost_hestbench}. Our main results are reported using PCA+Ridge (i) and XGBoost (iii). Directly applying Ridge regression may unfairly penalize models with larger embedding dimensions. To guarantee a fairer and more objective comparison, we chose to utilize PCA reduction.

\subsection{Scaling laws in HEST-Benchmark}

Overall, H-Optimus-0 brings the best average Pearson correlation in both PCA+Ridge and XGBoost evaluation, outperforming the second-best model, UNIv1.5, by 0.56\% and 0.69\%, respectively. ResNet50 (IN), the only model that was not pretrained on histology images, leads to the lowest performance in both PCA+Ridge and XGBoost. Legacy domain-specific models, such as CTransPath, are outperformed by all recent models, including UNIv1.5, UNI, GigaPath, Virchow, and H-Optimus-0. The disparity between the top and bottom domain-specific models is notable, showing an absolute improvement of 7.0\% for PCA+Ridge and 4.8\% for XGBoost. When inspecting individual performance, we observe large differences across tasks from 0.6432 Pearson correlation in SKCM to 0.2292 in READ for H-Optimus-0 evaluated using PCA+Ridge. 

\begin{figure}[t]
   \centering
   \resizebox{\linewidth}{!}{
       \includegraphics{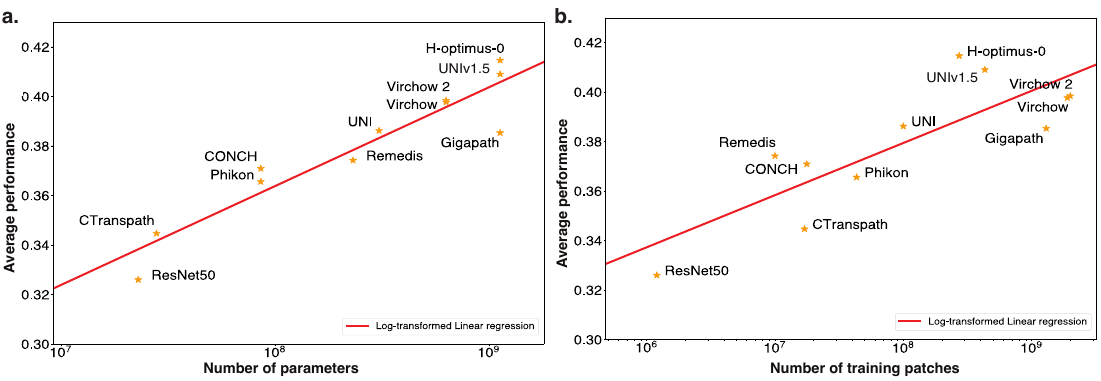} 
   }
   \caption{
       \textbf{Scaling laws in HEST-Benchmark.}
       \textbf{a.} Model scaling law comparing the number of training parameters in the vision encoder (log-scale) and the average performance on the HEST-Benchmark. Pearson correlation between parameters and performance of R=0.81 (P-value < 0.01). 
       \textbf{b.} Data scaling law comparing the number of image patches used for pretraining (log-scale) and the average performance on the HEST-Benchmark. Pearson correlation between number of patches and performance of R=0.48 (P-value=0.13). 
   } 
   \label{fig:scaling_law}
\end{figure}

\noindent\textbf{Model scaling law.} By inspecting the number of trainable parameters within the vision encoder for each model, we can describe how model size influences performance (measured with average Pearson correlation across all tasks, \autoref{fig:scaling_law}.a). Performance increases with model size following a logarithmic scaling law (Pearson correlation of R=0.81, P-value$<$0.01). Models considered ``parameter-efficient'' are represented on top of the log-transformed linear regression line (e.g., CONCH, UNIv1.5, and H-Optimus-0). This observation suggests a trade-off between downstream performance and model size. Despite the observation of a model scaling law, significant variations in performance among models of identical size persist, such as between H-Optimus-0 and GigaPath, both of which are ViT-giant models with over one billion parameters. 

\noindent\textbf{Data scaling law.} We further explored how the number of training samples used for pretraining each model (i.e., the number of image patches) affects performance. We observe that increasing the number of patches moderately correlates with the average performance (Pearson correlation of R=0.48, P-value=0.13). This correlation is weaker than model size, which we hypothesize is due to this analysis overlooking both the absolute number of WSIs used for pretraining (image patches are not independently and identically distributed per WSI) and disparities of morphological variety among WSIs (e.g., staining variation, disease diversity, artifacts, etc.).

Overall, HEST-Benchmark brings new insights into the performance of foundation models for pathology. We observe that
(1) Scaling the model size strongly correlates with average performance, but the gains grow logarithmically with the number of trainable parameters.
(2) Scaling the number of training patches weakly correlates with a higher performance (also on a logarithmic scale).  
(3) Performance remains low for some tasks (e.g., READ and ccRCC), which suggests that (i) the morphology might not be as reflective of gene expression for some cancer types or (ii) some cohorts have more noise than others (e.g., due to batch effects, low sensitivity, dropout events, or spillover between adjacent spots).

\section{HEST for biomarker exploration}

\begin{figure}
   \centering
   \resizebox{\linewidth}{!}{
       \includegraphics{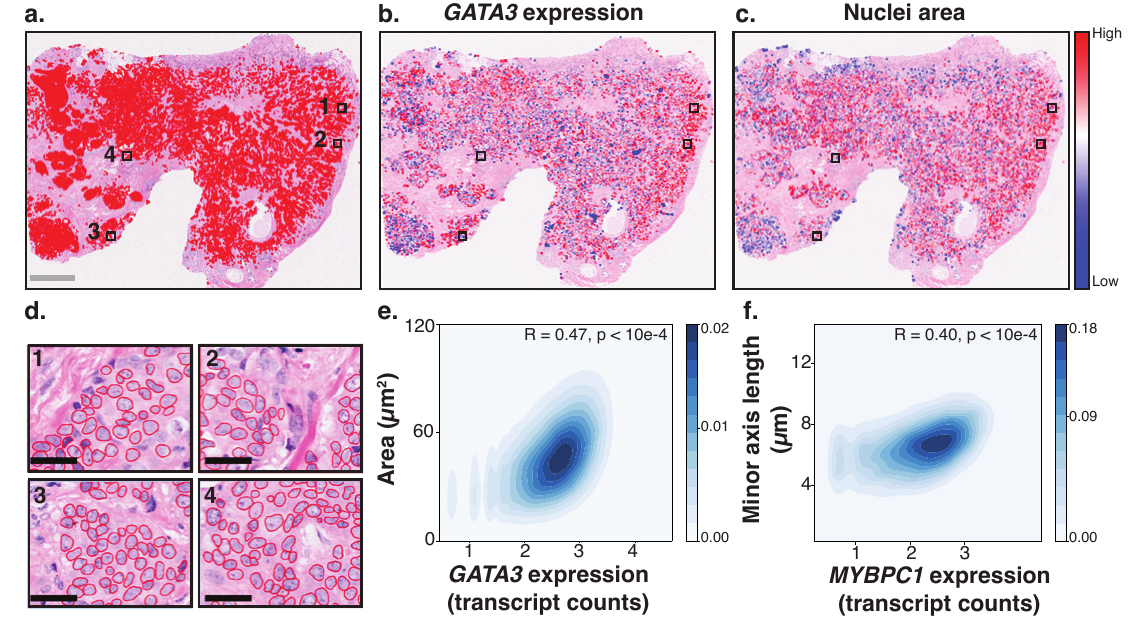} 
   }
   \caption{
       \textbf{HEST for biomarker exploration: Analysis of an invasive ductal carcinoma sample imaged with Xenium.}
       \textbf{a.} IDC Xenium sample with neoplastic nuclei overlaid in red ($n_c=168,033$ detected nuclei). Gray scale bar represents 2 mm.
       \textbf{b.} Heatmap of Xenium expression of gene \textit{GATA3}. Blue and red values indicate above and below the mean (in white), respectively.
        \textbf{c.} Heatmap of neoplastic nuclear area.
        \textbf{d.} Four randomly selected regions with CellViT segmentation of the neoplastic nuclei. Black scale bar represents 30 $\mu$m. 
       \textbf{e.}, \textbf{f.} Correlation between nuclear area and \textit{GATA3}, and minor axis length and \textit{MYBPC1}.
   } 
   \label{fig:discovery}
\end{figure}

HEST-1k also enables the analysis of interactions and correlations between tissue morphology (as seen in H\&E) and local gene expression (as provided in ST). Here, we showcase the capabilities of HEST-1k (1) by studying morphological correlates of expression changes in invasive breast cancer and (2) by visualizing tumor heterogeneity both on the morphological and molecular sides. Specifically, we focus on invasive ductal carcinoma (IDC) samples imaged with Xenium. Using CellViT nuclear segmentation and classification, we identified neoplastic nuclei (exemplified in two samples: \autoref{fig:discovery}.a with n=168,033 nuclei and Appendix~\autoref{fig:discovery_extended}.a with n=342,018 nuclei). We then overlay the WSI with the expression of specific genes, such as \textit{GATA3}, a known prognostic gene in breast cancer~\cite{mehra2005identification}(\autoref{fig:discovery}.b). This qualitatively shows that high \textit{GATA3} expression is associated with cancerous regions and reveals heterogeneity within invasive regions (e.g., the right-most region shows higher expression of \textit{GATA3} than the rest of the tumor, \autoref{fig:discovery}.b). Using the nuclear segmentation, we can compute human-interpretable features related to nuclear size (area, perimeter, major axis length, minor axis length, and equivalent diameter), topology and shape (roundness, ellipticity, eccentricity, extent, and roughness), and cell distribution (cell density and crowdedness). A heatmap of the nuclear area of neoplastic cells also indicates morphological heterogeneity among neoplastic regions (\autoref{fig:discovery}.c,d). Regions with a high nuclear area and elevated GATA3 expression notably overlap, suggesting that this tumor exhibits molecular heterogeneity, which to some degree is morphologically expressed. 

To investigate this hypothesis, we measured the Pearson correlation between the expression of \textit{GATA3} and nuclear area in neoplastic cells (\autoref{fig:discovery}.e). We observe a moderate correlation (R=0.47, P-value$<10^{-4}$), which is also observed in other genes and morphological features (\autoref{fig:discovery}.f). Overall, out of the 12 human-interpretable features we analyzed, we found the highest association with gene expression for size-related features, while features involving topology, shape, and cell distribution had a lower correlation (R<0.2). A similar analysis in another IDC sample (Appendix~\autoref{fig:discovery_extended}.b,c) further asserted these observations. In particular, we found the highest associations between nuclear size and expression for the genes \textit{FLNB} (R=0.45, P-value$<10^{-4}$) and \textit{TPD52} (R=0.47, P-value$<10^{-4}$), both involved in breast tumor growth and proliferation \cite{bandaru2014targeting,ren2021tumor}, and \textit{FOXA1} (R=0.47, P-value$<10^{-4}$), a known prognostic factor associated with better survival \cite{badve2007foxa1,wolf2007foxa1}. 

Such analysis highlights how HEST-1k can be used to identify fine-grained morphological correlates of expression. Similar approaches can be used to characterize morphological and molecular tumor heterogeneity at a larger scale.

\section{HEST for multimodal representation learning} \label{sec:dl}

Access to spatially-resolved expression--morphology pairs unlocks new directions for multimodal representation learning. Several problem statements can be explored, such as cross-modal alignment and retrieval, multimodal fusion, etc. Here, we fine-tune CONCH~\cite{lu2024towards} (ViT-Base model) on five Xenium invasive breast cancer cases (four ductal and one lobular case) using multimodal contrastive alignment. We hypothesize that the resulting breast cancer-specific patch encoder, termed CONCH-FT, can better encode the underlying molecular landscape associated with disease-specific morphologies. To validate the hypothesis, CONCH-FT is benchmarked on an independent breast cancer cohort for molecular subtyping against its non-finetuned version.

Specifically, for each Xenium sample, we extract 112$\times$112-$\mu$m image patches centered around each spot at 20$\times$ magnification (0.5$\mu$m/px). This yields 47,051 pairs of 224$\times$224-pixel patches and corresponding expression profile (n=238 common genes in the panel, log1p normalized). We then embed the data using modality-specific encoders: the image patches using a pretrained CONCH model and the expression data using a 3-layer MLP (normalized expression data are encoded as tabular data). The modality-specific embeddings are then aligned using a contrastive objective, i.e., InfoNCE loss~\cite{oord2018representation} by fine-tuning the image encoder and training the expression encoder from scratch. To mitigate over-fitting, we use the following training recipe: (1) Finetune only the last 3 layers of CONCH, (2) employ a layer-wise learning decay factor of 0.7, and (3) employ patch-level image augmentation. Additional details are provided in Appendix.

We evaluate the resulting CONCH-FT model to predict ER, PR, and HER2 expression status (binary) from WSIs in the BCNB dataset~\cite{xu2021predicting} (n=1,058 WSIs). To generate a slide representation for a WSI, we take the average of the patch embedding in the WSI (mean pooling), which is subsequently mapped to the expression status using logistic regression (Table~\ref{tab:finetuning}). The simple mean pooling approach to embedding the slide without additional fine-tuning on the downstream tasks highlights the expressivity of the learned latent space. We observe that CONCH-FT outperforms CONCH on most metrics, demonstrating that pan-tissue histology patch encoders can be further fine-tuned to obtain better tissue-specific patch encoders. This is further validated by the larger rank induced by the patch embedding space~\cite{garrido2023rankme} for CONCH-FT, suggesting better expressivity of the patch embeddings. While these results are based on only five paired WSIs, we anticipate additional benefits when training with larger disease-specific cohorts.

\begin{table}[t]
\centering
\caption{\textbf{CONCH fine-tuning on invasive breast cancer.} Logistic regression evaluation for ER/PR/HER2 status on BCNB (binary task, n=1,058 WSIs). A WSI is represented by the average of the patch embeddings within each WSI. We report the mean ± standard deviation computed over all folds (or patients) for ROC-AUC (AUC) and balanced accuracy (Bal.acc.). Best is \textbf{bold}.}
\label{tab:finetuning}
\small
\scalebox{1.0}{
\begin{tabular}{lccccccc}
\toprule
 & \multirow{2}{*}{\textbf{Rank}} & \multicolumn{2}{c}{\textbf{ER}} & \multicolumn{2}{c}{\textbf{PR}} & \multicolumn{2}{c}{\textbf{HER2}}  \\

 &&AUC & Bal.acc. & AUC & Bal.acc. & AUC & Bal.acc. \\
\midrule

\multirow{1}{*}{\textbf{CONCH}} & \multirow{1}{*}{144.66} & 0.881 & 0.745 & 0.810 & 0.698 & 0.715 & $\bm{0.624}$ \\ 

\multirow{1}{*}{\textbf{CONCH-FT}} & \multirow{1}{*}{$\bm{146.47}$} & $\bm{0.884}$ & $\bm{0.752}$& $\bm{0.818}$ & $\bm{0.714}$ & $\bm{0.724}$ & 0.615 \\ 

\bottomrule
\end{tabular}
}
\end{table}

\section{Discussion}

\noindent\textbf{Summary.} We assembled HEST-1k, a dataset comprising paired spatial transcriptomics, H\&E-stained whole-slide images, and comprehensive metadata built from public and internal cohorts. HEST-1k includes 1,229 samples, encompassing 2.1 million spots and over 76 million cells. The scale and comprehensiveness of HEST-1k, supported by the HEST-Library, enable exploring directions such as biomarker exploration and multimodal representation learning. Additionally, motivated by the need for new, diverse, and challenging patch-level benchmarks, we curated the HEST-Benchmark, a set of nine tasks covering eight cancer types and nine organs for gene expression prediction from histology. The HEST-Benchmark revealed data and model scaling laws across 11 foundation models of different dimensions and pretraining scale~\cite{lu2024towards}.

\noindent\textbf{Limitations.} Our study includes a few limitations. First, research data, such as those generated in spatial transcriptomic, are inherently noisy. While we tried to minimize ``label'' noise (e.g., by re-estimating image magnification and alignment, and unifying cancer samples using oncotree code taxonomy), staining and compression artifacts, varying acquisition protocols, among others, can negatively impact the quality of HEST-1k. Second, batch effects (on both the imaging and transcriptomic sides) can be significant across samples, datasets, and technologies. While this study does not explore batch effects quantification or mitigation, we provide a set of helpers in HEST-Library to let users explore this direction. Lastly, although the HEST-Library was designed for versatility, it cannot cover all existing formats and should rather be viewed as a blueprint for processing ST data in a consistent and unified manner.

\noindent\textbf{Future work.} Spatial transcriptomics is rapidly evolving, with new datasets frequently published. As they become available, we will keep updating HEST-1k with new resources. This study merely starts to uncover the potential of HEST-1k for advancing translational research and biomarker exploration, and we plan to explore these capabilities further. Additionally, the prospects for multimodal representation learning with HEST-1k are promising and are expected to grow with the addition of more data.

\section*{Acknowledgements}

We thank Dr. Maxime Meylan for his insights and guidance on accessing data published in~\cite{meylan2022tertiary}. We thank Rushin Gindra for his support in inspecting HEST-1k data, reporting issues, and providing references. HEST is supported by the Brigham and Women's Hospital (BWH) President's Fund, Mass General Hospital (MGH) Pathology, and the National Institute of Health (NIH) National Institute of General Medical Sciences (NIGMS) through R35GM138216. S.J.W. is supported by the Helmholtz Association under the joint research school ``Munich School for Data Science - MUDS'' and the Add-on Fellowship of the Joachim Herz Foundation.

\clearpage
\section*{Checklist}

\begin{enumerate}
  \item Do the main claims made in the abstract, and introduction accurately reflect the paper's contributions and scope?
    \answerYes{Each claim: HEST-1k, HEST-Library, HEST-Benchmark, HEST for biomarker exploration, and multimodal fine-tuning are supported by dedicated sections in the main text, in addition to supplementary information provided in the appendix. In addition, all the code to reproduce results is made available.}
  \item Did you describe the limitations of your work?
    \answerYes{We discuss limitations in the \textbf{Discussion}.}
  \item Did you discuss any potential negative societal impacts of your work?
    \answerYes{We discuss potential negative societal impacts in the section \textbf{Ethical considerations, intended usage, and license}.}
  \item Have you read the ethics review guidelines and ensured that your paper conforms to them?
    \answerYes{}
  \item Did you include the code, data, and instructions needed to reproduce the main experimental results (either in the supplemental material or as a URL)?
    \answerYes{In the abstract, we provide a link to access the HEST page on GitHub. HEST-Library includes a link to download all data and to run the HEST-Benchmark. Finally, we provide all metadata associated with HEST-1k in a CSV as part of the supplementary material.}
  \item Did you specify all the training details (e.g., data splits, hyperparameters, how they were chosen)?
    \answerYes{When relevant, we provide training details, such as in the \textbf{HEST-Benchmark}.}
	\item Did you report error bars?
    \answerYes{HEST-Benchmark results include standard deviation computed from cross-validation across all patients.}
	\item Did you include the total amount of compute and the type of resources used (e.g., type of GPUs, internal cluster, or cloud provider)?
    \answerYes{}
  \item If your work uses existing assets, did you cite the creators?
    \answerYes{All public resources used in this study are cited in Appendix \autoref{tab:tenx}, \ref{tab:ncbi}, \ref{tab:spa}, \ref{tab:mend}, \ref{tab:misc}, \ref{tab:internal} and \ref{tab:zenodo}.}
  \item Did you mention the license of the assets?
    \answerYes{Metadata associated with HEST-1k includes the license under which data were originally published. We ensured that the reported license allowed the distribution and creation of derivatives of the data.}
  \item Did you include any new assets either in the supplemental material or as a URL?
    \answerYes{As part of HEST-1k, we include internal datasets (see Appendix~\autoref{tab:internal}).}
  \item Did you discuss whether and how consent was obtained from people whose data you're using/curating?
    \answerNo{We used public resources for which the license was allowing redistributing the work. Users are welcome to inspect the individual IRBs of each publicly available resource.}
  \item Did you discuss whether the data you are using/curating contains personally identifiable information or offensive content?
    \answerYes{We manually ensured that none of the published and distributed data includes personally identifiable information or offensive content, such as personal health information.}
\end{enumerate}

\clearpage
\appendix

\renewcommand{\contentsname}{Appendix}
\tableofcontents
\addtocontents{toc}{\protect\setcounter{tocdepth}{2}}

\section{Ethical considerations, intended usage and license}

All resources provided as part of this study are strictly for research purposes and must not be utilized to support any diagnostic procedures. Users are hereby notified that the nuclear segmentation and classification components are derived from a publicly available model. Consequently, this model should not be regarded as the definitive standard, and users should exercise particular caution when utilizing this part of the dataset. Despite our efforts to exclude sensitive information, such as patient names, addresses, and social security numbers, users are expressly prohibited from attempting to reverse engineer the data to extract confidential patient information. In the presumption that users will adhere to the aforementioned restrictions, we have not identified any potential adverse social impacts that could arise from the use of HEST-1k.

The dataset is hosted on the HuggingFace Dataset webpage. All instructions are provided in the main README of HEST-Library. From there, users can choose to download HEST-1k in its entirety or a subset (e.g., only breast cancer samples). The HEST-1k, HEST-Benchmark, and HEST library are released under the Attribution-NonCommercial-ShareAlike 4.0 International license (CC BY-NC-SA 4.0 Deed)\footnote{https://creativecommons.org/licenses/by-nc-sa/4.0/}.

\section{Background}

This study connects two fields: (1) computational pathology, which primarily uses routinely acquired clinical data to determine outcomes such as disease diagnosis from H\&E-stained digitized tissue sections, and (2) spatial transcriptomics, which so far has been confined to biological research and aims to identify new biomarkers predictive of disease progression or response to treatment, among others. 

\subsection{Computational pathology}

Research in computational pathology~\cite{song2023artificial} has primarily focused on classifying digitized WSIs into clinical outcomes. Unlike natural image classification tasks such as ImageNet, a WSI may reach sizes of up to 150,000 $\times$ 150,000 pixels at 20$\times$ magnification (0.5$\mu$m/pixel). The challenge of managing the large size of WSIs has been one the central themes of the field, primarily through the adoption of multiple instance learning (MIL) for weakly-supervised classification~\cite{ilse2018attention}. MIL employs a two-step process: (1) Initially, the tissue is segmented from the background and then tessellated into patches, usually 256 $\times$ 256 pixels, akin to an ImageNet sample, and each patch is compressed into a patch embedding using a pretrained patch encoder. (2) Subsequently, these patch embeddings are aggregated using a learnable neural network, such as an attention-based network, a graph neural network, or a Transformer, to produce a slide embedding~\cite{ilse2018attention,shao2021transmil}. This slide embedding is then used to classify specific targets of interest, such as cancer histological subtyping, morphological subtyping, mutation prediction, or survival analysis.

Such frameworks have been shown to achieve better or similar performance than humans for Gleason grading in prostate cancer\cite{bulten2022artificial}, metastasis detection in lymph nodes\cite{bejnordi2017diagnostic}, determining the origin of a cancer of unknown primary\cite{lu2021ai}, predicting heart transplant rejection\cite{lipkova2022deep}, among others. 

\subsection{Spatial transcriptomics (ST)}

ST enables the measurement of gene activity and the mapping of its corresponding location in the tissue. In this study, we collected samples from two ST paradigms: sequencing-based (ST, Visium, Visium HD) and imaging-based (Xenium).

\noindent\textbf{Visium (HD) / Spatial Transcriptomics:} Visium-HD and its predecessors Visium and Spatial Transcriptomics (STv1) refer to a family of sequencing-based products for spatially resolving large transcript panels, whose main difference lies in the resolution and spacing between the expression measurement, called a spot. These spots capture mRNA from tissue sections placed on the chip, and the location-specific barcodes contained in each of the spots bind to the RNA to retain spatial information. The RNA molecules are then washed off the slides and processed by a sequencing instrument. Using a sequencing-based method allows the reuse of existing sequencing instruments developed in the fields of single-cell and bulk transcriptomics, hence benefiting from existing technological advancements and allowing whole transcriptome analysis. A fundamental drawback of current sequencing-based methods is the inherent RNA resolution limitation imposed by the size of the spots (e.g., 55$\mu$m in Visium).

\noindent\textbf{Xenium:} Xenium is an imaging-based spatial profiling technology that offers in situ RNA capturing on tissue sections by imaging fluorescent RNA markers derived from padlock probes and rolling circle amplification chemistry. This approach provides the exact 2D location of each measured transcript. As of 2024, Xenium cannot perform whole transcriptome measurements and is limited to gene panels of up to 5,000 genes. 

\section{HEST}

\subsection{HEST-1k}

We provide a comprehensive description of all publicly available and internal cohorts integrated into HEST-1k. 

\setcounter{table}{0}
\renewcommand{\thetable}{A\arabic{table}}
\renewcommand*{\theHtable}{\thetable}

\begin{table}[t]
    \centering
    \caption{\textbf{HEST-1k data overview.} All samples include a license that allows sharing and redistributing. National Center for Biotechnology Information: NCBI.
    }
    \label{tab:hest1k}
    \begin{tabular}{lccc}
        \toprule
        \textbf{Resource} & \textbf{Number of datasets} & \textbf{Number of samples} & \textbf{Size (GB, raw)}  \\ 
        \midrule
        10x Genomics & 87 & 112 & 275   \\
        Mendeley & 9 & 118 & 181   \\
        Spatial-Research & 4 & 139 & 18   \\
        Zenodo & 4 & 21 & 18   \\
        NCBI & 43 & 696 & 298   \\
        Internal & 3 & 28 & 60  \\
        Miscellaneous & 4 & 114 & 147  \\
        \bottomrule
    \end{tabular}
\end{table}

\begin{table}[ht]
\centering  
\small
\caption{\textbf{Datasets gathered from 10x Genomics portal.} $n$: number of samples in the cohort.}  
\label{tab:tenx}  
\begin{tabular}{m{7.5cm} m{1.3cm} m{1.5cm} >{\centering\arraybackslash}m{.2cm} >{\centering\arraybackslash}m{1cm}}
\toprule
\textbf{Collection name} & \textbf{Organ} & \textbf{Technology} & $\bm{n}$ & \textbf{Num. genes} \\
\midrule
Adult Mouse Brain (FFPE) & Brain & Visium & 1 & 19,465 \\
Adult Mouse Brain Coronal Section (Fresh Frozen) 1 & Brain & Visium & 1 & 32,285 \\
Adult Mouse Brain Coronal Section (Fresh Frozen) 2 & Brain & Visium & 1 & 32,285 \\
Adult Mouse Kidney (FFPE) & Kidney & Visium & 1 & 19,465 \\
Adult Mouse Olfactory Bulb & Brain & Visium & 1 & 32,285 \\

Characterization of immune cell populations in the tumor microenvironment of colorectal cancer using high definition spatial profiling \cite{oliveira2024characterization} & Bowel & Mixed & 8 & 18,085\\

FFPE Human Breast using the Entire Sample Area & Breast & Xenium & 2 & 541 \\
FFPE Human Breast with Custom Add-on Panel & Breast & Xenium & 2 & 541 \\
FFPE Human Breast with Pre-designed Panel & Breast & Xenium & 2 & 541 \\
FFPE Human Pancreas with Xenium Multimodal Cell Segmentation & Pancreas & Xenium & 1 & 541 \\
FFPE Human Prostate Adenocarcinoma with 5K Human Pan Tissue and Pathways Panel & Prostate & Xenium & 1 & 10,006\\

FFPE Human Skin Primary Dermal Melanoma with 5K Human Pan Tissue and Pathways Panel & Skin & Xenium & 1 & 10,017\\

Fresh Frozen Mouse Colon with Xenium Multimodal Cell Segmentation & Bowel & Xenium & 1 & 541 \\

Fresh Frozen Mouse Brain Hemisphere with 5K Mouse Pan Tissue and Pathways Panel & Brain & Xenium & 1 & 13,780\\

Fresh Frozen Visium on CytAssist: Human Breast Cancer, Probe-Based Whole Transcriptome Profiling & Breast & Visium & 1 & 18,085 \\
Fresh Frozen Visium on CytAssist: Mouse Brain, Probe-Based Whole Transcriptome Profiling & Brain & Visium & 1 & 19,465 \\
Human Bone and Bone Marrow Data with Custom Add-on Panel & Bone & Xenium & 3 & 541 \\
Human Brain Cancer, 11 mm Capture Area (FFPE) & Brain & Visium & 1 & 18,085 \\
Human Breast Cancer (Block A Section 1) & Breast & Visium & 1 & 33,538 \\
Human Breast Cancer (Block A Section 2) & Breast & Visium & 1 & 33,538 \\
Human Breast Cancer: Ductal Carcinoma In Situ, Invasive Carcinoma (FFPE) & Breast & Visium & 1 & 17,943 \\
Human Breast Cancer: Targeted, Immunology Panel & Breast & Visium & 1 & 1,056 \\
Human Breast Cancer: Visium Fresh Frozen, Whole Transcriptome & Breast & Visium & 1 & 36,601 \\
Human Breast Cancer: Whole Transcriptome Analysis & Breast & Visium & 1 & 32,285 \\
Human Cerebellum: Targeted, Neuroscience Panel & Brain & Visium & 1 & 1,186 \\
Human Cerebellum: Whole Transcriptome Analysis & Brain & Visium & 1 & 1,186 \\
Human Cervical Cancer (FFPE) & Cervix & Visium & 1 & 17,943 \\
Human Colon Preview Data (Xenium Human Colon Gene Expression Panel) & Bowel & Xenium & 2 & 541 \\
Human Colorectal Cancer, 11 mm Capture Area (FFPE) & Bowel & Visium & 1 & 18,085 \\
Human Colorectal Cancer: Targeted, Gene Signature Panel & Bowel & Visium & 1 & 1,142 \\
Human Colorectal Cancer: Whole Transcriptome Analysis & Bowel & Visium & 1 & 36,601 \\
Human Glioblastoma: Targeted, Pan-Cancer Panel & Brain & Visium & 1 & 1,253 \\
Human Glioblastoma: Whole Transcriptome Analysis & Brain & Visium & 1 & 36,601 \\
Human Heart & Heart & Visium & 1 & 36,601 \\
Human Heart Data with Xenium Human Multi-Tissue and Cancer Panel & Heart & Xenium & 1 & 541 \\
Human Intestine Cancer (FPPE) & Bowel & Visium & 1 & 17,943 \\
Human Kidney Preview Data (Xenium Human Multi-Tissue and Cancer Panel) & Kidney & Xenium & 2 & 541 \\
Human Kidney, 11 mm Capture Area (FFPE) & Kidney & Visium & 1 & 18,085 \\
Human Liver Data with Xenium Human Multi-Tissue and Cancer Panel & Liver & Xenium & 2 & 541 \\
Human Lung Cancer (FFPE) & Lung & Visium & 1 & 18,085 \\
Human Lung Cancer, 11 mm Capture Area (FFPE) & Lung & Visium & 1 & 18,085 \\
Human Lymph Node & Lymph node & Visium & 1 & 36,601 \\
\bottomrule
\end{tabular}
\end{table}

\begin{table}[ht]
\centering 
\small
\caption{\textbf{Datasets gathered from 10x Genomics portal. Continuation.}}
\label{tab:tenx2} 
\begin{tabular}{m{7.5cm} m{1.3cm} m{1.5cm} >{\centering\arraybackslash}m{.2cm} >{\centering\arraybackslash}m{1cm}}
\toprule
\textbf{Collection name} & \textbf{Organ} & \textbf{Technology} & $\bm{n}$ & \textbf{Num. genes} \\
\midrule
Human Ovarian Cancer (FFPE) & Ovary & Visium & 1 & 17,943 \\
Human Ovarian Cancer, 11 mm Capture Area (FFPE) & Ovary & Visium & 1 & 18,085 \\
Human Prostate Cancer, Acinar Cell Carcinoma (FFPE) & Prostate & Visium & 1 & 17,943 \\
Human Prostate Cancer, Adenocarcinoma with Invasive Carcinoma (FFPE) & Prostate & Visium & 1 & 17,943 \\
Human Skin Data with Xenium Human Multi-Tissue and Cancer Panel & Skin & Xenium & 2 & 541 \\
Human Skin Preview Data (Xenium Human Skin Gene Expression Panel with Custom Add-On) & Skin & Xenium & 1 & 541 \\
Human Skin Preview Data (Xenium Human Skin Gene Expression Panel) & Skin & Xenium & 1 & 541 \\
Human Tonsil Data with Xenium Human Multi-Tissue and Cancer Panel & Lymph node & Xenium & 2 & 541 \\
Mouse Bone Data with Custom Add-on Panel & Bone & Xenium & 3 & 541 \\
Mouse Brain Coronal Section 1 (FFPE) & Brain & Visium & 1 & 19,465 \\
Mouse Brain Coronal Section 2 (FFPE) & Brain & Visium & 1 & 19,465 \\
Mouse Brain Section (Coronal) & Brain & Visium & 1 & 31,053 \\
Mouse Brain Serial Section 1 (Sagittal-Anterior) & Brain & Visium & 1 & 31,053 \\
Mouse Brain Serial Section 1 (Sagittal-Posterior) & Brain & Visium & 1 & 31,053 \\
Mouse Brain Serial Section 2 (Sagittal-Anterior) & Brain & Visium & 1 & 31,053 \\
Mouse Brain Serial Section 2 (Sagittal-Posterior) & Brain & Visium & 1 & 31,053 \\
Mouse Embryo, 11 mm Capture Area (FFPE) & Embryo & Visium & 1 & 19,465 \\
Mouse Kidney Section (Coronal) & Kidney & Visium & 1 & 31,053 \\
Mouse Tissue Microarray in 3x3 Layout with 1 mm Edge to Edge Spacing (FFPE) & Lung/Brain & Visium & 1 & 19,465 \\
Mouse Tissue Microarray in 3x3 Layout with 2 mm Edge to Edge Spacing (FFPE) & Lung/Brain & Visium & 1 & 19,465 \\
Mouse Tissue Microarray in 5x5 Layout with 1 mm Edge to Edge Spacing (FFPE) & Kidney/Brain & Visium & 1 & 19,465 \\
Normal Human Prostate (FFPE) & Prostate & Visium & 1 & 17,943 \\
Pancreatic Cancer with Xenium Human Multi-Tissue and Cancer Panel & Pancreas & Xenium & 1 & 538 \\
Preservation Method Comparison on CytAssist: FFPE Mouse Brain (Sagittal), 11 mm Capture Area & Brain & Visium & 1 & 19,465 \\
Preservation Method Comparison on CytAssist: Fixed Frozen Mouse Brain (Sagittal), 11 mm Capture Area & Brain & Visium & 1 & 19,465 \\
Preservation Method Comparison on CytAssist: Fresh Frozen Mouse Brain (Sagittal), 11 mm Capture Area & Brain & Visium & 1 & 19,465 \\
Preservation Method Comparison on Visium CytAssist: FFPE Mouse Brain (Sagittal), 11 mm Capture Area & Brain & Visium & 1 & 19,465 \\
Preservation Method Comparison on Visium CytAssist: Fixed Frozen Mouse Brain (Sagittal), 11 mm Capture Area & Brain & Visium & 1 & 19,465 \\
Preservation Method Comparison on Visium CytAssist: Fresh Frozen Mouse Brain (Sagittal), 11 mm Capture Area & Brain & Visium & 1 & 19,465 \\
Preview Data: FFPE Human Lung Cancer with Xenium Multimodal Cell Segmentation & Lung & Xenium & 1 & 541 \\

Preview Data: FFPE Human Lymph Node with 5K Pan Tissue and Pathways Panel & Lymphoid & Xenium & 1 & 11,094\\

Visium CytAssist Gene Expression Libraries of Post-Xenium Human Colon Cancer (FFPE) & Bowel & Visium & 4 & 18,085 \\
Visium CytAssist Gene Expression Libraries of Post-Xenium Mouse Brain (FF) & Brain & Visium & 4 & 19,465 \\
Visium CytAssist, Mouse Embryo, 11 mm Capture Area (FFPE) & Embryo & Visium & 1 & 19,465 \\

Visium HD Spatial Gene Expression Library, Mouse Kidney (FFPE) & Kidney & Visium HD & 1 & 19,059\\

Visium HD Spatial Gene Expression Library, Mouse Embryo (FFPE) & Embryo & Visium HD & 1 & 19,059\\

Visium HD Spatial Gene Expression Library, Human Pancreas (FFPE) & Pancreas & Visium HD & 1 & 18,085\\

Whole Mouse Pup Preview Data (Xenium Mouse Tissue Atlassing Panel) & Whole organism & Xenium & 1 & 541 \\
\bottomrule
\end{tabular}
\end{table}

\begin{table}[ht]
\centering  
\small
\caption{\textbf{Datasets gathered from NCBI.}}  
\label{tab:ncbi}  
\begin{tabular}{m{7.5cm} m{1.3cm} m{1.5cm} >{\centering\arraybackslash}m{.2cm} >{\centering\arraybackslash}m{1cm}}
\toprule
\textbf{Publication} & \textbf{Organ} & \textbf{Technology} & $\bm{n}$ & \textbf{Num. genes} \\
\midrule
10X Visium Spatial transcriptomics of murine colon at d14 (mucosa healing) in B cell sufficient-deficient mice~\cite{parigi2022spatial} & Bowel & Visium & 2 & 31,053 \\
10X Visium Spatial transcriptomics of murine colon in steady state and during recovery after DSS colitis~\cite{parigi2022spatial} & Bowel & Visium & 2 & 31,053 \\
A Spatial Transcriptomic atlas of the human kidney papilla identifies significant immune injury in patients with stone disease~\cite{canela2023spatially} & Kidney & Visium & 7 & 36,601 \\
A cellular hierarchy in melanoma uncouples growth and metastasis~\cite{karras2022cellular} & Skin & Visium & 3 & 31,053 \\
A new epithelial cell subpopulation predicts response to surgery, chemotherapy, and immunotherapy in bladder cancer~\cite{abdelhafiz2023ychromosome,gouin2021ncadherin} & Bladder & Visium & 4 & 33,538 \\
A novel model of binge ethanol exposure reveals enhanced neurodegeneration with advanced age~\cite{anton2024binge} & Brain & Visium & 4 & 32,285 \\
A single-cell transcriptomic analysis of endometriosis~\cite{fonseca2023single} & Uterus & Visium & 2 & 36,601 \\
Distinct mesenchymal cell states mediate prostate cancer progression~\cite{pakula2024distinct} & Prostate & Visium & 2 & 32,589 \\
Epithelial Plasticity and Innate Immune Activation Promote Lung Tissue Remodeling following Respiratory Viral Infection~\cite{beppu2023epithelial} & Lung & Visium & 1 & 32,285 \\

Image-based spatial transcriptomics identifies molecular niche dysregulation associated with distal lung remodeling in pulmonary fibrosis \cite{Vannan2023.12.15.571954} & Lung & Xenium & 20 & 17,145\\

Gene expression within a human choroidal neovascular membrane using spatial transcriptomics~\cite{voigt2023gene} & Eye & Visium & 5 & 36,601 \\
Genome-wide Spatial Expression Profiling in Formalin-fixed Tissues~\cite{graciaVillacampa2021genome} & Kidney & Visium & 14 & 33,538 \\
High-resolution mapping of the tumor microenvironment using integrated single-cell, spatial, and in situ analysis~\cite{janesick2023high} & Breast & Xenium & 4 & 541 \\
Identification of TREM1+CD163+ myeloid cells as a deleterious immune subset in HCC~\cite{giraud2022trem} & Liver & Visium & 2 & 36,601 \\
Integration of spatial and single cell transcriptomics localizes epithelial-immune cross-talk in kidney injury~\cite{ferreira2021integration} & Kidney & Visium & 4 & 33,538 \\
Molecular Atlas of the Adult Mouse Brain~\cite{ortiz2020molecular} & Brain & Spatial Transcriptomics & 75 & 23,371 \\
Regional differential gene expression analyses of brains from four 24w-old Nf1+- mice & Brain & Visium & 4 & 32,285 \\
SARS-CoV-2 Niches in Human Placenta Revealed by Spatial Transcriptomics~\cite{barrozo2023sars} & Uterus & Visium & 16 & 36,612 \\
Schwann Cells Shape Tumor Cells and Cancer-Associated Fibroblasts in the Pancreatic Ductal Adenocarcinoma Microenvironment~\cite{xue2023schwann} & Pancreas & Visium & 4 & 20,615 \\
Single Cell and Spatial Analysis of Human Squamous Cell Carcinoma~\cite{ji2020multimodal} & Skin & Spatial Transcriptomics & 12 & 17,138 \\
Single-cell profiling of primary and paired metastatic lymph node tumors in breast cancer patients~\cite{liu2022single} & Lymph node & Visium & 4 & 33,931 \\
\bottomrule
\end{tabular}
\end{table}

\begin{table}[ht]
\centering  
\small
\caption{\textbf{Datasets gathered from NCBI. Continuation.}}  
\label{tab:ncbi2}  
\begin{tabular}{m{7.5cm} m{1.3cm} m{1.5cm} >{\centering\arraybackslash}m{.2cm} >{\centering\arraybackslash}m{1cm}}
\toprule
\textbf{Publication} & \textbf{Organ} & \textbf{Technology} & $\bm{n}$ & \textbf{Num. genes} \\
\midrule
Single-cell and spatial transcriptomics characterization of the immunological landscape in the healthy and PSC human liver~\cite{andrews2024single} & Liver & Visium & 4 & 36,601 \\
Single-nucleus Ribonucleic Acid-sequencing and Spatial Transcriptomics Reveal the Cardioprotection of Shexiang Baoxin Pill (MUSKARDIA) in Mice with Myocardial Ischemia-Reperfusion Injury~\cite{lin2023single} & Heart & Visium & 2 & 32,285 \\
Spatial Multimodal Analysis: MALDI-MSI and Spatial Transcriptomics within the same tissue section~\cite{vicari2023spatial} & Brain & Visium & 19 & 32,285 \\
Spatial RNA sequencing of regenerating mouse hindlimb muscle~\cite{mckellar2021large} & Muscle & Visium & 3 & 33,217 \\
Spatial Total RNA-Sequencing of regenerating mouse hindlimb muscle and Type 1-Lang reovirus-infected mouse heart~\cite{mcKellar2023spatial} & Muscle & Visium & 7 & 55,414 \\
Spatial localization with Spatial Transcriptomics for an atlas of healthy and injured cell states and niches in the human kidney~\cite{lake2023an} & Kidney & Visium & 23 & 33,538 \\
Spatial sequencing of Foreign body granuloma~\cite{krausgruber2023single} & None & Visium & 1 & 15,524 \\
Spatial transcriptomics landscape of non-communicable inflammatory skin diseases~\cite{schabitz2022spatial} & Skin & Visium & 59 & 20,613 \\
Spatial transcriptomics of adenoid cystic carcinoma of the lacrimal gland~\cite{moeyersoms2023spatial} & Eye & Visium & 1 & 17,943 \\
Spatial transcriptomics of the mouse brain across three age groups & Brain & Visium & 6 & 32,285 \\
Spatial transcriptomics reveal unnresolved wound repair as potential driver of PFA Ependymoma progression~\cite{fu2023spatial} & Brain & Visium & 14 & 36,601 \\
Spatiotemporal dynamics of molecular pathology in amyotrophic lateral sclerosis~\cite{maniatis2019spatiotemporal} & Spinal cord & Spatial Transcriptomics & 302 & 12,572 \\
The neurons that restore walking after paralysis~\cite{kathe2022neurons}  & Spinal cord & Visium & 16 & 22,127 \\
Visium spatial transcriptomics analysis of lacrimal gland during chronic inflammation progression~\cite{mauduit2022spatial} & Eye & Visium & 4 & 32,285 \\
YAP Drives Assembly of a Spatially Colocalized Cellular Triad Required for Heart Renewal~\cite{li2024yap} & Heart & Visium & 2 & 32,285 \\
Zika virus co-opts miRNA networks to persist in placental microenvironments detected by spatial transcriptomics~\cite{barrozo2024zika} & Placenta & Visium & 8 & 32,298 \\
Mouse model Heptablastoma spatial transcriptomics~\cite{pilet2023preneoplastic} & Liver & Visium & 10 & 31,053 \\
\bottomrule
\end{tabular}
\end{table}

\begin{table}[ht]
\centering  
\small
\caption{\textbf{Datasets gathered on Spatial-Research.}}  
\label{tab:spa}  
\begin{tabular}{m{7.5cm} m{1.3cm} m{1.5cm} >{\centering\arraybackslash}m{.2cm} >{\centering\arraybackslash}m{1cm}}
\toprule
\textbf{Publication} & \textbf{Organ} & \textbf{Technology} & $\bm{n}$ & \textbf{Num. genes} \\\midrule
A spatiotemporal organ-wide gene expression and cell atlas of the developing human heart~\cite{asp2019spatiotemporal} & Heart & Spatial Transcriptomics & 19 & 39,739 \\
Integrating spatial gene expression and breast tumour morphology via deep learning~\cite{he2020integrating} & Breast & Spatial Transcriptomics & 68 & 16,744 \\
Spatial deconvolution of HER2-positive breast cancer delineates tumor-associated cell type interactions~\cite{andersson2021spatial} & Breast & Spatial Transcriptomics & 36 & 15,045 \\
Visualization and analysis of gene expression in tissue sections by spatial transcriptomics~\cite{stahl2016visualization} & Brain & Spatial Transcriptomics & 16 & 16,573 \\
\bottomrule
\end{tabular}
\end{table}

\begin{table}[ht]
\centering  
\small
\caption{\textbf{Datasets gathered on Mendeley.}}  
\label{tab:mend}  
\begin{tabular}{m{7.5cm} m{1.3cm} m{1.5cm} >{\centering\arraybackslash}m{.2cm} >{\centering\arraybackslash}m{1cm}}
\toprule
\textbf{Publication} & \textbf{Organ} & \textbf{Technology} & $\bm{n}$ & \textbf{Num. genes} \\
\midrule
Ex-ST~\cite{andrusivova2023exst} & Brain & Visium & 5 & 31,053 \\
 Genome-wide spatial expression profiling in formalin-fixed tissues~\cite{villacampa2021genome} & Brain & Visium & 15 & 31,053 \\
Human ileum, Visium~\cite{mirzazadeh2021human} & Bowel & Visium & 4 & 33,538 \\
Human squamous cell carcinoma~\cite{abalo2021human} & Skin & Visium & 4 & 33,538 \\
Prostate needle biopsies pre- and post-ADT: Count matrices, histological-, and Androgen receptor immunohistochemistry images~\cite{marklund2022prostate} & Prostate & Spatial Transcriptomics & 24 & 26,437 \\
Spatially resolved clonal copy number alterations in benign and malignant tissue~\cite{erickson2022spatially} & Prostate & Visium & 23 & 33,538 \\
Spatially resolved transcriptomic profiling of degraded~\cite{mirzazadeh2023spatially} & Bowel & Visium & 35 & 17,943 \\
spatialRNAseq heart raw suppdata & Heart & Visium & 4 & 54,848 \\
spatialRNAseq ileum raw suppdata & Bowel & Visium & 4 & 54,848 \\
\bottomrule
\end{tabular}
\end{table}

\begin{table}[ht]
\centering  
\small
\caption{\textbf{Datasets gathered on Github and the Human Cell Atlas data explorer.}}  
\label{tab:misc}  
\begin{tabular}{m{7.5cm} m{1.3cm} m{1.5cm} >{\centering\arraybackslash}m{.2cm} >{\centering\arraybackslash}m{1cm}}
\toprule
\textbf{Publication} & \textbf{Organ} & \textbf{Technology} & $\bm{n}$ & \textbf{Num. genes} \\
\midrule
A spatially resolved atlas of the human lung characterizes a gland-associated immune niche\cite{madissoon2023spatially} & Lung & Visium & 20 & 17,922 \\
Molecular cartography uncovers evolutionary and microenvironmental dynamics in sporadic colorectal tumors\cite{heiser_molecular_2023} & Colon & Visium & 41 & 19,366 \\
Spatially resolved multiomics of human cardiac niches\cite{kanemaru_spatially_2023} & Heart & Visium & 41 & 33,538 \\
Transcriptome-scale spatial gene expression in the human dorsolateral prefrontal cortex\cite{maynard2021transcriptome} & Brain & Visium  & 12 & 33,538  \\
\bottomrule
\end{tabular}
\end{table}

\begin{table}[ht]
\centering  
\caption{\textbf{Internal datasets.}}  
\label{tab:internal}  
\begin{tabular}{m{7.5cm} m{1.3cm} m{1.5cm} >{\centering\arraybackslash}m{.2cm} >{\centering\arraybackslash}m{1cm}}
\toprule
\textbf{Publication} & \textbf{Organ} & \textbf{Technology} & $\bm{n}$ & \textbf{Num. genes} \\
\midrule
Prostate ST Internal & Prostate & Visium & 4 & 17,943 \\
Tertiary lymphoid structures generate and propagate anti-tumor antibody-producing plasma cells in renal cell cancer~\cite{meylan2022tertiary} & Lymph node & Visium & 24 & 17,943 \\
\bottomrule
\end{tabular}
\end{table}

\begin{table}[ht]
\centering  
\small
\caption{\textbf{Datasets gathered on Zenodo.}} 
\label{tab:zenodo} 
\begin{tabular}{m{7.5cm} m{1.3cm} m{1.5cm} >{\centering\arraybackslash}m{.2cm} >{\centering\arraybackslash}m{1cm}}
\toprule
\textbf{Publication} & \textbf{Organ} & \textbf{Technology} & $\bm{n}$ & \textbf{Num. genes} \\
\midrule
Charting the Heterogeneity of Colorectal Cancer Consensus Molecular Subtypes using Spatial Transcriptomics: datasets~\cite{valdeolivas2023charting} & Bowel & Visium & 14 & 36,601 \\
Demo 10x Visium dataset for STQ~\cite{zenodo2024demo}& Skin & Visium & 1 & 68,886 \\

Nextflow Pipeline for Visium and H\&E Data from Patient-Derived Xenograft Samples \cite{Domanskyi2023.07.27.550727} & Skin & Visium & 4 & 68886\\

Spotiphy: generative modeling in single-cell spatial whole transcriptomics~\cite{zenodo2024spotiphy} & Brain & Visium & 2 & 32,285 \\
\bottomrule
\end{tabular}
\end{table}

\subsection{HEST-Library}

\begin{figure}
   \centering
   \resizebox{\linewidth}{!}{
       \includegraphics{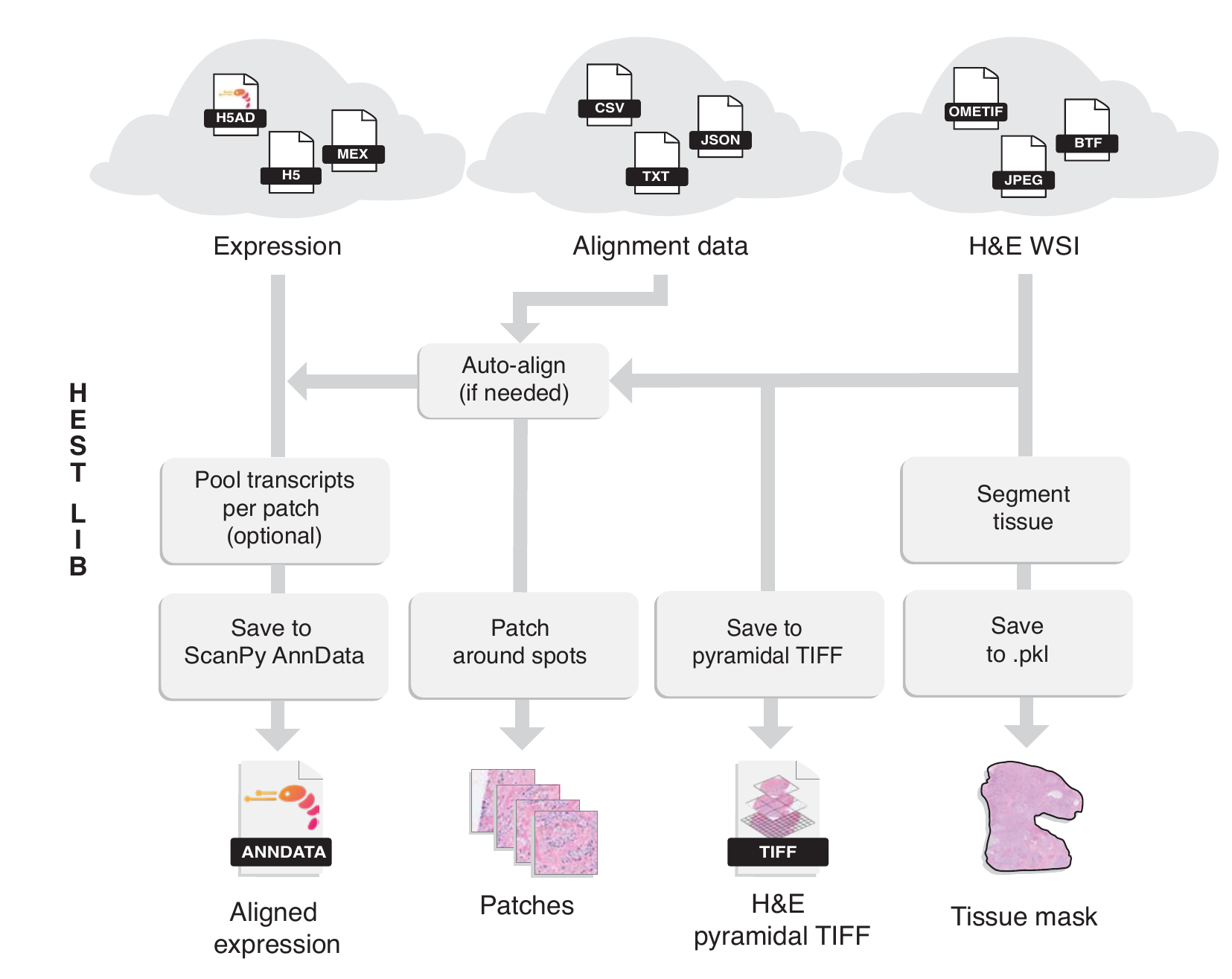}
   }
   \caption{
       \textbf{Overview of HEST-Library functionalities.} HEST was designed to transform legacy data scrapped in multiple public repositories, such as NCBI, into unified HEST objects that can easily be integrated into computational pipelines.
   }
   \label{fig:hest_library}
\end{figure}

\begin{figure}[t]
   \centering
   \resizebox{\linewidth}{!}{
    \includegraphics{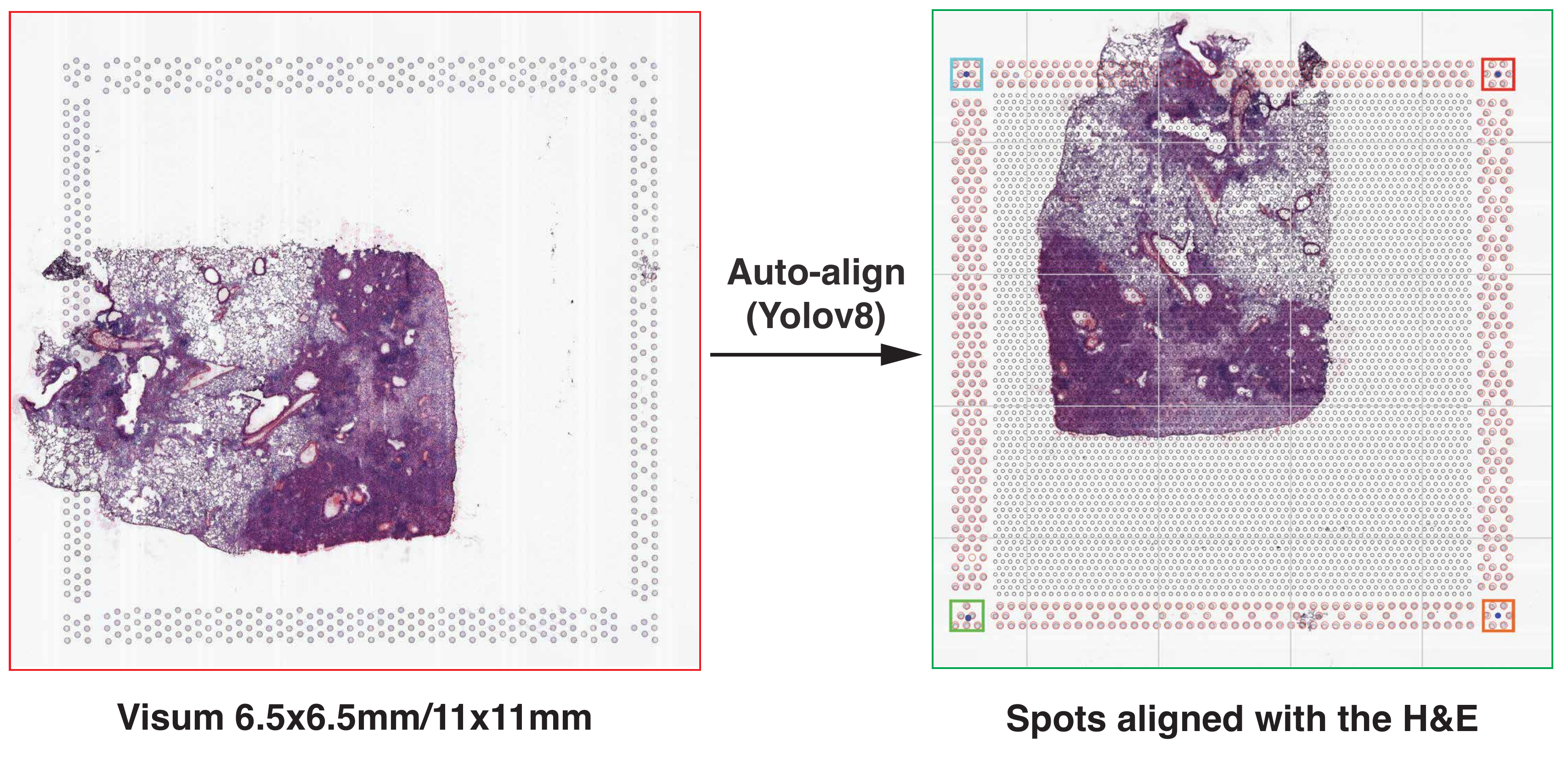}
    }
   \caption{
       \textbf{Fiducial detection and automatic alignment in Visium.} Corner fiducials on 6.5$\times$6.5mm and 11mm$\times$11mm Visium slides are automatically detected with a finetuned Yolov8 model. The spot coordinates are then derived if at least 3 of the 4 corner fiducials are detected. This process enables automatically estimating the pixel resolution.  
   }
   \label{fig:fig_autoalign}
\end{figure}

The HEST-Library helps transform unstructured spatial transcriptomics and histology data into a unified format. An overview of the HEST-Library is provided in~\autoref{fig:hest_library}. An example of fiducial detection is presented in~\autoref{fig:fig_autoalign}.

\subsection{HEST-Benchmark} \label{app:hestbenchmark}

\noindent\textbf{Gene selection, XGBoost Forest, and Ridge regression models:} We learn a regression model that maps the patch embeddings of each encoder to its corresponding gene expression profile.
The XGboost model uses 100 estimators, a 0.1 learning rate, a max depth of 3, 0.8 subsampling, gamma of 0.0, regression alpha of 0.0, and regression lambda of 1.0. Additional information can be found in the XGBoost API\footnote{\url{https://xgboost.readthedocs.io/en/stable/python/python_api.html}}. 
The Ridge regression uses a fixed $L2$ regularization coefficient $\lambda$ set to $100/MC$, where $M$ is the embedding dimension and $C=50$ is the number of targets trained with the Regularized Least-Squares Routine solver (sklearn implementation). Both regression models are trained to predict a panel constituted of the 50 most variable genes of each task. Specifically, for each task, we select the 50 most variable genes across all spots and samples after excluding the genes that have non-zero counts in less than 10\% of the spots. 

\noindent\textbf{Benchmark task description:} We provide complementary information on each task introduced as part of the HEST-Benchmark. 

\begin{table}[t]
\centering
\caption{\textbf{Overview of the HEST-Benchmark.} Each task involves predicting the expression levels of the 50 most variable genes from 112$\times$112 $\mu$m H\&E-stained image patches centered on each spatial transcriptomics spot. The tasks are formulated as multivariate regression problems. The Oncotree code describes the cancer type diagnosed in samples, e.g., PAAD denotes pancreatic adenocarcinoma. Additional information is provided in the Appendix.}
\label{tab:hestbenchmark}
\resizebox{\linewidth}{!}{
\begin{tabular}
{>{}m{3cm} m{1.5cm} m{2cm} >{\centering\arraybackslash}m{2cm} >{\centering\arraybackslash}m{2cm} m{2cm} m{2cm}}
\toprule
\textbf{Task ID} & \textbf{Oncotree} & \textbf{Organ} & \textbf{Number of Patients} & \textbf{Number of Samples} & \textbf{Technology} \\ 
\midrule
Task 1~\cite{janesick2023high} & IDC & Breast & 4 & 4 & Xenium  \\ 
Task 2 & PRAD & Prostate & 2 & 23 & Visium \\ 
Task 3 & PAAD & Pancreas & 3 & 3 & Xenium \\ 
Task 4 & SKCM & Skin & 2 & 2 & Xenium \\ 
Task 5~\cite{valdeolivas2023charting} & COAD & Colon & 2 & 4 & Xenium \\ 
Task 6~\cite{valdeolivas2023charting} & READ & Rectum & 2 & 4 & Visium \\ 
Task 7~\cite{meylan2022tertiary} & ccRCC & Kidney & 24 & 24 & Visium \\ 
Task 8 & LUAD & Lung & 2 & 2 & Xenium \\ 
Task 9~\cite{liu2022single} & IDC & Axillary lymph nodes & 4 & 4 & Visium \\ 
\bottomrule
\end{tabular}}
\end{table}

\noindent\textbf{Task 1: Prediction of expression in invasive ductal carcinoma (breast cancer, IDC).} We used all publicly available Xenium samples available on 10x Genomics (``FFPE Human Breast using the Entire Sample Area'', 2 patients) and two samples published in ~\cite{janesick2023high} (TENX95, TENX99, NCBI783, NCBI785). All samples are FFPE sections imaged with the Xenium pipeline v1. 

\noindent\textbf{Task 2: Prediction of expression in prostate adenocarcinoma (prostate cancer, PRAD).} 
We used all 23 Visium samples (fresh frozen sections) from 2 patients published in~\cite{erickson2022spatially} (MEND139 to MEND162). Both patients were diagnosed with prostatic acinar adenocarcinoma with a (4+3) Gleason score (ISUP group 4).

\noindent\textbf{Task 3: Prediction of expression in pancreatic adenocarcinoma (pancreatic cancer, PAAD).} We used 3 samples from 3 different patients from 10x Genomics (``FFPE Human Pancreas with Xenium Multimodal Cell Segmentation'' and ``Pancreatic Cancer with Xenium Human Multi-Tissue and Cancer Panel''). All samples are FFPE sections processed with Xenium pipeline v1 (TENX116, TENX126, TENX140). 

\noindent\textbf{Task 4: Expression prediction in skin cutaneous melanoma (skin cancer, SKCM).} 
We used 2 samples from 2 different patients from 10x Genomics website (``Human Skin Data with Xenium Human Multi-Tissue and
Cancer Panel''). All samples are FFPE sections processed with Xenium pipeline v1 (TENX115, TENX117). 

\noindent\textbf{Task 5: Prediction of expression in colon adenocarcinoma (colon cancer, COAD).} We used 4 COAD samples from 2 different patients available on 10x Genomics (TENX111, TENX147, TENX148, TENX149). All samples are fresh frozen sections processed with Visium.

\noindent\textbf{Task 6: Prediction of expression in rectal adenocarcinoma (rectum cancer, READ).} We used 4 READ samples from 2 different patients published in~\cite{valdeolivas2023charting}. All samples are fresh frozen sections processed with Visium (ZEN36, ZEN40, ZEN48, ZEN49).

\noindent\textbf{Task 7: Prediction of expression in clear cell renal cell carcinoma (kidney cancer, ccRCC).} We used the 24 ccRCC samples of 24 different patients published in~\cite{meylan2022tertiary}. All samples are fresh frozen sections processed with Visium (INT1 to INT24).

\noindent\textbf{Task 8: Prediction of expression in lung adenocarcinoma  (lung cancer, LUAD).} We used 2 LUAD samples from 2 different patients from 10x genomics (``Preview Data: FFPE Human Lung Cancer with Xenium Multimodal Cell Segmentation''). All samples are fresh frozen sections processed with Xenium pipeline v1 (TENX118, TENX141).

\noindent\textbf{Task 9: Prediction of expression in axillary lymph nodes in IDC patients.} We used 4 axillary lymph node samples from 2 IDC patients published in~\cite{liu2022single}. All samples are fresh frozen sections processed with Visium (NCBI681, NCBI682, NCBI683, NCBI684).

\begin{table}[t]
    \centering
    \caption{\textbf{State-of-the-art foundation models for histology evaluated on HEST-Benchmark.} B: Base, L: Large, H: Huge, G: Giant. $^{*}$: number of patches during pretraining, $^{**}$: number of patches during fine-tuning.}
    \label{tab:vision_encoders}
    \resizebox{\linewidth}{!}{
    \begin{tabular}{lcccccc}
        \toprule
        \textbf{Name} & \textbf{Number of} & \textbf{Number of } & \textbf{Magnification} & \textbf{Model} & \textbf{Training} & \textbf{Number of}  \\ 
        & \textbf{slides} & \textbf{patches} & & & \textbf{recipe} & \textbf{parameters} \\ 
        \midrule
        ResNet50 (IN)~\cite{lu2021data} & N/A & 1.2M & N/A & ResNet-50 & Supervised & 23M \\
        CTransPath~\cite{wang2021transpath} & 32k & 17M & 10$\times$ & Swin-T & MoCov3 & 28M \\
        Remedis~\cite{azizi2023robust}  & 10k & 10M & 20$\times$ & ResNet-152 & iBOT & 232M \\
        Phikon~\cite{filiot2023scaling} & 6k & 43.3M & 20$\times$ & ViT-B & iBOT & 86M \\
        UNI~\cite{chen2024towards} & 100k & 100M & 20$\times$ & ViT-L & DINOv2 & 307M \\
        CONCH~\cite{lu2024towards} & N/A & 16M$^{*}$ + 1.17M$^{**}$ & Many & ViT-B & CoCa &  86M \\
        GigaPath~\cite{xu2024wholeslide} & 171k & 1.3B & 20$\times$ & ViT-g & DINOv2 & 1.13B \\
        Virchow~\cite{vorontsov2024foundation} & 1.5M  & 2B & 20$\times$ & ViT-H & DINOv2 & 632M \\
        Virchow 2~\cite{Zimmermann2024Virchow2SS} & 3.1M & 1.9B & Many & ViT-H & DINOv2 & 632M \\
        H-Optimus-0 & 500K & 273M & 20x & ViT-g & DINOv2 & 1.13B \\
        UNIv1.5 & 350K & 432M & 20x & ViT-g & DINOv2 & 1.13B \\
        \bottomrule
    \end{tabular}}
\end{table}

We provide a brief description of each patch encoder assessed with the HEST-Benchmark. 

\noindent\textbf{ResNet50 (IN)~\cite{lu2020data}:} This model uses a ResNet50 backbone~\cite{he2016deep} trained on ImageNet~\cite{deng2009imagenet} (1.2 million natural images). Following prior work~\cite{lu2020data}, the patch embeddings are extracted by taking the representation at the penultimate layer before final classification. 

\noindent\textbf{CTransPath~\cite{wang2022transformer}:} This model uses a ``Tiny'' Swin Transformer backbone~\cite{liu2021swin} with a window size of 14 (Swin-T/14, 28 million parameters) pretrained on TCGA and PAIP datasets (17 million images) using MoCoV3~\cite{chen2021an}. 

\noindent\textbf{Remedis~\cite{azizi2023robust}:} This model uses a ResNet-152$\times$2 (232 million parameters) initialized with the ``Big Transfer''-medium protocol~\cite{kolesnikov2020big} on ImageNet-22K and pretrained with SimCLR~\cite{chen2020simple} on TCGA. 

\noindent\textbf{Phikon~\cite{filiot2023scaling}:} This model uses a Vision Transformer-Base (ViT-B, 86 million parameters)~\cite{dosovitskiy2021image} trained on TCGA data using iBOT~\cite{zhou2021ibot}.  

\noindent\textbf{UNI~\cite{chen2024towards}:} This model uses a ViT-Large (ViT-L, 307 million parameters)~\cite{dosovitskiy2021image} trained on 100 million histology images (over 100,000 slides) from proprietary and public data using DINOv2~\cite{oquab2023dinov2}. 

\noindent\textbf{CONCH~\cite{lu2024towards}:} This model uses a ViT-B (86 million parameters) trained on a smaller version of UNI using iBOT, and then fine-tuned on 1.17 million histology image--caption pairs extracted from online educational and research resources using CoCa~\cite{yu2022coca}. 

\noindent\textbf{GigaPath~\cite{xu2024wholeslide}:} This model uses a ViT-giant (1.13 billion parameters) trained on 1.3 billion image patches from 171,189 WSIs at 20$\times$ magnification using DINOv2. 

\noindent\textbf{Virchow~\cite{vorontsov2024foundation}:} This model uses a ViT-Huge (632M parameters) trained on 2 billion image patches and 1.5M WSIs at 20$\times$ magnification using DINOv2.

\noindent\textbf{Virchow 2~\cite{Zimmermann2024Virchow2SS}:} This model uses a ViT-Huge (632M parameters) trained on 1.9B patches and 3.1M WSIs using DINOv2

\noindent\textbf{H-Optimus-0:} This model uses a ViT-giant (1.13B parameters) trained 273 million image patches from 500,000 WSIs at 20$\times$ magnification using DINOv2.

\noindent\textbf{UNIv1.5:} This model uses a ViT-giant (1.13B parameters) trained on 432 million image patches from 350,000 WSIs using DINOv2. 

\section{HEST for multimodal representation learning}\label{sec:appendix_multimodal}

We provide additional information regarding CONCH fine-tuning using multimodal alignment. CONCH-FT model, a ViT-Base model initialized with CONCH weights, was fine-tuned for 50 epochs using a cosine learning rate scheduler, with a base learning rate of $10^{-4}$ for the image encoder and $10^{-3}$ for the expression encoder. Only the last 3 layers of the model were fine-tuned, with a layer-wise learning decay rate of 0.7. For training with the infoNCE loss, a contrastive temperature of $10^{-2}$ and batch size of 1,024 pairs of patch and transcriptomics were used. A combination of random horizontal/vertical flip and color jittering was employed for image patch augmentation.

The rank of the embedding space (also referred to as smooth rank measure~\cite{garrido2023rankme}) measures the quality of the embeddings produced from encoders trained in unsupervised or self-supervised manners. Given the patch embedding matrix $H \in \mathcal{R}^{N \times d}$ and $d<N$, where $N$ is the number of patches and $d$ is the feature dimension, we compute the rank as the entropy of the $d$ L1-normalized singular values of $H$.

\section{HEST for discovery}

Cells were segmented and classified using CellViT~\cite{horst2024cellvit}. To find the gene expression profile of each neoplastic cell, we matched each cell to its corresponding cell index in Xenium by assigning the index for which the distance between the cell centroids was the smallest. After matching all neoplastic cells, only those cells for which the assignment was unique were kept. After this filtering step, an average of 91\% of the cells per sample were kept while 9\% of the cells were discarded.

\begin{figure}[ht]
   \centering
   \resizebox{\linewidth}{!}{
       \includegraphics{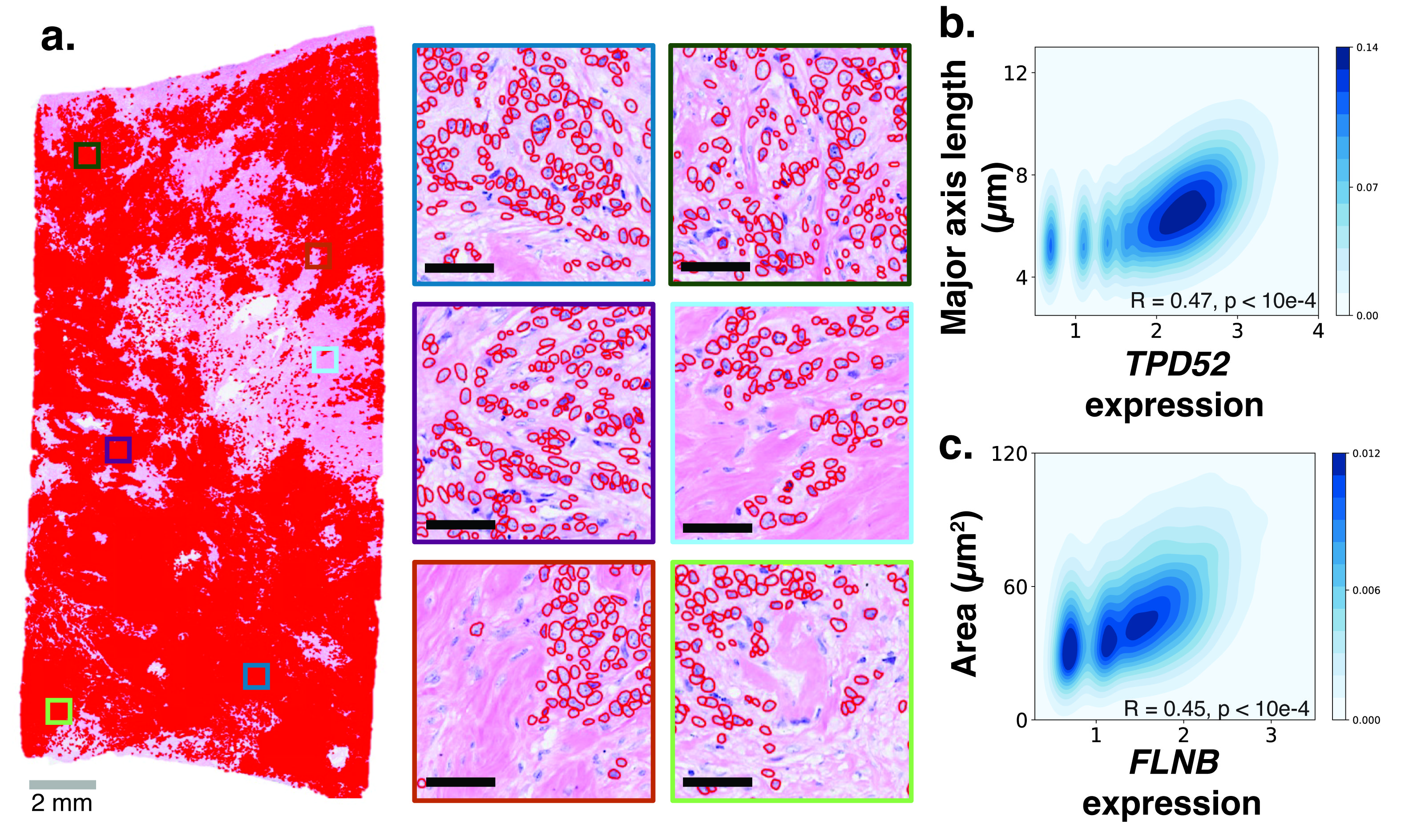}
   }
   \caption{
       \textbf{HEST for biomarker discovery: Analysis of an invasive ductal carcinoma Xenium sample.}
       \textbf{a.} IDC Xenium sample with neoplastic nuclei overlaid in red ($n_c$=342,018 detected nuclei). Six randomly selected regions with CellViT segmentation of the neoplastic nuclei. Black scale bar represents 30 $\mu$m. 
       \textbf{b.} Pearson correlation between the major axis length of neoplastic nuclei and the log1p-normalized expression of \textit{TPD52}. 
       \textbf{c.} Analogous analysis between nuclear area and \textit{FLNB} expression. 
   }
   \label{fig:discovery_extended}
\end{figure}

\section{Datasheet for HEST-1k}

We provide a DataSheet for HEST-1k that summarizes the contributions, analyses, and intended usages presented in the study. 

\subsection{Motivation for dataset creation}

\begin{itemize}
    \item \textbf{Why was the dataset created?}
    HEST-1k was designed with three key applications: (1) multimodal representation learning of histology and transcriptomics, (2) biomarker exploration and characterization, and (3) benchmarking foundation models for pathology. Despite many publicly available resources, no existing unified and user-friendly formatting was available to bring ST into the world of deep learning.  
    \item \textbf{What (other) tasks could the dataset be used for? Are there obvious tasks for which it should not be used?}
    Users are welcome to introduce new, creative ways to use the dataset. However, users are not allowed to try to retrieve patient information from the existing data. A dedicated section is provided to discuss ethical considerations and intended usage. 
    \item \textbf{Has the dataset been used for any tasks already? If so, where are the results so others can compare (e.g., links to published papers)?}
    The metadata attached to HEST-1k reports all samples that were made public as part of a publication (peer-reviewed or not). 
    \item \textbf{Who funded the creation of the dataset?}
    HEST is supported by the Brigham and Women's Hospital (BWH) President's Fund, Mass General Hospital (MGH) Pathology, and the National Institute of Health (NIH) National Institute of General Medical Sciences (NIGMS) through R35GM138216.
\end{itemize}

\subsection{Dataset composition}

\begin{itemize}
    \item \textbf{What are the instances?}
    The modalities used in this study are histopathology whole-slide images, gene expression data, and derivatives of these two modalities, such as nuclear segmentation and classification maps.
    \item \textbf{Are relationships between instances made explicit in the data}
    Each whole-slide image maps to a unique gene expression profile in an unequivocal way. 
    \item \textbf{What data does each instance consist of?}
    Imaging data consists of Generic TIFF objects stored in a pyramidal format, and gene expression data consists of \textit{scanpy} objects. Derivatives are stored in JSON files, parquet files, and Hierarchical Data Format (HDF) files. 
    \item \textbf{Is there a label/target associated with instances? If the instances are related to people, are subpopulations identified (e.g., by age, gender, etc.), and what is their distribution?}
    Each sample pair (slide and expression profile) is associated with comprehensive metadata. All metadata information is thoroughly described in the main paper. Age and gender are only reported in a subset of cases.  
    \item \textbf{Is everything included or does the data rely on external resources? (e.g., websites, tweets, datasets) If external resources, a) are there guarantees that they will exist, and remain constant, over time; b) is there an official archival version. Are there licenses, fees or rights associated with any of the data?}
    We provide all data as part of the HEST-1k release. In addition, a link to the original data is provided in the metadata. Each sample is associated with a license as provided by the original publication, where we ensured that the reported license allowed for distributing and creating derivatives of the data.
    \item \textbf{Are there recommended data splits or evaluation measures?}
    HEST-1k comes with the HEST-Benchmark, a series of tasks for gene expression prediction from histology images. All patient-stratified splits are specified in the attached comma-separated values (CSV) files. 
    \item \textbf{What experiments were initially run on this dataset? Have a summary of those results and, if available, provide the link to a paper with more information here.}
    All experiments run with HEST-1k are described in this study. The reader can refer to the main text for a thorough description of all experiments (see \textbf{HEST-Benchmark}, \textbf{HEST for biomarker exploration}, \textbf{HEST for multimodal representation learning}).
\end{itemize}

\subsection{Data collection process}

\begin{itemize}
    \item \textbf{How was the data collected?}
    The data were manually inspected and curated by the authors of the present study. 
    \item \textbf{Who was involved in the data collection process?}
    All authors of the present study were involved in the data collection, inspection, and curation. The reader can refer to the original publication to understand how the data were originally acquired.  
    \item \textbf{Over what time frame was the data collected? Does the collection time frame match the creation time frame?}
    The original data comprise publications from 2016 to 2024. As the dataset grows, more recent publications might be included in HEST-1k. 
    \item \textbf{Does the dataset contain all possible instances? Or is it, for instance, a sample (not necessarily random) from a larger set of instances?}
    All pairs of gene expression data and whole-slide images of the underlying studies were included and are unique.
    \item \textbf{Is there information missing from the dataset and why? (this does not include intentionally dropped instances; it might include, e.g., redacted text, and withheld documents) Is this data missing because it was unavailable?}
    Original publications may include some missing information, such as the alignment file between the slide and the expression profile. We developed computational tools to minimize missing information and reach near-complete metadata. 
    \item \textbf{Are there any known errors, sources of noise, or redundancies in the data?}
    All whole-slide images have been manually inspected. The quality from one sample to another varies significantly, for instance, due to poor staining, compression artifact, lower resolution, etc. Gene expression data are inherently noisy. Users can decide to apply post-hoc normalization methods to reduce noise, e.g., stain normalization on the imaging side or batch effect mitigation on the transcriptomics side.  
\end{itemize}

\subsection{Data preprocessing}

\begin{itemize}
    \item \textbf{What preprocessing/cleaning was done?}
    All whole-slide images were converted into pyramidal TIFF objects with re-estimated pixel resolution. All alignment files have been manually inspected and included if missing. All gene expression data have been transformed into \textit{scanpy} objects following the same process. 
    \item \textbf{Was the ``raw'' data saved in addition to the preprocessed/cleaned data? (e.g., to support unanticipated future uses)}
    Raw data are downloaded but not publicly shared. In the case of public samples, users can re-download them using the metadata provided as part of the dataset release. 
    \item \textbf{Is the preprocessing software available?}
    Yes, the source code to preprocess HEST-1k is made publicly available as part of the HEST library. 
\end{itemize}

\subsection{Dataset distribution}

\begin{itemize}
    \item \textbf{How is the dataset distributed?}
    HEST-1k is distributed using HuggingFace Datasets. 
    \item \textbf{When will the dataset be released/first distributed?}
    The dataset is public and can be accessed through the HuggingFace Datasets interface. 
    \item \textbf{What license (if any) is it distributed under? Are there any copyrights on the data?}
    The dataset is distributed under the Attribution-NonCommercial-ShareAlike 4.0 International license (CC BY-NC-SA 4.0 Deed). 
    \item \textbf{Are there any fees or access/export restrictions?}
    No access/export restrictions unless they violate the terms of the above-mentioned license (CC BY-NC-SA 4.0 Deed).
    \item \textbf{Who is supporting/hosting/maintaining the dataset? How does one contact the owner/curator/manager of the dataset?}
    The dataset is maintained by the authors of the publication.
    \item \textbf{Will the dataset be updated? How often and by whom? How will updates/revisions be documented and communicated (e.g., mailing list, GitHub)? Is there an erratum?}
    The dataset might evolve as additional samples become publicly available. Dataset versioning will be put in place. 
    \item \textbf{If the dataset becomes obsolete how will this be communicated?}
    The GitHub README will be updated. 
    \item \textbf{Is there a repository to link to any/all papers/systems that use this dataset?}
    There is no repository to link papers that use HEST-1k. Users are required to cite HEST-1k if they use it in their own research. 
    \item \textbf{If others want to extend/augment/build on this dataset, is there a mechanism for them to do so? If so, is there a process for tracking/assessing the quality of those contributions. What is the process for communicating/distributing these contributions to users?}
    Users are welcome to contact us if they would like to provide additional data that meets our standards. We do not have a dedicated system to communicate these contributions. Newly added data will be tracked in the versioning. 
\end{itemize}

\subsection{Legal and ethical considerations}

\begin{itemize}
    \item \textbf{If the dataset relates to people (e.g., their attributes) or was generated by people, were they informed about the data collection? (e.g., datasets that collect writing, photos, interactions, transactions, etc.)}
    HEST-1k does not include patient information (such as name, address, etc.).
    \item \textbf{If it relates to other ethically protected subjects, have appropriate obligations been met? (e.g., medical data might include information collected from animals)}
    For animal samples (\textit{Mus musculus} tissue), we refer to the original publication for an in-depth analysis.  
    \item \textbf{If it relates to people, were there any ethical review applications/reviews/approvals? (e.g. Institutional Review Board applications)}
    For human tissue, we refer to the original publication for an in-depth analysis. Internal cohorts were ethically reviewed and collected as part of dedicated IRBs. 
    \item \textbf{If it relates to people, could this dataset expose people to harm or legal action? (e.g., financial, social or otherwise) What was done to mitigate or reduce the potential for harm?}
    No, patients cannot be linked to the corresponding histology and gene expression profile. 
    \item \textbf{If it relates to people, does it unfairly advantage or disadvantage a particular social group? In what ways? How was this mitigated?}
    Most datasets do not include specific demographics. When reported, we include this information in the metadata associated with each sample. To our knowledge, the representation of HEST-1k does not unfairly advantage or disadvantage a particular social group. 
    \item \textbf{Does the dataset contain information that might be considered sensitive or confidential? (e.g., personally identifying information)}
    No. 
    \item \textbf{Does the dataset contain information that might be considered inappropriate or offensive?}
    No. 
\end{itemize}

\section{Author statement}

The authors of this paper bear all responsibility in case of violation of rights associated with HEST-1k, HEST-Library, and HEST-Benchmark.

\begin{table}[h!]
\centering
\caption{\textbf{HEST-Benchmark evaluated using Ridge regression}. Model performance measured with Pearson correlation. Best is \textbf{bold}, second best is \underline{underlined}.}
\scalebox{0.8}{
\begin{tabular}{lccccccccc|c}
\toprule
{} &               \textbf{IDC} &              \textbf{PRAD} &              \textbf{PAAD} &              \textbf{SKCM} &              \textbf{COAD} &              \textbf{READ} &             \textbf{ccRCC} &              \textbf{LUAD} &         \textbf{LYMPH IDC} &           \textbf{Average} \\
\midrule

\textbf{REMEDIS     } &                     0.4936 &                     0.2632 &                     0.2881 &                     0.4117 &                      0.151 &                     0.0776 &                     0.2201 &                     0.3114 &                     0.1694 &                     0.2651 \\
            &  \scriptsize{$\pm$ 0.0725} &  \scriptsize{$\pm$ 0.0821} &  \scriptsize{$\pm$ 0.0544} &  \scriptsize{$\pm$ 0.0384} &  \scriptsize{$\pm$ 0.0147} &  \scriptsize{$\pm$ 0.0684} &  \scriptsize{$\pm$ 0.0418} &  \scriptsize{$\pm$ 0.0432} &  \scriptsize{$\pm$ 0.0365} &   \\

\textbf{GigaPath   } &                      0.532 &         0.3035 &                     0.3172 &                     0.2231 &                      0.163 &                     0.1236 &                     0.2172 &                     0.3144 &                     0.1925 &                     0.2652 \\
            &  \scriptsize{$\pm$ 0.0812} &  \scriptsize{$\pm$ 0.0279} &  \scriptsize{$\pm$ 0.0165} &  \scriptsize{$\pm$ 0.0071} &   \scriptsize{$\pm$ 0.041} &  \scriptsize{$\pm$ 0.0379} &  \scriptsize{$\pm$ 0.0479} &  \scriptsize{$\pm$ 0.0871} &  \scriptsize{$\pm$ 0.0304} &   \\

\textbf{UNIv1.5} & 0.5657 & \underline{0.3065} & 0.3004 & 0.258 & 0.1982 & 0.1077 &                     0.2023 & 0.3084 & 0.1998 & 0.2719 \\
            &  \scriptsize{$\pm$ 0.0866} &  \scriptsize{$\pm$ 0.02} &  \scriptsize{$\pm$ 0.0199} &  \scriptsize{$\pm$ 0.0514} &  \scriptsize{$\pm$ 0.0262} &  \scriptsize{$\pm$ 0.0222} &  \scriptsize{$\pm$ 0.0595} &   \scriptsize{$\pm$ 0.0747} &  \scriptsize{$\pm$ 0.0129} &   \\

\textbf{H-Optimus-0} &            \textbf{0.5789} &                     0.2561 &                     0.3367 &                     0.2778 &                     0.1605 &                     0.1228 &                     0.2342 &                     0.3143 &                     0.1976 &                     0.2754 \\
            &  \scriptsize{$\pm$ 0.0899} &  \scriptsize{$\pm$ 0.0003} &  \scriptsize{$\pm$ 0.0428} &  \scriptsize{$\pm$ 0.0048} &  \scriptsize{$\pm$ 0.0522} &  \scriptsize{$\pm$ 0.0309} &  \scriptsize{$\pm$ 0.0373} &   \scriptsize{$\pm$ 0.083} &  \scriptsize{$\pm$ 0.0253} &   \\
            
\textbf{Virchow2        } &         \underline{0.5666} &                     0.2972 &                     0.2718 &                      0.303 &                     0.1814 &                     0.1208 &                     0.2257 &                     0.3017 &                     0.2172 &                     0.2762 \\
            &  \scriptsize{$\pm$ 0.0848} &   \scriptsize{$\pm$ 0.037} &  \scriptsize{$\pm$ 0.0387} &  \scriptsize{$\pm$ 0.0184} &  \scriptsize{$\pm$ 0.0326} &  \scriptsize{$\pm$ 0.0526} &  \scriptsize{$\pm$ 0.0433} &  \scriptsize{$\pm$ 0.1199} &   \scriptsize{$\pm$ 0.019} &  \\

\textbf{Virchow    } &                     0.5583 &                     0.2744 &                     0.3361 &                     0.3389 &                     0.1825 &                     0.0955 &                     0.2375 &                     0.2897 &                     0.2081 &                     0.2801 \\
            &  \scriptsize{$\pm$ 0.0876} &  \scriptsize{$\pm$ 0.0019} &   \scriptsize{$\pm$ 0.037} &  \scriptsize{$\pm$ 0.0063} &  \scriptsize{$\pm$ 0.0369} &  \scriptsize{$\pm$ 0.0527} &  \scriptsize{$\pm$ 0.0371} &  \scriptsize{$\pm$ 0.0785} &  \scriptsize{$\pm$ 0.0234} & \\
            
\textbf{ResNet50   } &                     0.4453 &                     0.2753 &                     0.3432 &          \underline{0.413} &         \underline{0.2009} &                     0.0669 &                     0.2103 &                     0.4001 &                      0.203 &                     0.2842 \\
            &  \scriptsize{$\pm$ 0.0377} &  \scriptsize{$\pm$ 0.0622} &  \scriptsize{$\pm$ 0.0654} &  \scriptsize{$\pm$ 0.0814} &   \scriptsize{$\pm$ 0.061} &  \scriptsize{$\pm$ 0.0646} &  \scriptsize{$\pm$ 0.0548} &  \scriptsize{$\pm$ 0.0637} &  \scriptsize{$\pm$ 0.0536} &  \\

\textbf{CTransPath } &                     0.4996 &                     0.2895 &         \underline{0.3826} &                     0.4038 &                     0.1751 &                     0.0909 &                     0.2139 &                     0.4026 &                     0.2089 &                     0.2963 \\
            &  \scriptsize{$\pm$ 0.0594} &  \scriptsize{$\pm$ 0.0724} &   \scriptsize{$\pm$ 0.066} &   \scriptsize{$\pm$ 0.065} &  \scriptsize{$\pm$ 0.0423} &  \scriptsize{$\pm$ 0.0808} &  \scriptsize{$\pm$ 0.0438} &    \scriptsize{$\pm$ 0.07} &  \scriptsize{$\pm$ 0.0367} &  \\

\textbf{Phikon    } &                     0.5259 &                     0.2493 &                     0.3594 &                     0.3684 &                     0.1697 &                     0.1136 &             \textbf{0.253} &         \underline{0.4224} &                     0.2151 &                     0.2974 \\
            &  \scriptsize{$\pm$ 0.0791} &  \scriptsize{$\pm$ 0.1264} &  \scriptsize{$\pm$ 0.0707} &  \scriptsize{$\pm$ 0.1061} &  \scriptsize{$\pm$ 0.0562} &  \scriptsize{$\pm$ 0.0749} &  \scriptsize{$\pm$ 0.0483} &  \scriptsize{$\pm$ 0.0579} &  \scriptsize{$\pm$ 0.0416} &  \\

\textbf{UNI   } &                      0.563 &                      0.257 &                     0.3768 &                     0.3433 &                     0.1839 &         \underline{0.1239} &         \underline{0.2395} &                     0.3714 &         \underline{0.2236} &         \underline{0.2981} \\
            &  \scriptsize{$\pm$ 0.0771} &  \scriptsize{$\pm$ 0.0819} &  \scriptsize{$\pm$ 0.0555} &  \scriptsize{$\pm$ 0.0556} &  \scriptsize{$\pm$ 0.0509} &  \scriptsize{$\pm$ 0.0434} &  \scriptsize{$\pm$ 0.0557} &  \scriptsize{$\pm$ 0.1098} &  \scriptsize{$\pm$ 0.0289} &   \\

\textbf{CONCH      } &                      0.528 &            \textbf{0.3604} &            \textbf{0.4224} &            \textbf{0.5079} &            \textbf{0.2467} &            \textbf{0.1443} &                     0.2356 &            \textbf{0.4957} &            \textbf{0.2462} &            \textbf{0.3541} \\
            &  \scriptsize{$\pm$ 0.0794} &  \scriptsize{$\pm$ 0.0135} &  \scriptsize{$\pm$ 0.0773} &  \scriptsize{$\pm$ 0.0281} &  \scriptsize{$\pm$ 0.0045} &  \scriptsize{$\pm$ 0.0455} &  \scriptsize{$\pm$ 0.0387} &  \scriptsize{$\pm$ 0.0203} &  \scriptsize{$\pm$ 0.0349} &   \\
            
\bottomrule
\end{tabular}
}
\label{tab:ridge_hestbench}
\end{table}

\begin{table}[h!]
\centering
\caption{\textbf{HEST-Benchmark evaluated using XGBoost regression.} Model performance measured with Pearson correlation. Best is \textbf{bold}, second best is \underline{underlined}.}
\scalebox{0.8}{
\begin{tabular}{lccccccccc|c}
\toprule
{} &               \textbf{IDC} &              \textbf{PRAD} &              \textbf{PAAD} &              \textbf{SKCM} &              \textbf{COAD} &              \textbf{READ} &             \textbf{ccRCC} &              \textbf{LUAD} &         \textbf{LYMPH IDC} &           \textbf{Average} \\
\midrule

\textbf{ResNet50 (IN)  } &                     0.4646 &                     0.3433 &                     0.4017 &                     0.4707 &                     0.2892 &                     0.0586 &                      0.181 &                     0.4967 &                     0.2284 &                      0.326 \\
            &  \scriptsize{$\pm$ 0.0353} &  \scriptsize{$\pm$ 0.0168} &  \scriptsize{$\pm$ 0.0648} &  \scriptsize{$\pm$ 0.0834} &  \scriptsize{$\pm$ 0.0115} &   \scriptsize{$\pm$ 0.069} &  \scriptsize{$\pm$ 0.0502} &    \scriptsize{$\pm$ 0.01} &  \scriptsize{$\pm$ 0.0511} &  \scriptsize{} \\

\textbf{CTransPath } &                     0.4738 &                     0.3514 &                     0.4257 &                     0.5304 &                     0.2921 &                     0.0996 &                     0.2026 &                     0.5225 &                      0.234 &                      0.348 \\
            &  \scriptsize{$\pm$ 0.0394} &  \scriptsize{$\pm$ 0.0032} &  \scriptsize{$\pm$ 0.0701} &   \scriptsize{$\pm$ 0.073} &  \scriptsize{$\pm$ 0.0018} &  \scriptsize{$\pm$ 0.0766} &  \scriptsize{$\pm$ 0.0387} &  \scriptsize{$\pm$ 0.0063} &  \scriptsize{$\pm$ 0.0613} &  \scriptsize{} \\
            
\textbf{Phikon     } &                     0.4704 &            \textbf{0.3943} &                     0.3988 &                     0.5323 &                      0.277 &                     0.1451 &                      0.213 &                      0.542 &                     0.2443 &                     0.3575 \\
            &  \scriptsize{$\pm$ 0.0672} &  \scriptsize{$\pm$ 0.0123} &  \scriptsize{$\pm$ 0.0598} &  \scriptsize{$\pm$ 0.0607} &  \scriptsize{$\pm$ 0.0098} &  \scriptsize{$\pm$ 0.0851} &  \scriptsize{$\pm$ 0.0362} &   \scriptsize{$\pm$ 0.017} &  \scriptsize{$\pm$ 0.0632} &  \scriptsize{} \\

\textbf{GigaPath   } &                     0.5222 &                     0.3749 &                     0.4415 &                     0.5297 &                     0.2876 &                     0.1609 &                     0.2207 &                     0.5506 &                     0.2464 &                     0.3705 \\
            &  \scriptsize{$\pm$ 0.0641} &  \scriptsize{$\pm$ 0.0103} &   \scriptsize{$\pm$ 0.058} &  \scriptsize{$\pm$ 0.0376} &  \scriptsize{$\pm$ 0.0039} &  \scriptsize{$\pm$ 0.0777} &  \scriptsize{$\pm$ 0.0402} &  \scriptsize{$\pm$ 0.0108} &  \scriptsize{$\pm$ 0.0526} &  \scriptsize{} \\

\textbf{CONCH      } &                     0.5175 &                     0.3784 &                     0.4428 &                     0.5766 &         0.3215 &                     0.1431 &                     0.1738 &                     0.5581 &                     0.2554 &                     0.3742 \\
            &  \scriptsize{$\pm$ 0.0602} &  \scriptsize{$\pm$ 0.0124} &  \scriptsize{$\pm$ 0.0657} &  \scriptsize{$\pm$ 0.0519} &  \scriptsize{$\pm$ 0.0062} &  \scriptsize{$\pm$ 0.0665} &  \scriptsize{$\pm$ 0.0544} &  \scriptsize{$\pm$ 0.0081} &  \scriptsize{$\pm$ 0.0605} &  \scriptsize{} \\

\textbf{REMEDIS    } &                     0.5116 &                     0.3526 &            \textbf{0.4621} &                     0.5885 &                      0.319 &                     0.1129 &                     0.2303 &                      0.562 &                     0.2521 &                     0.3768 \\
            &  \scriptsize{$\pm$ 0.0594} &  \scriptsize{$\pm$ 0.0073} &  \scriptsize{$\pm$ 0.0555} &  \scriptsize{$\pm$ 0.0253} &  \scriptsize{$\pm$ 0.0101} &  \scriptsize{$\pm$ 0.0846} &  \scriptsize{$\pm$ 0.0393} &  \scriptsize{$\pm$ 0.0057} &  \scriptsize{$\pm$ 0.0601} &  \scriptsize{} \\

\textbf{Virchow2   } &                     0.5378 &                     0.3772 &                     0.4237 &                     0.5565 &                      0.281 &            \textbf{0.1779} &            \textbf{0.2428} &                     0.5641 &         \underline{0.2582} &                     0.3799 \\
            &  \scriptsize{$\pm$ 0.0685} &   \scriptsize{$\pm$ 0.007} &  \scriptsize{$\pm$ 0.0525} &  \scriptsize{$\pm$ 0.0152} &  \scriptsize{$\pm$ 0.0162} &   \scriptsize{$\pm$ 0.077} &  \scriptsize{$\pm$ 0.0361} &  \scriptsize{$\pm$ 0.0069} &  \scriptsize{$\pm$ 0.0504} &  \scriptsize{} \\

\textbf{UNI        } &          0.538 &                     0.3513 &          \underline{0.451} &         \underline{0.6089} &                     0.2921 &                     0.1679 &                      0.235 &                     0.5357 &                     0.2456 &                     0.3806 \\
            &  \scriptsize{$\pm$ 0.0603} &  \scriptsize{$\pm$ 0.0162} &  \scriptsize{$\pm$ 0.0587} &  \scriptsize{$\pm$ 0.0294} &  \scriptsize{$\pm$ 0.0191} &  \scriptsize{$\pm$ 0.0641} &  \scriptsize{$\pm$ 0.0381} &  \scriptsize{$\pm$ 0.0057} &    \scriptsize{$\pm$ 0.05} &   \scriptsize{} \\

\textbf{Virchow    } &                     0.5309 &                     0.3447 &                     0.4448 &         \underline{0.6089} &            \underline{0.3275} &                     0.1419 &                     0.2307 &         \underline{0.5643} &            \textbf{0.2617} &         0.3839 \\
            &  \scriptsize{$\pm$ 0.0764} &  \scriptsize{$\pm$ 0.0117} &  \scriptsize{$\pm$ 0.0501} &  \scriptsize{$\pm$ 0.0165} &  \scriptsize{$\pm$ 0.0254} &  \scriptsize{$\pm$ 0.0669} &  \scriptsize{$\pm$ 0.0336} &  \scriptsize{$\pm$ 0.0091} &  \scriptsize{$\pm$ 0.0537} &  \scriptsize{} \\

\textbf{UNIv1.5} & \underline{0.555} & 0.3654 & 0.434 & 0.6025 & \textbf{0.336} & \underline{0.1742} & 0.2166 & 0.5634 & 0.2515 & \underline{0.3887} \\
            &  \scriptsize{$\pm$ 0.0763} &  \scriptsize{$\pm$ 0.0098} &  \scriptsize{$\pm$ 0.0568} &  \scriptsize{$\pm$ 0.0385} &  \scriptsize{$\pm$ 0.0179} &  \scriptsize{$\pm$ 0.0568} &  \scriptsize{$\pm$ 0.0337} &  \scriptsize{$\pm$ 0.0054} &  \scriptsize{$\pm$ 0.0434} &   \scriptsize{} \\

\textbf{H-Optimus-0} &            \textbf{0.5564} &         \underline{0.3829} &                     0.4445 &            \textbf{0.6502} &                     0.2922 &         0.1731 &         \underline{0.2402} &            \textbf{0.5654} &                     0.2555 &            \textbf{0.3956} \\
            &  \scriptsize{$\pm$ 0.0777} &  \scriptsize{$\pm$ 0.0049} &  \scriptsize{$\pm$ 0.0563} &  \scriptsize{$\pm$ 0.0326} &  \scriptsize{$\pm$ 0.0063} &  \scriptsize{$\pm$ 0.0777} &  \scriptsize{$\pm$ 0.0348} &  \scriptsize{$\pm$ 0.0084} &  \scriptsize{$\pm$ 0.0522} &   \scriptsize{} \\

\bottomrule
\end{tabular}
}
\label{tab:xgboost_hestbench}
\end{table}

\clearpage
\printbibliography

@article{xu2021predicting,
  title={Predicting axillary lymph node metastasis in early breast cancer using deep learning on primary tumor biopsy slides},
  author={Xu, Feng and Zhu, Chuang and Tang, Wenqi and Wang, Ying and Zhang, Yu and Li, Jie and Jiang, Hongchuan and Shi, Zhongyue and Liu, Jun and Jin, Mulan},
  journal={Frontiers in oncology},
  volume={11},
  pages={759007},
  year={2021},
  publisher={Frontiers Media SA}
}

@article{hu2021spagcn,
  title={SpaGCN: Integrating gene expression, spatial location and histology to identify spatial domains and spatially variable genes by graph convolutional network},
  author={Hu, Jian and Li, Xiangjie and Coleman, Kyle and Schroeder, Amelia and Ma, Nan and Irwin, David J and Lee, Edward B and Shinohara, Russell T and Li, Mingyao},
  journal={Nature methods},
  volume={18},
  number={11},
  pages={1342--1351},
  year={2021},
  publisher={Nature Publishing Group US New York}
}

@article{blampey2024sopa,
	title = {Sopa: a technology-invariant pipeline for analyses of image-based spatial omics},
	volume = {15},
	url = {https://www.nature.com/articles/s41467-024-48981-z},
	doi = {10.1038/s41467-024-48981-z},
	journal = {Nature Communications},
	author = {Blampey, Quentin and Mulder, Kevin and Gardet, Margaux and Christodoulidis, Stergios and Dutertre, Charles-Antoine and André, Fabrice and Ginhoux, Florent and Cournède, Paul-Henry},
	year = {2024},
	note = {Publisher: Nature Publishing Group},
	pages = {4981},
}

@inproceedings{garrido2023rankme,
  title={Rankme: Assessing the downstream performance of pretrained self-supervised representations by their rank},
  author={Garrido, Quentin and Balestriero, Randall and Najman, Laurent and Lecun, Yann},
  booktitle={International Conference on Machine Learning},
  pages={10929--10974},
  year={2023},
  organization={PMLR}
}

@article{staahl2016visualization,
  title={Visualization and analysis of gene expression in tissue sections by spatial transcriptomics},
  author={St{\aa}hl, Patrik L and Salm{\'e}n, Fredrik and Vickovic, Sanja and Lundmark, Anna and Navarro, Jos{\'e} Fern{\'a}ndez and Magnusson, Jens and Giacomello, Stefania and Asp, Michaela and Westholm, Jakub O and Huss, Mikael and others},
  journal={Science},
  volume={353},
  number={6294},
  pages={78--82},
  year={2016},
  publisher={American Association for the Advancement of Science}
}

@article{marx2021method,
  title={Method of the Year: spatially resolved transcriptomics},
  author={Marx, Vivien},
  journal={Nature methods},
  volume={18},
  number={1},
  pages={9--14},
  year={2021},
  publisher={Nature Publishing Group US New York}
}

@article{horst2024cellvit,
    title = {CellViT: Vision Transformers for precise cell segmentation and classification},
    journal = {Medical Image Analysis},
    volume = {94},
    pages = {103143},
    year = {2024},
    issn = {1361-8415},
    doi = {https://doi.org/10.1016/j.media.2024.103143},
    url = {https://www.sciencedirect.com/science/article/pii/S1361841524000689},
    author = {Fabian Hörst and Moritz Rempe and Lukas Heine and Constantin Seibold and Julius Keyl and Giulia Baldini and Selma Ugurel and Jens Siveke and Barbara Grünwald and Jan Egger and Jens Kleesiek},
}

@article{lu2021data,
  title={Data-efficient and weakly supervised computational pathology on whole-slide images},
  author={Lu, Ming Y and Williamson, Drew FK and Chen, Tiffany Y and Chen, Richard J and Barbieri, Matteo and Mahmood, Faisal},
  journal={Nature biomedical engineering},
  volume={5},
  number={6},
  pages={555--570},
  year={2021},
  publisher={Nature Publishing Group UK London}
}

@article{aresta2019bach,
  title={Bach: Grand challenge on breast cancer histology images},
  author={Aresta, Guilherme and Ara{\'u}jo, Teresa and Kwok, Scotty and Chennamsetty, Sai Saketh and Safwan, Mohammed and Alex, Varghese and Marami, Bahram and Prastawa, Marcel and Chan, Monica and Donovan, Michael and others},
  journal={Medical image analysis},
  volume={56},
  pages={122--139},
  year={2019},
  publisher={Elsevier}
}

@inproceedings{he2016deep,
  title={Deep residual learning for image recognition},
  author={He, Kaiming and Zhang, Xiangyu and Ren, Shaoqing and Sun, Jian},
  booktitle={Proceedings of the IEEE conference on computer vision and pattern recognition},
  pages={770--778},
  year={2016}
}

@article{bankhead2017qupath,
author = {Bankhead, Peter and Loughrey, Maurice and Fernandez, Jose and Dombrowski, Yvonne and Mcart, Darragh and Dunne, Philip and Mcquaid, Stephen and Gray, Ronan and Murray, Liam and Coleman, Helen and James, Jacqueline and Salto-Tellez, Manuel and Hamilton, Peter},
year = {2017},
month = {12},
pages = {},
title = {QuPath: Open source software for digital pathology image analysis},
volume = {7},
journal = {Scientific Reports},
doi = {10.1038/s41598-017-17204-5}
}

@article{wang2022transformer,
  title={Transformer-based unsupervised contrastive learning for histopathological image classification},
  author={Wang, Xiyue and Yang, Sen and Zhang, Jun and Wang, Minghui and Zhang, Jing and Yang, Wei and Huang, Junzhou and Han, Xiao},
  journal={Medical Image Analysis},
  volume={81},
  pages={102559},
  year={2022},
  publisher={Elsevier}
}

@inproceedings{deng2009imagenet,
  title={Imagenet: A large-scale hierarchical image database},
  author={Deng, Jia and Dong, Wei and Socher, Richard and Li, Li-Jia and Li, Kai and Fei-Fei, Li},
  booktitle={2009 IEEE conference on computer vision and pattern recognition},
  pages={248--255},
  year={2009},
  organization={Ieee}
}

@article{oord2018representation,
  title={Representation learning with contrastive predictive coding},
  author={Oord, Aaron van den and Li, Yazhe and Vinyals, Oriol},
  journal={arXiv preprint arXiv:1807.03748},
  year={2018}
}

@inproceedings{chen2020simple,
  title={A simple framework for contrastive learning of visual representations},
  author={Chen, Ting and Kornblith, Simon and Norouzi, Mohammad and Hinton, Geoffrey},
  booktitle={International conference on machine learning},
  pages={1597--1607},
  year={2020},
  organization={PMLR}
}

@article{parigi2022spatial,
  author       = {Parigi, S. M. and Larsson, L. and Das, S. and Ramirez Flores, R. O. and others},
  title        = {The spatial transcriptomic landscape of the healing mouse intestine following damage},
  journal      = {Nature Communications},
  year         = 2022,
  volume       = {13},
  number       = {1},
  pages        = {828},
  doi          = {10.1038/s41467-022-28423-2},
  pmid         = {35149721}
}

@article{gouin2021ncadherin,
  author       = {Gouin, K. H. 3rd and Ing, N. and Plummer, J. T. and Rosser, C. J. and others},
  title        = {An N-Cadherin 2 expressing epithelial cell subpopulation predicts response to surgery, chemotherapy and immunotherapy in bladder cancer},
  journal      = {Nature Communications},
  year         = 2021,
  volume       = {12},
  number       = {1},
  pages        = {4906},
  doi          = {10.1038/s41467-021-25205-2},
  pmid         = {34385456}
}

@article{liu2022single,
	author = {Liu, Tong and Liu, Cheng and Yan, Meisi and Zhang, Lei and Zhang, Jing and Xiao, Min and Li, Zhigao and Wei, Xiaofan and Zhang, Hongquan},
	title = {{Single cell profiling of primary and paired metastatic lymph node tumors in breast cancer patients}},
	journal = {Nat. Commun.},
	volume = {13},
	number = {6823},
	pages = {1--17},
	year = {2022},
	month = nov,
	issn = {2041-1723},
	publisher = {Nature Publishing Group},
	doi = {10.1038/s41467-022-34581-2}
}

@article{ji2020multimodal,
	author = {Ji, Andrew L. and Rubin, Adam J. and Thrane, Kim and Jiang, Sizun and Reynolds, David L. and Meyers, Robin M. and Guo, Margaret G. and George, Benson M. and Mollbrink, Annelie and Bergenstr{\aa}hle, Joseph and Larsson, Ludvig and Bai, Yunhao and Zhu, Bokai and Bhaduri, Aparna and Meyers, Jordan M. and Rovira-Clav{\ifmmode\acute{e}\else\'{e}\fi}, Xavier and Hollmig, S. Tyler and Aasi, Sumaira Z. and Nolan, Garry P. and Lundeberg, Joakim and Khavari, Paul A.},
	title = {{Multimodal Analysis of Composition and Spatial Architecture in Human Squamous Cell Carcinoma}},
	journal = {Cell},
	volume = {182},
	number = {2},
	pages = {497--51422},
	year = {2020},
	month = jul,
	issn = {1097-4172},
	publisher = {Elsevier Inc.},
	eprint = {32579974},
	doi = {10.1016/j.cell.2020.05.039}
}

@article{xue2023schwann,
	author = {Xue, Meilin and Zhu, Youwei and Jiang, Yongsheng and Han, Lijie and Shi, Minmin and Su, Rui and Wang, Liwen and Xiong, Cheng and Wang, Chaofu and Wang, Ting and Deng, Shijie and Wu, Dong and Cao, Yizhi and Dong, Lei and Bai, Fan and Zhao, Shulin and Deng, Xiaxing and Peng, Chenghong and Li, Hongwei and Chen, Jianjun and Shen, Baiyong and Jiang, Lingxi and Chen, Hao},
	title = {{Schwann cells regulate tumor cells and cancer-associated fibroblasts in the pancreatic ductal adenocarcinoma microenvironment}},
	journal = {Nat. Commun.},
	volume = {14},
	number = {4600},
	pages = {1--18},
	year = {2023},
	month = jul,
	issn = {2041-1723},
	publisher = {Nature Publishing Group},
	doi = {10.1038/s41467-023-40314-w}
}

@article{barrozo2023sars,
	author = {Barrozo, Enrico R. and Seferovic, Maxim D. and Castro, Eumenia C. C. and Major, Angela M. and Moorshead, David N. and Jochum, Michael D. and Rojas, Ricardo Ferral and Shope, Cynthia D. and Aagaard, Kjersti M.},
	title = {{SARS-CoV-2 niches in human placenta revealed by spatial transcriptomics}},
	journal = {Med},
	volume = {4},
	number = {9},
	pages = {612--634.e4},
	year = {2023},
	month = sep,
	issn = {2666-6340},
	publisher = {Cell Press},
	doi = {10.1016/j.medj.2023.06.003}
}

@article{ortiz2020molecular,
	author = {Ortiz, Cantin and Navarro, Jose Fernandez and Jurek, Aleksandra and M{\ifmmode\ddot{a}\else\"{a}\fi}rtin, Antje and Lundeberg, Joakim and Meletis, Konstantinos},
	title = {{Molecular atlas of the adult mouse brain}},
	journal = {Sci. Adv.},
	volume = {6},
	number = {26},
	year = {2020},
	month = jun,
	issn = {2375-2548},
	publisher = {American Association for the Advancement of Science},
	doi = {10.1126/sciadv.abb3446}
}

@article{ferreira2021integration,
	author = {Ferreira, Ricardo Melo and Sabo, Angela R. and Winfree, Seth and Collins, Kimberly S. and Janosevic, Danielle and Gulbronson, Connor J. and Cheng, Ying-Hua and Casbon, Lauren and Barwinska, Daria and Ferkowicz, Michael J. and Xuei, Xiaoling and Zhang, Chi and Dunn, Kenneth W. and Kelly, Katherine J. and Sutton, Timothy A. and Hato, Takashi and Dagher, Pierre C. and El-Achkar, Tarek M. and Eadon, Michael T.},
	title = {{Integration of spatial and single-cell transcriptomics localizes epithelial cell{\textendash}immune cross-talk in kidney injury}},
	journal = {JCI Insight},
	volume = {6},
	number = {12},
	year = {2021},
	month = jun,
	issn = {0021-9738},
	publisher = {American Society for Clinical Investigation},
	doi = {10.1172/jci.insight.147703}
}

@article{giraud2022trem,
	author = {Giraud, Julie and Chalopin, Domitille and Ramel, Elo{\ifmmode\ddot{\imath}\else\"{\i}\fi}se and Boyer, Thomas and Zouine, Atika and Derieppe, Marie-Alix and Larmonier, Nicolas and Adotevi, Olivier and Le Bail, Brigitte and Blanc, Jean-Fr{\ifmmode\acute{e}\else\'{e}\fi}d{\ifmmode\acute{e}\else\'{e}\fi}ric and Chiche, Laurence and Nikolski, Macha and Saleh, Maya},
	title = {{TREM1+CD163+ regulatory myeloid cells expand in steatohepatitis-HCC and associate with poor prognosis and therapeutic resistance to immune checkpoint blockade}},
	journal = {bioRxiv},
	pages = {2022.11.09.515839},
	year = {2022},
	month = nov,
	publisher = {Cold Spring Harbor Laboratory},
	eprint = {2022.11.09.515839},
	url = {https://doi.org/10.1101/2022.11.09.515839}
}

@article{janesick2023high,
	author = {Janesick, Amanda and Shelansky, Robert and Gottscho, Andrew D. and Wagner, Florian and Williams, Stephen R. and Rouault, Morgane and Beliakoff, Ghezal and Morrison, Carolyn A. and Oliveira, Michelli F. and Sicherman, Jordan T. and Kohlway, Andrew and Abousoud, Jawad and Drennon, Tingsheng Yu and Mohabbat, Seayar H. and Taylor, Sarah E. B.},
	title = {{High resolution mapping of the tumor microenvironment using integrated single-cell, spatial and in situ analysis}},
	journal = {Nat. Commun.},
	volume = {14},
	number = {8353},
	pages = {1--15},
	year = {2023},
	month = dec,
	issn = {2041-1723},
	publisher = {Nature Publishing Group},
	doi = {10.1038/s41467-023-43458-x}
}

@article{graciaVillacampa2021genome,
	author = {Gracia Villacampa, Eva and Larsson, Ludvig and Mirzazadeh, Reza and Kvastad, Linda and Andersson, Alma and Mollbrink, Annelie and Kokaraki, Georgia and Monteil, Vanessa and Schultz, Niklas and Appelberg, Karin Sofia and Montserrat, Nuria and Zhang, Haibo and Penninger, Josef M. and Miesbach, Wolfgang and Mirazimi, Ali and Carlson, Joseph and Lundeberg, Joakim},
	title = {{Genome-wide spatial expression profiling in formalin-fixed tissues}},
	journal = {Cell Genomics},
	volume = {1},
	number = {3},
	pages = {100065},
	year = {2021},
	month = dec,
	issn = {2666-979X},
	publisher = {Elsevier},
	doi = {10.1016/j.xgen.2021.100065}
}

@article{voigt2023gene,
	author = {Voigt, Andrew P. and Mullin, Nathaniel K. and Navratil, Emma M. and Flamme-Wiese, Miles J. and Lin, Li-Chun and Scheetz, Todd E. and Han, Ian C. and Stone, Edwin M. and Tucker, Budd A. and Mullins, Robert F.},
	title = {{Gene Expression Within a Human Choroidal Neovascular Membrane Using Spatial Transcriptomics}},
	journal = {Invest. Ophthalmol. Visual Sci.},
	volume = {64},
	number = {13},
	pages = {40},
	year = {2023},
	month = oct,
	issn = {1552-5783},
	publisher = {The Association for Research in Vision and Ophthalmology},
	doi = {10.1167/iovs.64.13.40}
}

@article{beppu2023epithelial,
	author = {Beppu, Andrew K. and Zhao, Juanjuan and Yao, Changfu and Carraro, Gianni and Israely, Edo and Coelho, Anna Lucia and Drake, Katherine and Hogaboam, Cory M. and Parks, William C. and Kolls, Jay K. and Stripp, Barry R.},
	title = {{Epithelial plasticity and innate immune activation promote lung tissue remodeling following respiratory viral infection}},
	journal = {Nat. Commun.},
	volume = {14},
	number = {5814},
	pages = {1--16},
	year = {2023},
	month = sep,
	issn = {2041-1723},
	publisher = {Nature Publishing Group},
	doi = {10.1038/s41467-023-41387-3}
}

@article{pakula2024distinct,
	author = {Pakula, Hubert and Omar, Mohamed and Carelli, Ryan and Pederzoli, Filippo and Fanelli, Giuseppe Nicol{\ifmmode\grave{o}\else\`{o}\fi} and Pannellini, Tania and Socciarelli, Fabio and Van Emmenis, Lucie and Rodrigues, Silvia and Fidalgo-Ribeiro, Caroline and Nuzzo, Pier Vitale and Brady, Nicholas J. and Dinalankara, Wikum and Jere, Madhavi and Valencia, Itzel and Saladino, Christopher and Stone, Jason and Unkenholz, Caitlin and Garner, Richard and Alexanderani, Mohammad K. and Khani, Francesca and de Almeida, Francisca Nunes and Abate-Shen, Cory and Greenblatt, Matthew B. and Rickman, David S. and Barbieri, Christopher E. and Robinson, Brian D. and Marchionni, Luigi and Loda, Massimo},
	title = {{Distinct mesenchymal cell states mediate prostate cancer progression}},
	journal = {Nat. Commun.},
	volume = {15},
	number = {363},
	pages = {1--21},
	year = {2024},
	month = jan,
	issn = {2041-1723},
	publisher = {Nature Publishing Group},
	doi = {10.1038/s41467-023-44210-1}
}

@article{oliveira2024characterization,
	author = {Oliveira, Michelli F. and Romero, Juan P. and Chung, Meii and Williams, Stephen and Gottscho, Andrew D. and Gupta, Anushka and Pilipauskas, Susan E. and Mohabbat, Syrus and Raman, Nandhini and Sukovich, David and Patterson, David and Taylor, Sarah E. B.},
	title = {Characterization of immune cell populations in the tumor microenvironment of colorectal cancer using high definition spatial profiling},
	elocation-id = {2024.06.04.597233},
	year = {2024},
	doi = {10.1101/2024.06.04.597233},
	publisher = {Cold Spring Harbor Laboratory},
	URL = {https://www.biorxiv.org/content/early/2024/06/05/2024.06.04.597233},
	eprint = {https://www.biorxiv.org/content/early/2024/06/05/2024.06.04.597233.full.pdf},
	journal = {bioRxiv}
}

@article{Domanskyi2023.07.27.550727,
	author = {Domanskyi, Sergii and Srivastava, Anuj and Kaster, Jessica and Li, Haiyin and Herlyn, Meenhard and Rubinstein, Jill C. and Chuang, Jeffrey H.},
	title = {Nextflow Pipeline for Visium and H\&E Data from Patient-Derived Xenograft Samples},
	elocation-id = {2023.07.27.550727},
	year = {2023},
	doi = {10.1101/2023.07.27.550727},
	publisher = {Cold Spring Harbor Laboratory},
	URL = {https://www.biorxiv.org/content/early/2023/07/30/2023.07.27.550727},
	eprint = {https://www.biorxiv.org/content/early/2023/07/30/2023.07.27.550727.full.pdf},
	journal = {bioRxiv}
}

@article{Vannan2023.12.15.571954,
	author = {Vannan, Annika and Lyu, Ruqian and Williams, Arianna L. and Negretti, Nicholas M. and Mee, Evan D. and Hirsh, Joseph and Hirsh, Samuel and Nichols, David S. and Calvi, Carla L. and Taylor, Chase J. and Polosukhin, Vasiliy. V. and Serezani, Ana PM and McCall, A. Scott and Gokey, Jason J. and Shim, Heejung and Ware, Lorraine B. and Bacchetta, Matthew J. and Shaver, Ciara M. and Blackwell, Timothy S. and Walia, Rajat and Sucre, Jennifer MS and Kropski, Jonathan A. and McCarthy, Davis J and Banovich, Nicholas E.},
	title = {Image-based spatial transcriptomics identifies molecular niche dysregulation associated with distal lung remodeling in pulmonary fibrosis},
	elocation-id = {2023.12.15.571954},
	year = {2023},
	doi = {10.1101/2023.12.15.571954},
	publisher = {Cold Spring Harbor Laboratory},
	URL = {https://www.biorxiv.org/content/early/2023/12/17/2023.12.15.571954},
	eprint = {https://www.biorxiv.org/content/early/2023/12/17/2023.12.15.571954.full.pdf},
	journal = {bioRxiv}
}

@article{kanemaru_spatially_2023,
	title = {Spatially resolved multiomics of human cardiac niches},
	volume = {619},
	issn = {1476-4687},
	url = {https://doi.org/10.1038/s41586-023-06311-1},
	doi = {10.1038/s41586-023-06311-1},
	number = {7971},
	journal = {Nature},
	author = {Kanemaru, Kazumasa and Cranley, James and Muraro, Daniele and Miranda, Antonio M. A. and Ho, Siew Yen and Wilbrey-Clark, Anna and Patrick Pett, Jan and Polanski, Krzysztof and Richardson, Laura and Litvinukova, Monika and Kumasaka, Natsuhiko and Qin, Yue and Jablonska, Zuzanna and Semprich, Claudia I. and Mach, Lukas and Dabrowska, Monika and Richoz, Nathan and Bolt, Liam and Mamanova, Lira and Kapuge, Rakeshlal and Barnett, Sam N. and Perera, Shani and Talavera-López, Carlos and Mulas, Ilaria and Mahbubani, Krishnaa T. and Tuck, Liz and Wang, Lu and Huang, Margaret M. and Prete, Martin and Pritchard, Sophie and Dark, John and Saeb-Parsy, Kourosh and Patel, Minal and Clatworthy, Menna R. and Hübner, Norbert and Chowdhury, Rasheda A. and Noseda, Michela and Teichmann, Sarah A.},
	month = jul,
	year = {2023},
	pages = {801--810},
}

@article{heiser_molecular_2023,
	title = {Molecular cartography uncovers evolutionary and microenvironmental dynamics in sporadic colorectal tumors},
	volume = {186},
	issn = {0092-8674},
	url = {https://doi.org/10.1016/j.cell.2023.11.006},
	doi = {10.1016/j.cell.2023.11.006},
	number = {25},
	urldate = {2024-10-25},
	journal = {Cell},
	author = {Heiser, Cody N. and Simmons, Alan J. and Revetta, Frank and McKinley, Eliot T. and Ramirez-Solano, Marisol A. and Wang, Jiawei and Kaur, Harsimran and Shao, Justin and Ayers, Gregory D. and Wang, Yu and Glass, Sarah E. and Tasneem, Naila and Chen, Zhengyi and Qin, Yan and Kim, William and Rolong, Andrea and Chen, Bob and Vega, Paige N. and Drewes, Julia L. and Markham, Nicholas O. and Saleh, Nabil and Nikolos, Fotis and Vandekar, Simon and Jones, Angela L. and Washington, M. Kay and Roland, Joseph T. and Chan, Keith S. and Schürpf, Thomas and Sears, Cynthia L. and Liu, Qi and Shrubsole, Martha J. and Coffey, Robert J. and Lau, Ken S.},
	month = dec,
	year = {2023},
	note = {Publisher: Elsevier},
	pages = {5620--5637.e16},
}

@article{fonseca2023single,
	author = {Fonseca, Marcos A. S. and Haro, Marcela and Wright, Kelly N. and Lin, Xianzhi and Abbasi, Forough and Sun, Jennifer and Hernandez, Lourdes and Orr, Natasha L. and Hong, Jooyoon and Choi-Kuaea, Yunhee and Maluf, Horacio M. and Balzer, Bonnie L. and Fishburn, Aaron and Hickey, Ryan and Cass, Ilana and Goodridge, Helen S. and Truong, Mireille and Wang, Yemin and Pisarska, Margareta D. and Dinh, Huy Q. and E. L.-Naggar, Amal and Huntsman, David G. and Anglesio, Michael S. and Goodman, Marc T. and Medeiros, Fabiola and Siedhoff, Matthew and Lawrenson, Kate},
	title = {{Single-cell transcriptomic analysis of endometriosis}},
	journal = {Nat. Genet.},
	volume = {55},
	pages = {255--267},
	year = {2023},
	month = feb,
	issn = {1546-1718},
	publisher = {Nature Publishing Group},
	doi = {10.1038/s41588-022-01254-1}
}

@article{anton2024binge,
	author = {Anton, Paige E. and Rutt, Lauren N. and Kaufman, Michael L. and Busquet, Nicolas and Kovacs, Elizabeth J. and McCullough, Rebecca L.},
	title = {{Binge ethanol exposure in advanced age elevates neuroinflammation and early indicators of neurodegeneration and cognitive impairment in female mice}},
	journal = {Brain Behav. Immun.},
	volume = {116},
	pages = {303--316},
	year = {2024},
	month = feb,
	issn = {0889-1591},
	publisher = {Academic Press},
	doi = {10.1016/j.bbi.2023.12.034}
}

@article{abdelhafiz2023ychromosome,
  author       = {Abdel-Hafiz, H. A. and Schafer, J. M. and Chen, X. and Xiao, T. and others},
  title        = {Y chromosome loss in cancer drives growth by evasion of adaptive immunity},
  journal      = {Nature},
  year         = 2023,
  volume       = {619},
  number       = {7970},
  pages        = {624-631},
  doi          = {10.1038/s41586-023-06083-w},
  pmid         = {37344596}
}

@article{karras2022cellular,
	author = {Karras, Panagiotis and Bordeu, Ignacio and Pozniak, Joanna and Nowosad, Ada and Pazzi, Cecilia and Van Raemdonck, Nina and Landeloos, Ewout and Van Herck, Yannick and Pedri, Dennis and Bervoets, Greet and Makhzami, Samira and Khoo, Jia Hui and Pavie, Benjamin and Lamote, Jochen and Marin-Bejar, Oskar and Dewaele, Michael and Liang, Han and Zhang, Xingju and Hua, Yichao and Wouters, Jasper and Browaeys, Robin and Bergers, Gabriele and Saeys, Yvan and Bosisio, Francesca and van den Oord, Joost and Lambrechts, Diether and Rustgi, Anil K. and Bechter, Oliver and Blanpain, Cedric and Simons, Benjamin D. and Rambow, Florian and Marine, Jean-Christophe},
	title = {{A cellular hierarchy in melanoma uncouples growth and metastasis}},
	journal = {Nature},
	volume = {610},
	pages = {190--198},
	year = {2022},
	month = oct,
	issn = {1476-4687},
	publisher = {Nature Publishing Group},
	doi = {10.1038/s41586-022-05242-7}
}

@article{canela2023spatially,
  author       = {Canela, V. H. and Bowen, W. S. and Ferreira, R. M. and Syed, F. and others},
  title        = {A spatially anchored transcriptomic atlas of the human kidney papilla identifies significant immune injury in patients with stone disease},
  journal      = {Nature Communications},
  year         = 2023,
  volume       = {14},
  number       = {1},
  pages        = {4140},
  doi          = {10.1038/s41467-023-41340-8},
  pmid         = {37468493}
}

@article{stahl2016visualization,
	author = {St{\aa}hl, Patrik L. and Salm{\ifmmode\acute{e}\else\'{e}\fi}n, Fredrik and Vickovic, Sanja and Lundmark, Anna and Navarro, Jos{\ifmmode\acute{e}\else\'{e}\fi} Fern{\ifmmode\acute{a}\else\'{a}\fi}ndez and Magnusson, Jens and Giacomello, Stefania and Asp, Michaela and Westholm, Jakub O. and Huss, Mikael and Mollbrink, Annelie and Linnarsson, Sten and Codeluppi, Simone and Borg, {\AA}ke and Pont{\ifmmode\acute{e}\else\'{e}\fi}n, Fredrik and Costea, Paul Igor and Sahl{\ifmmode\acute{e}\else\'{e}\fi}n, Pelin and Mulder, Jan and Bergmann, Olaf and Lundeberg, Joakim and Fris{\ifmmode\acute{e}\else\'{e}\fi}n, Jonas},
	title = {{Visualization and analysis of gene expression in tissue sections by spatial transcriptomics}},
	journal = {Science},
	volume = {353},
	number = {6294},
	pages = {78--82},
	year = {2016},
	month = jul,
	issn = {0036-8075},
	publisher = {American Association for the Advancement of Science},
	doi = {10.1126/science.aaf2403}
}

@article{andersson2021spatial,
	author = {Andersson, Alma and Larsson, Ludvig and Stenbeck, Linnea and Salm{\ifmmode\acute{e}\else\'{e}\fi}n, Fredrik and Ehinger, Anna and Wu, Sunny Z. and Al-Eryani, Ghamdan and Roden, Daniel and Swarbrick, Alex and Borg, {\AA}ke and Fris{\ifmmode\acute{e}\else\'{e}\fi}n, Jonas and Engblom, Camilla and Lundeberg, Joakim},
	title = {{Spatial deconvolution of HER2-positive breast cancer delineates tumor-associated cell type interactions}},
	journal = {Nat. Commun.},
	volume = {12},
	number = {6012},
	pages = {1--14},
	year = {2021},
	month = oct,
	issn = {2041-1723},
	publisher = {Nature Publishing Group},
	doi = {10.1038/s41467-021-26271-2}
}

@article{pilet2023preneoplastic,
	author = {Pilet, Jill and Hirsch, Theo Z. and Gupta, Barkha and Roehrig, Am{\ifmmode\acute{e}\else\'{e}\fi}lie and Morcrette, Guillaume and Pire, Aurore and Letouz{\ifmmode\acute{e}\else\'{e}\fi}, Eric and Fresneau, Brice and Taque, Sophie and Brugi{\ifmmode\grave{e}\else\`{e}\fi}res, Laurence and Branchereau, Sophie and Chardot, Christophe and Aerts, Isabelle and Sarnacki, Sabine and Fabre, Monique and Guettier, Catherine and Rebouissou, Sandra and Zucman-Rossi, Jessica},
	title = {{Preneoplastic liver colonization by 11p15.5 altered mosaic cells in young children with hepatoblastoma}},
	journal = {Nat. Commun.},
	volume = {14},
	year = {2023},
	publisher = {Nature Publishing Group},
	doi = {10.1038/s41467-023-42418-9}
}

@article{barrozo2024zika,
	author = {Barrozo, Enrico R. and Seferovic, Maxim D. and Hamilton, Mark P. and Moorshead, David N. and Jochum, Michael D. and Do, Trang and O'Neil, Derek S. and Suter, Melissa A. and Aagaard, Kjersti M.},
	title = {{Zika virus co-opts microRNA networks to persist in placental niches detected by spatial transcriptomics}},
	journal = {Am. J. Obstet. Gynecol.},
	volume = {230},
	number = {2},
	pages = {2511--25117},
	year = {2024},
	month = feb,
	issn = {1097-6868},
	publisher = {Elsevier Inc},
	eprint = {37598997},
	doi = {10.1016/j.ajog.2023.08.012}
}

@article{li2024yap,
	author = {Li, Rich Gang and Li, Xiao and Morikawa, Yuka and Grisanti-Canozo, Francisco J. and Meng, Fansen and Tsai, Chang-Ru and Zhao, Yi and Liu, Lin and Kim, Jong and Xie, Bing and Klysik, Elzbieta and Liu, Shijie and Samee, Md Abul Hassan and Martin, James F.},
	title = {{YAP induces a neonatal-like pro-renewal niche in the adult heart}},
	journal = {Nature cardiovascular research},
	volume = {3},
	number = {3},
	pages = {283},
	year = {2024},
	month = mar,
	publisher = {NIH Public Access},
	doi = {10.1038/s44161-024-00428-w}
}

@article{mauduit2022spatial,
	author = {Mauduit, Olivier and Delcroix, Vanessa and Umazume, Takeshi and de Paiva, Cintia S. and Dartt, Darlene A. and Makarenkova, Helen P.},
	title = {{Spatial transcriptomics of the lacrimal gland features macrophage activity and epithelium metabolism as key alterations during chronic inflammation}},
	journal = {Front. Immunol.},
	volume = {13},
	pages = {1011125},
	year = {2022},
	month = oct,
	issn = {1664-3224},
	publisher = {Frontiers},
	doi = {10.3389/fimmu.2022.1011125}
}

@article{kathe2022neurons,
	author = {Kathe, Claudia and Skinnider, Michael A. and Hutson, Thomas H. and Regazzi, Nicola and Gautier, Matthieu and Demesmaeker, Robin and Komi, Salif and Ceto, Steven and James, Nicholas D. and Cho, Newton and Baud, Laetitia and Galan, Katia and Matson, Kaya J. E. and Rowald, Andreas and Kim, Kyungjin and Wang, Ruijia and Minassian, Karen and Prior, John O. and Asboth, Leonie and Barraud, Quentin and Lacour, St{\ifmmode\acute{e}\else\'{e}\fi}phanie P. and Levine, Ariel J. and Wagner, Fabien and Bloch, Jocelyne and Squair, Jordan W. and Courtine, Gr{\ifmmode\acute{e}\else\'{e}\fi}goire},
	title = {{The neurons that restore walking after paralysis}},
	journal = {Nature},
	volume = {611},
	pages = {540--547},
	year = {2022},
	month = nov,
	issn = {1476-4687},
	publisher = {Nature Publishing Group},
	doi = {10.1038/s41586-022-05385-7}
}

@article{maniatis2019spatiotemporal,
  title={Spatiotemporal dynamics of molecular pathology in amyotrophic lateral sclerosis},
  author={Maniatis, Silas and {\"A}ij{\"o}, Tarmo and Vickovic, Sanja and Braine, Catherine and Kang, Kristy and Mollbrink, Annelie and Fagegaltier, Delphine and Andrusivov{\'a}, {\v{Z}}aneta and Saarenp{\"a}{\"a}, Sami and Saiz-Castro, Gonzalo and others},
  journal={Science},
  volume={364},
  number={6435},
  pages={89--93},
  year={2019},
  publisher={American Association for the Advancement of Science}
}

@article{fu2023spatial,
  author       = {Fu, R. and Norris, G. A. and Willard, N. and Griesinger, A. M. and others},
  title        = {Spatial transcriptomic analysis delineates epithelial and mesenchymal subpopulations and transition stages in childhood ependymoma},
  journal      = {Neuro-Oncology},
  year         = 2023,
  volume       = {25},
  number       = {4},
  pages        = {786-798},
  doi          = {10.1093/neuonc/noad070},
  pmid         = {36215273}
}

@article{moeyersoms2023spatial,
	author = {Moeyersoms, Acadia H. M. and Gallo, Ryan A. and Zhang, Michelle G. and Stathias, Vasileios and Maeng, Michelle M. and Owens, Dawn and Khzam, Rayan Abou and Sayegh, Yoseph and Maza, Cynthia and Dubovy, Sander R. and Tse, David T. and Pelaez, Daniel},
	title = {{Spatial Transcriptomics Identifies Expression Signatures Specific to Lacrimal Gland Adenoid Cystic Carcinoma Cells}},
	journal = {Cancers},
	volume = {15},
	number = {12},
	pages = {3211.},
	year = {2023},
	month = jun,
	issn = {2072-6694},
	publisher = {See full text options at MDPI},
	eprint = {37370820},
	doi = {10.3390/cancers15123211}
}

@article{schabitz2022spatial,
	author = {Sch{\ifmmode\ddot{a}\else\"{a}\fi}bitz, A. and Hillig, C. and Mubarak, M. and Jargosch, M. and Farnoud, A. and Scala, E. and Kurzen, N. and Pilz, A. C. and Bhalla, N. and Thomas, J. and Stahle, M. and Biedermann, T. and Schmidt-Weber, C. B. and Theis, F. and Garzorz-Stark, N. and Eyerich, K. and Menden, M. P. and Eyerich, S.},
	title = {{Spatial transcriptomics landscape of lesions from non-communicable inflammatory skin diseases}},
	journal = {Nat. Commun.},
	volume = {13},
	number = {7729},
	pages = {1--13},
	year = {2022},
	month = dec,
	issn = {2041-1723},
	publisher = {Nature Publishing Group},
	doi = {10.1038/s41467-022-35319-w}
}

@article{krausgruber2023single,
	author = {Krausgruber, Thomas and Redl, Anna and Barreca, Daniele and Doberer, Konstantin and Romanovskaia, Daria and Dobnikar, Lina and Guarini, Maria and Unterluggauer, Luisa and Kleissl, Lisa and Atzm{\ifmmode\ddot{u}\else\"{u}\fi}ller, Denise and Mayerhofer, Carolina and Kopf, Aglaja and Saluzzo, Simona and Lim, Clarice X. and Rexie, Praveen and Weichhart, Thomas and Bock, Christoph and Stary, Georg},
	title = {{Single-cell and spatial transcriptomics reveal aberrant lymphoid developmental programs driving granuloma formation}},
	journal = {Immunity},
	volume = {56},
	number = {2},
	pages = {289--306.e7},
	year = {2023},
	month = feb,
	issn = {1074-7613},
	publisher = {Cell Press},
	doi = {10.1016/j.immuni.2023.01.014}
}

@article{lake2023an,
	author = {Lake, Blue B. and Menon, Rajasree and Winfree, Seth and Hu, Qiwen and Melo Ferreira, Ricardo and Kalhor, Kian and Barwinska, Daria and Otto, Edgar A. and Ferkowicz, Michael and Diep, Dinh and Plongthongkum, Nongluk and Knoten, Amanda and Urata, Sarah and Mariani, Laura H. and Naik, Abhijit S. and Eddy, Sean and Zhang, Bo and Wu, Yan and Salamon, Diane and Williams, James C. and Wang, Xin and Balderrama, Karol S. and Hoover, Paul J. and Murray, Evan and Marshall, Jamie L. and Noel, Teia and Vijayan, Anitha and Hartman, Austin and Chen, Fei and Waikar, Sushrut S. and Rosas, Sylvia E. and Wilson, Francis P. and Palevsky, Paul M. and Kiryluk, Krzysztof and Sedor, John R. and Toto, Robert D. and Parikh, Chirag R. and Kim, Eric H. and Satija, Rahul and Greka, Anna and Macosko, Evan Z. and Kharchenko, Peter V. and Gaut, Joseph P. and Hodgin, Jeffrey B. and Eadon, Michael T. and Dagher, Pierre C. and El-Achkar, Tarek M. and Zhang, Kun and Kretzler, Matthias and Jain, Sanjay},
	title = {{An atlas of healthy and injured cell states and niches in the human kidney}},
	journal = {Nature},
	volume = {619},
	pages = {585--594},
	year = {2023},
	month = jul,
	issn = {1476-4687},
	publisher = {Nature Publishing Group},
	doi = {10.1038/s41586-023-05769-3}
}

@article{moses2022museum,
	author = {Moses, Lambda and Pachter, Lior},
	title = {{Museum of spatial transcriptomics}},
	journal = {Nat. Methods},
	volume = {19},
	pages = {534--546},
	year = {2022},
	month = may,
	issn = {1548-7105},
	publisher = {Nature Publishing Group},
	doi = {10.1038/s41592-022-01409-2}
}

@article{mcKellar2023spatial,
	author = {McKellar, David W. and Mantri, Madhav and Hinchman, Meleana M. and Parker, John S. L. and Sethupathy, Praveen and Cosgrove, Benjamin D. and De Vlaminck, Iwijn},
	title = {{Spatial mapping of the total transcriptome by in situ polyadenylation}},
	journal = {Nat. Biotechnol.},
	volume = {41},
	pages = {513--520},
	year = {2023},
	month = apr,
	issn = {1546-1696},
	publisher = {Nature Publishing Group},
	doi = {10.1038/s41587-022-01517-6}
}

@article{mckellar2021large,
  author       = {McKellar, D. W. and Walter, L. D. and Song, L. T. and Mantri, M. and others},
  title        = {Large-scale integration of single-cell transcriptomic data captures transitional progenitor states in mouse skeletal muscle regeneration},
  journal      = {Communications Biology},
  year         = 2021,
  volume       = {4},
  number       = {1},
  pages        = {1280},
  doi          = {10.1038/s42003-021-02756-8},
  pmid         = {34773081}
}

@article{vicari2023spatial,
	author = {Vicari, Marco and Mirzazadeh, Reza and Nilsson, Anna and Shariatgorji, Reza and Bj{\ifmmode\ddot{a}\else\"{a}\fi}rterot, Patrik and Larsson, Ludvig and Lee, Hower and Nilsson, Mats and Foyer, Julia and Ekvall, Markus and Czarnewski, Paulo and Zhang, Xiaoqun and Svenningsson, Per and K{\ifmmode\ddot{a}\else\"{a}\fi}ll, Lukas and Andr{\ifmmode\acute{e}\else\'{e}\fi}n, Per E. and Lundeberg, Joakim},
	title = {{Spatial multimodal analysis of transcriptomes and metabolomes in tissues}},
	journal = {Nat. Biotechnol.},
	pages = {1--5},
	year = {2023},
	month = sep,
	issn = {1546-1696},
	publisher = {Nature Publishing Group},
	doi = {10.1038/s41587-023-01937-y}
}

@article{lin2023single,
	author = {Lin, Wenyong and Chen, Xin and Wang, Dongyuan and Lu, Ruixia and Zhang, Chunling and Niu, Zhenchao and Chen, Jie and Ruan, Xiaofen and Wang, Xiaolong},
	title = {{Single-nucleus ribonucleic acid-sequencing and spatial transcriptomics reveal the cardioprotection of Shexiang Baoxin Pill (SBP) in mice with myocardial ischemia-reperfusion injury}},
	journal = {Front. Pharmacol.},
	volume = {14},
	pages = {1173649},
	year = {2023},
	month = may,
	issn = {1663-9812},
	publisher = {Frontiers},
	doi = {10.3389/fphar.2023.1173649}
}

@article{andrews2024single,
	author = {Andrews, Tallulah S. and Nakib, Diana and Perciani, Catia T. and Ma, Xue Zhong and Liu, Lewis and Winter, Erin and Camat, Damra and Chung, Sai W. and Lumanto, Patricia and Manuel, Justin and Mangroo, Shantel and Hansen, Bettina and Arpinder, Bal and Thoeni, Cornelia and Sayed, Blayne and Feld, Jordan and Gehring, Adam and Gulamhusein, Aliya and Hirschfield, Gideon M. and Ricciuto, Amanda and Bader, Gary D. and McGilvray, Ian D. and MacParland, Sonya},
	title = {{Single-cell, single-nucleus, and spatial transcriptomics characterization of the immunological landscape in the healthy and PSC human liver}},
	journal = {J. Hepatol.},
	volume = {80},
	number = {5},
	pages = {730--743},
	year = {2024},
	month = may,
	issn = {1600-0641},
	publisher = {The Author(sElsevier B.V},
	eprint = {38199298},
	doi = {10.1016/j.jhep.2023.12.023}
}

@article{asp2019spatiotemporal,
	author = {Asp, Michaela and Giacomello, Stefania and Larsson, Ludvig and Wu, Chenglin and F{\ifmmode\ddot{u}\else\"{u}\fi}rth, Daniel and Qian, Xiaoyan and W{\ifmmode\ddot{a}\else\"{a}\fi}rdell, Eva and Custodio, Joaquin and Reimeg{\aa}rd, Johan and Salm{\ifmmode\acute{e}\else\'{e}\fi}n, Fredrik and {\ifmmode\ddot{O}\else\"{O}\fi}sterholm, Cecilia and St{\aa}hl, Patrik L. and Sundstr{\ifmmode\ddot{o}\else\"{o}\fi}m, Erik and {\AA}kesson, Elisabet and Bergmann, Olaf and Bienko, Magda and M{\aa}nsson-Broberg, Agneta and Nilsson, Mats and Sylv{\ifmmode\acute{e}\else\'{e}\fi}n, Christer and Lundeberg, Joakim},
	title = {{A Spatiotemporal Organ-Wide Gene Expression and Cell Atlas of the Developing Human Heart}},
	journal = {Cell},
	volume = {179},
	number = {7},
	pages = {1647--166019},
	year = {2019},
	month = dec,
	issn = {1097-4172},
	publisher = {Elsevier Inc},
	eprint = {31835037},
	doi = {10.1016/j.cell.2019.11.025}
}

@misc{andrusivova2023exst,
  author       = {Andrusivova, Zaneta and Fan, Yuhang},
  title        = {Ex-ST},
  year         = 2023,
  publisher    = {Mendeley Data},
  version      = {V1},
  doi          = {10.17632/nrbsxrk9mp.1},
  url          = {https://data.mendeley.com/datasets/nrbsxrk9mp/1}
}

@article{mirzazadeh2023spatially,
	author = {Mirzazadeh, Reza and Andrusivova, Zaneta and Larsson, Ludvig and Newton, Phillip T. and Galicia, Leire Alonso and Abalo, Xes{\ifmmode\acute{u}\else\'{u}\fi}s M. and Avijgan, Mahtab and Kvastad, Linda and Denadai-Souza, Alexandre and Stakenborg, Nathalie and Firsova, Alexandra B. and Shamikh, Alia and Jurek, Aleksandra and Schultz, Niklas and Nist{\ifmmode\acute{e}\else\'{e}\fi}r, Monica and Samakovlis, Christos and Boeckxstaens, Guy and Lundeberg, Joakim},
	title = {{Spatially resolved transcriptomic profiling of degraded and challenging fresh frozen samples}},
	journal = {Nat. Commun.},
	volume = {14},
	number = {509},
	pages = {1--16},
	year = {2023},
	month = jan,
	issn = {2041-1723},
	publisher = {Nature Publishing Group},
	doi = {10.1038/s41467-023-36071-5}
}

@article{erickson2022spatially,
	author = {Erickson, Andrew and He, Mengxiao and Berglund, Emelie and Marklund, Maja and Mirzazadeh, Reza and Schultz, Niklas and Kvastad, Linda and Andersson, Alma and Bergenstr{\aa}hle, Ludvig and Bergenstr{\aa}hle, Joseph and Larsson, Ludvig and Alonso Galicia, Leire and Shamikh, Alia and Basmaci, Elisa and D{\ifmmode\acute{\imath}\else\'{\i}\fi}az De St{\aa}hl, Teresita and Rajakumar, Timothy and Doultsinos, Dimitrios and Thrane, Kim and Ji, Andrew L. and Khavari, Paul A. and Tarish, Firaz and Tanoglidi, Anna and Maaskola, Jonas and Colling, Richard and Mirtti, Tuomas and Hamdy, Freddie C. and Woodcock, Dan J. and Helleday, Thomas and Mills, Ian G. and Lamb, Alastair D. and Lundeberg, Joakim},
	title = {{Spatially resolved clonal copy number alterations in benign and malignant tissue}},
	journal = {Nature},
	volume = {608},
	pages = {360--367},
	year = {2022},
	month = aug,
	issn = {1476-4687},
	publisher = {Nature Publishing Group},
	doi = {10.1038/s41586-022-05023-2}
}

@article{xiao2021tumor,
	author = {Xiao, Yi and Yu, Dihua},
	title = {{Tumor microenvironment as a therapeutic target in cancer}},
	journal = {Pharmacol. Ther.},
	volume = {221},
	pages = {107753},
	year = {2021},
	month = may,
	issn = {0163-7258},
	publisher = {Pergamon},
	doi = {10.1016/j.pharmthera.2020.107753}
}

@article{devisser2023evolving,
	author = {de Visser, Karin E. and Joyce, Johanna A.},
	title = {{The evolving tumor microenvironment: From cancer initiation to metastatic outgrowth}},
	journal = {Cancer Cell},
	volume = {41},
	number = {3},
	pages = {374--403},
	year = {2023},
	month = mar,
	issn = {1535-6108},
	publisher = {Cell Press},
	doi = {10.1016/j.ccell.2023.02.016}
}

@misc{abalo2021human,
  author       = {Abalo, Xes{\'u}s and Thrane, Kim and Ji, Andrew Lin and Mirzazadeh, Reza and Khavari, Paul and Lundeberg, Joakim},
  title        = {Human squamous cell carcinoma, Visium},
  year         = 2021,
  publisher    = {Mendeley Data},
  version      = {V1},
  doi          = {10.17632/2bh5fchcv6.1},
  url          = {https://data.mendeley.com/datasets/2bh5fchcv6/1}
}

@misc{marklund2022prostate,
  author       = {Marklund, Maja},
  title        = {Prostate needle biopsies pre- and post-ADT: Count matrices, histological-, and Androgen receptor immunohistochemistry images},
  year         = 2022,
  publisher    = {Mendeley Data},
  version      = {V1},
  doi          = {10.17632/mdt8n2xgf4.1},
  url          = {https://data.mendeley.com/datasets/mdt8n2xgf4/1}
}

@misc{mirzazadeh2021human,
  author       = {Mirzazadeh, Reza and Larsson, Ludvig and Stakenborg, Nathalie and Abalo, Xes{\'u}s and Boeckxstaens, Guy and Lundeberg, Joakim},
  title        = {Human ileum, Visium},
  year         = 2021,
  publisher    = {Mendeley Data},
  version      = {V1},
  doi          = {10.17632/v8s9nz948s.1},
  url          = {https://data.mendeley.com/datasets/v8s9nz948s/1}
}

@article{villacampa2021genome,
	author = {Villacampa, Eva Gracia and Larsson, Ludvig and Mirzazadeh, Reza and Kvastad, Linda and Andersson, Alma and Mollbrink, Annelie and Kokaraki, Georgia and Monteil, Vanessa and Schultz, Niklas and Appelberg, Karin Sofia and Montserrat, Nuria and Zhang, Haibo and Penninger, Josef M. and Miesbach, Wolfgang and Mirazimi, Ali and Carlson, Joseph and Lundeberg, Joakim},
	title = {{Genome-wide spatial expression profiling in formalin-fixed tissues}},
	journal = {Cell Genom.},
	volume = {1},
	number = {3},
	pages = {100065.},
	year = {2021},
	month = dec,
	issn = {2666-979X},
	publisher = {2021 The Authors.},
	eprint = {36776149},
	doi = {10.1016/j.xgen.2021.100065}
}

@article{meylan2022tertiary,
	author = {Meylan, Maxime and Petitprez, Florent and Becht, Etienne and Bougo{\ifmmode\ddot{u}\else\"{u}\fi}in, Antoine and Pupier, Guilhem and Calvez, Anne and Giglioli, Ilenia and Verkarre, Virginie and Lacroix, Guillaume and Verneau, Johanna and Sun, Chen-Ming and Laurent-Puig, Pierre and Vano, Yann-Alexandre and Ela{\ifmmode\ddot{\imath}\else\"{\i}\fi}di, Reza and M{\ifmmode\acute{e}\else\'{e}\fi}jean, Arnaud and Sanchez-Salas, Rafa{\ifmmode\ddot{e}\else\"{e}\fi}l and Barret, Eric and Cathelineau, Xavier and Oudard, Stephane and Reynaud, Claude-Agn{\ifmmode\grave{e}\else\`{e}\fi}s and de Reyni{\ifmmode\grave{e}\else\`{e}\fi}s, Aur{\ifmmode\acute{e}\else\'{e}\fi}lien and Saut{\ifmmode\grave{e}\else\`{e}\fi}s-Fridman, Catherine and Fridman, Wolf Herman},
	title = {{Tertiary lymphoid structures generate and propagate anti-tumor antibody-producing plasma cells in renal cell cancer}},
	journal = {Immunity},
	volume = {55},
	number = {3},
	pages = {527--541.e5},
	year = {2022},
	month = mar,
	issn = {1074-7613},
	publisher = {Cell Press},
	doi = {10.1016/j.immuni.2022.02.001}
}

@article{chen2021an,
	author = {Chen, Xinlei and Xie, Saining and He, Kaiming},
	title = {{An Empirical Study of Training Self-Supervised Vision Transformers}},
	journal = {arXiv},
	year = {2021},
	month = apr,
	eprint = {2104.02057},
	doi = {10.48550/arXiv.2104.02057}
}

@inproceedings{yu2022coca,
  title   = {CoCa: Contrastive Captioners are Image-Text Foundation Models},
  author  = {Jiahui Yu and Zirui Wang and Vijay Vasudevan and Legg Yeung and Mojtaba Seyedhosseini and Yonghui Wu},
  year    = {2022}
}

@article{gatenbee2023valis,
	author = {Gatenbee, Chandler D. and Baker, Ann-Marie and Prabhakaran, Sandhya and Swinyard, Ottilie and Slebos, Robbert J. C. and Mandal, Gunjan and Mulholland, Eoghan and Andor, Noemi and Marusyk, Andriy and Leedham, Simon and Conejo-Garcia, Jose R. and Chung, Christine H. and Robertson-Tessi, Mark and Graham, Trevor A. and Anderson, Alexander R. A.},
	title = {{Virtual alignment of pathology image series for multi-gigapixel whole slide images}},
	journal = {Nat. Commun.},
	volume = {14},
	number = {4502},
	pages = {1--14},
	year = {2023},
	month = jul,
	issn = {2041-1723},
	publisher = {Nature Publishing Group},
	doi = {10.1038/s41467-023-40218-9}
}

@misc{zenodo2024spotiphy,
	title = {{Spotiphy: generative modeling in single-cell spatial whole transcriptomics}},
	journal = {Zenodo},
	year = {2024},
	month = jan,
	note = {[Online; accessed 16. May 2024]},
	doi = {10.5281/zenodo.10520022}
}

@misc{zenodo2024demo,
	title = {{Demo 10x Visium dataset for STQ}},
	journal = {Zenodo},
	year = {2024},
	month = feb,
	note = {[Online; accessed 16. May 2024]},
	doi = {10.5281/zenodo.10654467}
}

@article{valdeolivas2023charting,
	author = {Valdeolivas, Alberto and Amberg, Bettina and Giroud, Nicolas and Richardson, Marion and G{\ifmmode\acute{a}\else\'{a}\fi}lvez, Eric J. C. and Badillo, Solveig and Julien-Laferri{\ifmmode\grave{e}\else\`{e}\fi}re, Alice and Turos, Demeter and von Voithenberg, Lena Voith and Wells, Isabelle and Lo, Amy A. and Y{\ifmmode\acute{a}\else\'{a}\fi}ng{\ifmmode\ddot{u}\else\"{u}\fi}ez, Emilio and Das Thakur, Meghna and Bscheider, Michael and Sultan, Marc and Kumpesa, Nadine and Jacobsen, Bj{\ifmmode\ddot{o}\else\"{o}\fi}rn and Bergauer, Tobias and Saez-Rodriguez, Julio and Rottenberg, Sven and Schwalie, Petra C. and Hahn, Kerstin},
	title = {{Charting the Heterogeneity of Colorectal Cancer Consensus Molecular Subtypes using Spatial Transcriptomics}},
	journal = {bioRxiv},
	pages = {2023.01.23.525135},
	year = {2023},
	month = jan,
	publisher = {Cold Spring Harbor Laboratory},
	eprint = {2023.01.23.525135},
	url = {https://doi.org/10.1101/2023.01.23.525135}
}

@incollection{kolesnikov2020big,
	author = {Kolesnikov, Alexander and Beyer, Lucas and Zhai, Xiaohua and Puigcerver, Joan and Yung, Jessica and Gelly, Sylvain and Houlsby, Neil},
	title = {{Big Transfer (BiT): General Visual Representation Learning}},
	booktitle = {{Computer Vision {\textendash} ECCV 2020: 16th European Conference, Glasgow, UK, August 23{\textendash}28, 2020, Proceedings, Part V}},
	pages = {491--507},
	isbn = {978-3-030-58557-0},
	publisher = {Springer-Verlag},
	address = {Heidelberg, Germany},
	doi = {10.1007/978-3-030-58558-7_29},
    year={2020}
}

@inproceedings{wang2021transpath,
  title={TransPath: Transformer-Based Self-supervised Learning for Histopathological Image Classification},
  author={Wang, Xiyue and Yang, Sen and Zhang, Jun and Wang, Minghui and Zhang, Jing and Huang, Junzhou and Yang, Wei and Han, Xiao},
  booktitle={International Conference on Medical Image Computing and Computer-Assisted Intervention},
  pages={186--195},
  year={2021},
  organization={Springer}
}

@article{koohbanani2021self,
  title={Self-Path: Self-supervision for Classification of Pathology Images with Limited Annotations},
  author={Koohbanani, Navid Alemi and Unnikrishnan, Balagopal and Khurram, Syed Ali and Krishnaswamy, Pavitra and Rajpoot, Nasir},
  journal={IEEE Transactions on Medical Imaging},
  year={2021},
  publisher={IEEE}
}

@article{li2021from,
	author = {Li, Xinmin and Wang, Cun-Yu},
	title = {{From bulk, single-cell to spatial RNA sequencing}},
	journal = {Int. J. Oral Sci.},
	volume = {13},
	number = {36},
	pages = {1--6},
	year = {2021},
	month = nov,
	issn = {2049-3169},
	publisher = {Nature Publishing Group},
	doi = {10.1038/s41368-021-00146-0}
}

@article{mehra2005identification,
	author = {Mehra, Rohit and Varambally, Sooryanarayana and Ding, Lei and Shen, Ronglai and Sabel, Michael S. and Ghosh, Debashis and Chinnaiyan, Arul M. and Kleer, Celina G.},
	title = {{Identification of GATA3 as a Breast Cancer Prognostic Marker by Global Gene Expression Meta-analysis}},
	journal = {Cancer Res.},
	volume = {65},
	number = {24},
	pages = {11259--11264},
	year = {2005},
	month = dec,
	issn = {0008-5472},
	publisher = {American Association for Cancer Research},
	doi = {10.1158/0008-5472.CAN-05-2495}
}

@inproceedings{ilse2018attention,
  title={Attention-based Deep Multiple Instance Learning},
  author={Ilse, Maximilian and Tomczak, Jakub and Welling, Max},
  booktitle={Proceedings of the 35th International Conference on Machine Learning},
  pages={2132--2141},
  year={2018}
}

@article{lu2020data,
  title={Data Efficient and Weakly Supervised Computational Pathology on Whole Slide Images},
  author={Lu, Ming Y and Williamson, Drew FK and Chen, Tiffany Y and Chen, Richard J and Barbieri, Matteo and Mahmood, Faisal},
  journal={Nature Biomedical Engineering},
  year={2020}
}

@article{kather2020pan,
  title={Pan-cancer image-based detection of clinically actionable genetic alterations},
  author={Kather, Jakob Nikolas and Heij, Lara R and Grabsch, Heike I and Loeffler, Chiara and Echle, Amelie and Muti, Hannah Sophie and Krause, Jeremias and Niehues, Jan M and Sommer, Kai AJ and Bankhead, Peter and others},
  journal={Nature Cancer},
  volume={1},
  number={8},
  pages={789--799},
  year={2020},
  publisher={Nature Publishing Group}
}

@article{huang2023visual,
author = {Huang, Zhi and Bianchi, Federico and Yuksekgonul, Mert and Montine, Thomas and Zou, James},
year = {2023},
month = {08},
pages = {1-10},
title = {A visual–language foundation model for pathology image analysis using medical Twitter},
volume = {29},
journal = {Nature Medicine},
doi = {10.1038/s41591-023-02504-3}
}

@article{brancati2021bracs,
  title={BRACS: A Dataset for BReAst Carcinoma Subtyping in H\&E Histology Images},
  author={Brancati, Nadia and Anniciello, Anna Maria and Pati, Pushpak and Riccio, Daniel and Scognamiglio, Giosu{\`e} and Jaume, Guillaume and De Pietro, Giuseppe and Di Bonito, Maurizio and Foncubierta, Antonio and Botti, Gerardo and others},
  journal={arXiv preprint arXiv:2111.04740},
  year={2021}
}

@article{song2023artificial,
author={Song, Andrew H.
and Jaume, Guillaume
and Williamson, Drew F. K.
and Lu, Ming Y.
and Vaidya, Anurag
and Miller, Tiffany R.
and Mahmood, Faisal},
title={Artificial intelligence for digital and computational pathology},
journal={Nature Reviews Bioengineering},
year={2023},
day={02},
issn={2731-6092},
doi={10.1038/s44222-023-00096-8},
url={https://doi.org/10.1038/s44222-023-00096-8} 
}

@inproceedings{dosovitskiy2021image,
title={An Image is Worth 16x16 Words: Transformers for Image Recognition at Scale},
author={Alexey Dosovitskiy and Lucas Beyer and Alexander Kolesnikov and Dirk Weissenborn and Xiaohua Zhai and Thomas Unterthiner and Mostafa Dehghani and Matthias Minderer and Georg Heigold and Sylvain Gelly and Jakob Uszkoreit and Neil Houlsby},
booktitle={International Conference on Learning Representations},
year={2021},
url={https://openreview.net/forum?id=YicbFdNTTy}
}

@article{ciga2022self,
  title={Self supervised contrastive learning for digital histopathology},
  author={Ciga, Ozan and Xu, Tony and Martel, Anne Louise},
  journal={Machine Learning with Applications},
  volume={7},
  year={2022},
}

@article{chen2024towards,
    title={Towards a General-Purpose Foundation Model for Computational Pathology},
    author={
        Chen, Richard J. and Ding, Tong and Lu, Ming Y. and Williamson, Drew F. K. and Jaume, Guillaume and Song, Andrew H. and Chen, Bowen and Zhang, Andrew and Shao, Daniel and Shaban, Muhammad and Williams, Mane and Oldenburg, Lukas and Weishaupt, Luca L. and Wang, Judy J. and Vaidya, Anurag and Le, Long Phi and Gerber, Georg and Sahai, Sharifa and Williams, Walt and Mahmood, Faisal},
    journal={Nature Medicine},
    year={2024}
}

@article{lu2024towards,
title={A visual-language foundation model for computational pathology},
  author={Lu, Ming Y and Chen, Bowen and Williamson, Drew FK and Chen, Richard J and Liang, Ivy and Ding, Tong and Jaume, Guillaume and Odintsov, Igor and Le, Long Phi and Gerber, Georg and others},
  journal={Nature Medicine},
  volume={30},
  number={3},
  pages={863--874},
  year={2024},
  publisher={Nature Publishing Group US New York}
}

@article{filiot2023scaling,
author = {Filiot, Alexandre and Ghermi, Ridouane and Olivier, Antoine and Jacob, Paul and Fidon, Lucas and Kain, Alice and Saillard, Charlie and Schiratti, Jean-Baptiste},
year = {2023},
month = {07},
pages = {},
title = {Scaling Self-Supervised Learning for Histopathology with Masked Image Modeling},
doi = {10.1101/2023.07.21.23292757},
journal = {medRxiv}
}

@article{vorontsov2024foundation,
  title={A foundation model for clinical-grade computational pathology and rare cancers detection},
  author={Vorontsov, Eugene and Bozkurt, Alican and Casson, Adam and Shaikovski, George and Zelechowski, Michal and Severson, Kristen and Zimmermann, Eric and Hall, James and Tenenholtz, Neil and Fusi, Nicolo and others},
  journal={Nature medicine},
  pages={1--12},
  year={2024},
  publisher={Nature Publishing Group US New York}
}

@inproceedings{gamper2019pannuke,
  title={PanNuke: an open pan-cancer histology dataset for nuclei instance segmentation and classification},
  author={Gamper, Jevgenij and Koohbanani, Navid Alemi and Benet, Ksenija and Khuram, Ali and Rajpoot, Nasir},
  booktitle={European Congress on Digital Pathology},
  pages={11--19},
  year={2019},
  organization={Springer}
}

@article{gamper2020pannuke,
  title={PanNuke Dataset Extension, Insights and Baselines},
  author={Gamper, Jevgenij and Koohbanani, Navid Alemi and Graham, Simon and Jahanifar, Mostafa and Khurram, Syed Ali and Azam, Ayesha and Hewitt, Katherine and Rajpoot, Nasir},
  journal={arXiv preprint arXiv:2003.10778},
  year={2020}
}

@inproceedings{Zimmermann2024Virchow2SS,
  title={Virchow2: Scaling Self-Supervised Mixed Magnification Models in Pathology},
  author={Eric Zimmermann and Eugene Vorontsov and Julian Viret and Adam Casson and Michal Zelechowski and George Shaikovski and Neil Tenenholtz and James Hall and Thomas J Fuchs and Nicol{\`o} Fusi and Siqi Liu and Kristen Severson},
  year={2024}
}

@inproceedings{liu2021swin,
  title={Swin transformer: Hierarchical vision transformer using shifted windows},
  author={Liu, Ze and Lin, Yutong and Cao, Yue and Hu, Han and Wei, Yixuan and Zhang, Zheng and Lin, Stephen and Guo, Baining},
  booktitle={Proceedings of the IEEE/CVF International Conference on Computer Vision},
  pages={10012--10022},
  year={2021}
}

@inproceedings{jaume2024transcriptomics,
  title={Transcriptomics-guided Slide Representation Learning in Computational Pathology},
  author={Jaume, Guillaume and Oldenburg, Lukas and Vaidya, Anurag Jayant and Chen, Richard J. and Williamson, Drew FK and Peeters, Thomas and Song, Andrew H. and Mahmood, Faisal},
  booktitle={Proceedings of the IEEE/CVF Conference on Computer Vision and Pattern Recognition (CVPR)},
  year={2024}
}

@inproceedings{jaume2024multistain,
  title={Multistain Pretraining for Slide Representation Learning in Pathology},
  author={Jaume, Guillaume and Vaidya, Anurag and Zhang, Andrew and Song, Andrew and Chen, Richard and Sahai, Sharifa and Mo, Dandan and Madrigal, Emilio and Le, Long and Mahmood, Faisal},
  booktitle = {Proceedings of the European Conference on Computer Vision (ECCV)},
  year={2024},
  organization={Springer}
}

@article{lu2024multimodal,
	author = {Lu, Ming Y. and Chen, Bowen and Williamson, Drew F. K. and Chen, Richard J. and Zhao, Melissa and Chow, Aaron K. and Ikemura, Kenji and Kim, Ahrong and Pouli, Dimitra and Patel, Ankush and Soliman, Amr and Chen, Chengkuan and Ding, Tong and Wang, Judy J. and Gerber, Georg and Liang, Ivy and Le, Long Phi and Parwani, Anil V. and Weishaupt, Luca L. and Mahmood, Faisal},
	title = {{A Multimodal Generative AI Copilot for Human Pathology}},
	journal = {Nature},
	pages = {1--3},
	year = {2024},
	month = jun,
	issn = {1476-4687},
	publisher = {Nature Publishing Group},
	doi = {10.1038/s41586-024-07618-3}
}

@article{fu2020pan,
  title={Pan-cancer computational histopathology reveals mutations, tumor composition and prognosis},
  author={Fu, Yu and Jung, Alexander W and Torne, Ramon Vi{\~n}as and Gonzalez, Santiago and V{\"o}hringer, Harald and Shmatko, Artem and Yates, Lucy R and Jimenez-Linan, Mercedes and Moore, Luiza and Gerstung, Moritz},
  journal={Nature Cancer},
  volume={1},
  number={8},
  pages={800--810},
  year={2020},
  publisher={Nature Publishing Group}
}

@article{bejnordi2017diagnostic,
  title={Diagnostic assessment of deep learning algorithms for detection of lymph node metastases in women with breast cancer},
  author={Bejnordi, Babak Ehteshami and Veta, Mitko and Van Diest, Paul Johannes and Van Ginneken, Bram and Karssemeijer, Nico and Litjens, Geert and Van Der Laak, Jeroen AWM and Hermsen, Meyke and Manson, Quirine F and Balkenhol, Maschenka and others},
  journal={JAMA},
  volume={318},
  number={22},
  pages={2199--2210},
  year={2017},
  publisher={American Medical Association}
}

@article{barbano2021unitopatho,
	author = {Barbano, Carlo Alberto and Perlo, Daniele and Tartaglione, Enzo and Fiandrotti, Attilio and Bertero, Luca and Cassoni, Paola and Grangetto, Marco},
	title = {{UniToPatho, a labeled histopathological dataset for colorectal polyps classification and adenoma dysplasia grading}},
	journal = {arXiv},
	year = {2021},
	month = jan,
	eprint = {2101.09991},
	doi = {10.1109/ICIP42928.2021.9506198}
}

@article{kather2019predicting,
	author = {Kather, Jakob Nikolas and Krisam, Johannes and Charoentong, Pornpimol and Luedde, Tom and Herpel, Esther and Weis, Cleo-Aron and Gaiser, Timo and Marx, Alexander and Valous, Nektarios A. and Ferber, Dyke and Jansen, Lina and Reyes-Aldasoro, Constantino Carlos and Z{\ifmmode\ddot{o}\else\"{o}\fi}rnig, Inka and J{\ifmmode\ddot{a}\else\"{a}\fi}ger, Dirk and Brenner, Hermann and Chang-Claude, Jenny and Hoffmeister, Michael and Halama, Niels},
	title = {{Predicting survival from colorectal cancer histology slides using deep learning: A retrospective multicenter study}},
	journal = {PLoS Med.},
	volume = {16},
	number = {1},
	year = {2019},
	month = jan,
	publisher = {PLOS},
	doi = {10.1371/journal.pmed.1002730}
}

@article{pataki2022huncrc,
	author = {Pataki, B{\ifmmode\acute{a}\else\'{a}\fi}lint {\ifmmode\acute{A}\else\'{A}\fi}rmin and Olar, Alex and Ribli, Dezs{\ifmmode\mbox{\H{o}}\else\H{o}\fi} and Pesti, Adri{\ifmmode\acute{a}\else\'{a}\fi}n and Kontsek, Endre and Gy{\ifmmode\ddot{o}\else\"{o}\fi}ngy{\ifmmode\ddot{o}\else\"{o}\fi}si, Benedek and Bilecz, {\ifmmode\acute{A}\else\'{A}\fi}gnes and Kov{\ifmmode\acute{a}\else\'{a}\fi}cs, Tekla and Kov{\ifmmode\acute{a}\else\'{a}\fi}cs, Krist{\ifmmode\acute{o}\else\'{o}\fi}f Attila and Kramer, Zs{\ifmmode\acute{o}\else\'{o}\fi}fia and Kiss, Andr{\ifmmode\acute{a}\else\'{a}\fi}s and Sz{\ifmmode\acute{o}\else\'{o}\fi}cska, Mikl{\ifmmode\acute{o}\else\'{o}\fi}s and Pollner, P{\ifmmode\acute{e}\else\'{e}\fi}ter and Csabai, Istv{\ifmmode\acute{a}\else\'{a}\fi}n},
	title = {{HunCRC: annotated pathological slides to enhance deep learning applications in colorectal cancer screening}},
	journal = {Sci. Data},
	volume = {9},
	number = {370},
	pages = {1--7},
	year = {2022},
	month = jun,
	issn = {2052-4463},
	publisher = {Nature Publishing Group},
	doi = {10.1038/s41597-022-01450-y}
}

@inproceedings{wei2021petri,
title={A Petri Dish for Histopathology Image Analysis},
author={Wei, Jerry and Suriawinata, Arief and Ren, Bing and Liu, Xiaoying and Lisovsky, Mikhail and Vaickus, Louis and Brown, Charles and Baker, Michael and Tomita, Naofumi and Torresani, Lorenzo and others},
booktitle={International Conference on Artificial Intelligence in Medicine},
pages={11--24},
year={2021},
organization={Springer}
}

@article{silvarodriguez2020going,
	author = {Silva-Rodr{\ifmmode\acute{\imath}\else\'{\i}\fi}guez, Julio and Colomer, Adri{\ifmmode\acute{a}\else\'{a}\fi}n and Sales, Mar{\ifmmode\acute{\imath}\else\'{\i}\fi}a A. and Molina, Rafael and Naranjo, Valery},
	title = {{Going deeper through the Gleason scoring scale: An automatic end-to-end system for histology prostate grading and cribriform pattern detection}},
	journal = {Comput. Methods Programs Biomed.},
	volume = {195},
	pages = {105637},
	year = {2020},
	month = oct,
	issn = {0169-2607},
	publisher = {Elsevier},
	doi = {10.1016/j.cmpb.2020.105637}
}

@misc{huo2022comprehensive,
	author = {Huo, Xinmi and Ong, Kok Haur and Lau, Kah Weng and Gole, Laurent and Tan, Char Loo and Zhang, Chongchong and Zhang, Yonghui and Zhu, Xiaohui and Li, Longjie and Han, Hao and Young, David and Lu, Haoda and Xu, Jun and Chen, Wanyuan and Sanders, Stephan J. and Kuan, Lee Hwee and Hue, Susan Swee-Shan and Yu, Weimiao and Tan, Soo Yong},
	title = {{Comprehensive AI Model Development for Gleason Grading: From Scanning, Cloud-Based Annotation to Pathologist-AI Interaction}},
	year = {2022},
	month = jul,
	note = {[Online; accessed 7. May 2024]},
	doi = {10.2139/ssrn.4172090}
}

@article{koziarski2024diagset,
	author = {Koziarski, Micha{\l} and Cyganek, Bogus{\l}aw and Niedziela, Przemys{\l}aw and Olborski, Bogus{\l}aw and Antosz, Zbigniew and {\ifmmode\dot{Z}\else\.{Z}\fi}ydak, Marcin and Kwolek, Bogdan and W{\k{a}}sowicz, Pawe{\l} and Buka{\l}a, Andrzej and Swad{\ifmmode\acute{z}\else\'{z}\fi}ba, Jakub and Sitkowski, Piotr},
	title = {{DiagSet: a dataset for prostate cancer histopathological image classification}},
	journal = {Sci. Rep.},
	volume = {14},
	number = {6780},
	pages = {1--14},
	year = {2024},
	month = mar,
	issn = {2045-2322},
	publisher = {Nature Publishing Group},
	doi = {10.1038/s41598-024-52183-4}
}

@inproceedings{gamper2021multiple,
  title={Multiple Instance Captioning: Learning Representations from Histopathology Textbooks and Articles},
  author={Gamper, Jevgenij and Rajpoot, Nasir},
  booktitle={Proceedings of the IEEE/CVF Conference on Computer Vision and Pattern Recognition},
  year={2021}
}

@article{lu2021ai,
  title={AI-based pathology predicts origins for cancers of unknown primary},
  author={Lu, Ming Y and Chen, Tiffany Y and Williamson, Drew FK and Zhao, Melissa and Shady, Maha and Lipkova, Jana and Mahmood, Faisal},
  journal={Nature},
  volume={594},
  number={7861},
  pages={106--110},
  year={2021},
  publisher={Nature Publishing Group}
}

@article{shao2021transmil,
  title={Transmil: Transformer based correlated multiple instance learning for whole slide image classification},
  author={Shao, Zhuchen and Bian, Hao and Chen, Yang and Wang, Yifeng and Zhang, Jian and Ji, Xiangyang and others},
  journal={Advances in Neural Information Processing Systems},
  volume={34},
  year={2021}
}

@article{zhou2021ibot,
  title={iBOT: Image BERT Pre-Training with Online Tokenizer},
  author={Zhou, Jinghao and Wei, Chen and Wang, Huiyu and Shen, Wei and Xie, Cihang and Yuille, Alan and Kong, Tao},
  journal={International Conference on Learning Representations (ICLR)},
  year={2022}
}

@article{kather2019deep,
  title={Deep learning can predict microsatellite instability directly from histology in gastrointestinal cancer},
  author={Kather, Jakob Nikolas and Pearson, Alexander T and Halama, Niels and J{\"a}ger, Dirk and Krause, Jeremias and Loosen, Sven H and Marx, Alexander and Boor, Peter and Tacke, Frank and Neumann, Ulf Peter and others},
  journal={Nature medicine},
  volume={25},
  number={7},
  pages={1054--1056},
  year={2019},
  publisher={Nature Publishing Group US New York}
}

@article{echle2021deep,
  title={Deep learning in cancer pathology: a new generation of clinical biomarkers},
  author={Echle, Amelie and Rindtorff, Niklas Timon and Brinker, Titus Josef and Luedde, Tom and Pearson, Alexander Thomas and Kather, Jakob Nikolas},
  journal={British journal of cancer},
  volume={124},
  number={4},
  pages={686--696},
  year={2021},
  publisher={Nature Publishing Group UK London}
}

@article{lipkova2022deep,
  title={Deep learning-enabled assessment of cardiac allograft rejection from endomyocardial biopsies},
  author={Lipkova, Jana and Chen, Tiffany Y and Lu, Ming Y and Chen, Richard J and Shady, Maha and Williams, Mane and Wang, Jingwen and Noor, Zahra and Mitchell, Richard N and Turan, Mehmet and others},
  journal={Nature medicine},
  volume={28},
  number={3},
  pages={575--582},
  year={2022},
  publisher={Nature Publishing Group US New York}
}

@article{azizi2023robust,
  title={Robust and data-efficient generalization of self-supervised machine learning for diagnostic imaging},
  author={Azizi, Shekoofeh and Culp, Laura and Freyberg, Jan and Mustafa, Basil and Baur, Sebastien and Kornblith, Simon and Chen, Ting and Tomasev, Nenad and Mitrovi{\'c}, Jovana and Strachan, Patricia and others},
  journal={Nature Biomedical Engineering},
  pages={1--24},
  year={2023},
  publisher={Nature Publishing Group UK London}
}

@article{xu2024wholeslide,
	author = {Xu, Hanwen and Usuyama, Naoto and Bagga, Jaspreet and Zhang, Sheng and Rao, Rajesh and Naumann, Tristan and Wong, Cliff and Gero, Zelalem and Gonz{\ifmmode\acute{a}\else\'{a}\fi}lez, Javier and Gu, Yu and Xu, Yanbo and Wei, Mu and Wang, Wenhui and Ma, Shuming and Wei, Furu and Yang, Jianwei and Li, Chunyuan and Gao, Jianfeng and Rosemon, Jaylen and Bower, Tucker and Lee, Soohee and Weerasinghe, Roshanthi and Wright, Bill J. and Robicsek, Ari and Piening, Brian and Bifulco, Carlo and Wang, Sheng and Poon, Hoifung},
	title = {{A whole-slide foundation model for digital pathology from real-world data}},
	journal = {Nature},
	volume = {630},
	pages = {181--188},
	year = {2024},
	month = jun,
	issn = {1476-4687},
	publisher = {Nature Publishing Group},
	doi = {10.1038/s41586-024-07441-w}
}

@article{chen2017rethinking,
	author = {Chen, Liang-Chieh and Papandreou, George and Schroff, Florian and Adam, Hartwig},
	title = {{Rethinking Atrous Convolution for Semantic Image Segmentation}},
	journal = {arXiv},
	year = {2017},
	month = jun,
	eprint = {1706.05587},
	doi = {10.48550/arXiv.1706.05587}
}

@article{rao2021exploring,
	author = {Rao, Anjali and Barkley, Dalia and Fran{\ifmmode\mbox{\c{c}}\else\c{c}\fi}a, Gustavo S. and Yanai, Itai},
	title = {{Exploring tissue architecture using spatial transcriptomics}},
	journal = {Nature},
	volume = {596},
	pages = {211--220},
	year = {2021},
	month = aug,
	issn = {1476-4687},
	publisher = {Nature Publishing Group},
	doi = {10.1038/s41586-021-03634-9}
}

@article{bulten2022artificial,
  title={Artificial intelligence for diagnosis and Gleason grading of prostate cancer: the PANDA challenge},
  author={Bulten, Wouter and Kartasalo, Kimmo and Chen, Po-Hsuan Cameron and Str{\"o}m, Peter and Pinckaers, Hans and Nagpal, Kunal and Cai, Yuannan and Steiner, David F and van Boven, Hester and Vink, Robert and others},
  journal={Nature Medicine},
  volume={28},
  number={1},
  pages={154--163},
  year={2022},
  publisher={Nature Publishing Group}
}

@article{chen2022pan,
  title={Pan-cancer integrative histology-genomic analysis via multimodal deep learning},
  author={Chen, Richard J and Lu, Ming Y and Williamson, Drew FK and Chen, Tiffany Y and Lipkova, Jana and Noor, Zahra and Shaban, Muhammad and Shady, Maha and Williams, Mane and Joo, Bumjin and others},
  journal={Cancer Cell},
  volume={40},
  number={8},
  pages={865--878},
  year={2022},
  publisher={Elsevier}
}

@inproceedings{kang2023benchmarking,
  title={Benchmarking Self-Supervised Learning on Diverse Pathology Datasets},
  author={Kang, Mingu and Song, Heon and Park, Seonwook and Yoo, Donggeun and Pereira, S{\'e}rgio},
  booktitle={Proceedings of the IEEE/CVF Conference on Computer Vision and Pattern Recognition},
  pages={3344--3354},
  year={2023}
}

@article{haghverdi2018mnn,
	author = {Haghverdi, Laleh and Lun, Aaron T. L. and Morgan, Michael D. and Marioni, John C.},
	title = {{Batch effects in single-cell RNA-sequencing data are corrected by matching mutual nearest neighbors}},
	journal = {Nat. Biotechnol.},
	volume = {36},
	pages = {421--427},
	year = {2018},
	month = may,
	issn = {1546-1696},
	publisher = {Nature Publishing Group},
	doi = {10.1038/nbt.4091}
}

@article{zhang2020combat,
	author = {Zhang, Yuqing and Parmigiani, Giovanni and Johnson, W. Evan},
	title = {{ComBat-seq: batch effect adjustment for RNA-seq count data}},
	journal = {NAR Genomics Bioinf.},
	volume = {2},
	number = {3},
	pages = {lqaa078},
	year = {2020},
	month = sep,
	issn = {2631-9268},
	publisher = {Oxford Academic}
}

@article{korsunsky2019harmony,
	author = {Korsunsky, Ilya and Millard, Nghia and Fan, Jean and Slowikowski, Kamil and Zhang, Fan and Wei, Kevin and Baglaenko, Yuriy and Brenner, Michael and Loh, Po-ru and Raychaudhuri, Soumya},
	title = {{Fast, sensitive and accurate integration of single-cell data with Harmony}},
	journal = {Nat. Methods},
	volume = {16},
	pages = {1289--1296},
	year = {2019},
	month = dec,
	issn = {1548-7105},
	publisher = {Nature Publishing Group},
	doi = {10.1038/s41592-019-0619-0}
}

@article{oquab2023dinov2,
  title={Dinov2: Learning robust visual features without supervision},
  author={Oquab, Maxime and Darcet, Timoth{\'e}e and Moutakanni, Th{\'e}o and Vo, Huy and Szafraniec, Marc and Khalidov, Vasil and Fernandez, Pierre and Haziza, Daniel and Massa, Francisco and El-Nouby, Alaaeldin and others},
  journal={arXiv preprint arXiv:2304.07193},
  year={2023}
}

@InProceedings{lu2023visual,
    author    = {Lu, Ming Y. and Chen, Bowen and Zhang, Andrew and Williamson, Drew F. K. and Chen, Richard J. and Ding, Tong and Le, Long Phi and Chuang, Yung-Sung and Mahmood, Faisal},
    title     = {Visual Language Pretrained Multiple Instance Zero-Shot Transfer for Histopathology Images},
    booktitle = {Proceedings of the IEEE/CVF Conference on Computer Vision and Pattern Recognition (CVPR)},
    year      = {2023},
    pages     = {19764-19775}
}

@article{wagner2023fully,
title = {Transformer-based biomarker prediction from colorectal cancer histology: A large-scale multicentric study},
  volume = {41},
  ISSN = {1535-6108},
  number = {9},
  journal = {Cancer Cell},
  publisher = {Elsevier BV},
  author = {Wagner,  Sophia J. and Reisenb\"{u}chler,  Daniel and West,  Nicholas P. and Niehues,  Jan Moritz and Zhu,  Jiefu and Foersch,  Sebastian and Veldhuizen,  Gregory Patrick and Quirke,  Philip and Grabsch,  Heike I. and van den Brandt,  Piet A. and Hutchins,  Gordon G.A. and Richman,  Susan D. and Yuan,  Tanwei and Langer,  Rupert and Jenniskens,  Josien C.A. and Offermans,  Kelly and Mueller,  Wolfram and Gray,  Richard and Gruber,  Stephen B. and Greenson,  Joel K. and Rennert,  Gad and Bonner,  Joseph D. and Schmolze,  Daniel and Jonnagaddala,  Jitendra and Hawkins,  Nicholas J. and Ward,  Robyn L. and Morton,  Dion and Seymour,  Matthew and Magill,  Laura and Nowak,  Marta and Hay,  Jennifer and Koelzer,  Viktor H. and Church,  David N. and Matek,  Christian and Geppert,  Carol and Peng,  Chaolong and Zhi,  Cheng and Ouyang,  Xiaoming and James,  Jacqueline A. and Loughrey,  Maurice B. and Salto-Tellez,  Manuel and Brenner,  Hermann and Hoffmeister,  Michael and Truhn,  Daniel and Schnabel,  Julia A. and Boxberg,  Melanie and Peng,  Tingying and Kather,  Jakob Nikolas and Church,  David and Domingo,  Enric and Edwards,  Joanne and Glimelius,  Bengt and Gogenur,  Ismail and Harkin,  Andrea and Hay,  Jen and Iveson,  Timothy and Jaeger,  Emma and Kelly,  Caroline and Kerr,  Rachel and Maka,  Noori and Morgan,  Hannah and Oien,  Karin and Orange,  Clare and Palles,  Claire and Roxburgh,  Campbell and Sansom,  Owen and Saunders,  Mark and Tomlinson,  Ian},
  year = {2023},
  month = sep,
}

@article{bergenstrahle2022super,
	author = {Bergenstr{\aa}hle, Ludvig and He, Bryan and Bergenstr{\aa}hle, Joseph and Abalo, Xes{\ifmmode\acute{u}\else\'{u}\fi}s and Mirzazadeh, Reza and Thrane, Kim and Ji, Andrew L. and Andersson, Alma and Larsson, Ludvig and Stakenborg, Nathalie and Boeckxstaens, Guy and Khavari, Paul and Zou, James and Lundeberg, Joakim and Maaskola, Jonas},
	title = {{Super-resolved spatial transcriptomics by deep data fusion}},
	journal = {Nat. Biotechnol.},
	volume = {40},
	pages = {476--479},
	year = {2022},
	month = apr,
	issn = {1546-1696},
	publisher = {Nature Publishing Group},
	doi = {10.1038/s41587-021-01075-3}
}

@article{zhang2024inferring,
	author = {Zhang, Daiwei and Schroeder, Amelia and Yan, Hanying and Yang, Haochen and Hu, Jian and Lee, Michelle Y. Y. and Cho, Kyung S. and Susztak, Katalin and Xu, George X. and Feldman, Michael D. and Lee, Edward B. and Furth, Emma E. and Wang, Linghua and Li, Mingyao},
	title = {{Inferring super-resolution tissue architecture by integrating spatial transcriptomics with histology}},
	journal = {Nat. Biotechnol.},
	pages = {1--6},
	year = {2024},
	month = jan,
	issn = {1546-1696},
	publisher = {Nature Publishing Group},
	doi = {10.1038/s41587-023-02019-9}
}

@article{wang2024m2ort,
	author = {Wang, Hongyi and Du, Xiuju and Liu, Jing and Ouyang, Shuyi and Chen, Yen-Wei and Lin, Lanfen},
	title = {{M2ORT: Many-To-One Regression Transformer for Spatial Transcriptomics Prediction from Histopathology Images}},
	journal = {arXiv},
	year = {2024},
	month = jan,
	eprint = {2401.10608},
	doi = {10.48550/arXiv.2401.10608}
}

@article{pang2021leveraging,
	author = {Pang, Minxing and Su, Kenong and Li, Mingyao},
	title = {{Leveraging information in spatial transcriptomics to predict super-resolution gene expression from histology images in tumors}},
	journal = {bioRxiv},
	pages = {2021.11.28.470212},
	year = {2021},
	month = nov,
	publisher = {Cold Spring Harbor Laboratory},
	eprint = {2021.11.28.470212},
	url = {https://doi.org/10.1101/2021.11.28.470212}
}

@inproceedings{xie2023spatially,
 author = {Xie, Ronald and Pang, Kuan and Chung, Sai and Perciani, Catia and MacParland, Sonya and Wang, Bo and Bader, Gary},
 booktitle = {Advances in Neural Information Processing Systems},
 editor = {A. Oh and T. Neumann and A. Globerson and K. Saenko and M. Hardt and S. Levine},
 pages = {70626--70637},
 publisher = {Curran Associates, Inc.},
 title = {Spatially Resolved Gene Expression Prediction from Histology Images via Bi-modal Contrastive Learning},
 url = {https://proceedings.neurips.cc/paper_files/paper/2023/file/df656d6ed77b565e8dcdfbf568aead0a-Paper-Conference.pdf},
 volume = {36},
 year = {2023}
}

@inproceedings{chung2024accurate,
  title={Accurate Spatial Gene Expression Prediction by integrating Multi-resolution features},
  author={Chung, Youngmin and Ha, Ji Hun and Im, Kyeong Chan and Lee, Joo Sang},
  booktitle={Proceedings of the IEEE/CVF Conference on Computer Vision and Pattern Recognition},
  pages={11591--11600},
  year={2024}
}

@article{loeffler2022artificial,
  title={Artificial Intelligence--based Detection of FGFR3 Mutational Status Directly from Routine Histology in Bladder Cancer: A Possible Preselection for Molecular Testing?},
  author={Loeffler, Chiara Maria Lavinia and Bruechle, Nadina Ortiz and Jung, Max and Seillier, Lancelot and Rose, Michael and Laleh, Narmin Ghaffari and Knuechel, Ruth and Brinker, Titus J and Trautwein, Christian and Gaisa, Nadine T and others},
  journal={European Urology Focus},
  volume={8},
  number={2},
  pages={472--479},
  year={2022},
  publisher={Elsevier}
}

@article{wang2019predicting,
  title={Predicting EGFR mutation status in lung adenocarcinoma on computed tomography image using deep learning},
  author={Wang, Shuo and Shi, Jingyun and Ye, Zhaoxiang and Dong, Di and Yu, Dongdong and Zhou, Mu and Liu, Ying and Gevaert, Olivier and Wang, Kun and Zhu, Yongbei and others},
  journal={European Respiratory Journal},
  volume={53},
  number={3},
  year={2019},
  publisher={Eur Respiratory Soc}
}

@article{saldanha2023self,
  title={Self-supervised attention-based deep learning for pan-cancer mutation prediction from histopathology},
  author={Saldanha, Oliver Lester and Loeffler, Chiara ML and Niehues, Jan Moritz and van Treeck, Marko and Seraphin, Tobias P and Hewitt, Katherine Jane and Cifci, Didem and Veldhuizen, Gregory Patrick and Ramesh, Siddhi and Pearson, Alexander T and others},
  journal={NPJ Precision Oncology},
  volume={7},
  number={1},
  pages={35},
  year={2023},
  publisher={Nature Publishing Group UK London}
}

@article{schmauch2020deep,
  title={A deep learning model to predict RNA-Seq expression of tumours from whole slide images},
  author={Schmauch, Beno{\^\i}t and Romagnoni, Alberto and Pronier, Elodie and Saillard, Charlie and Maill{\'e}, Pascale and Calderaro, Julien and Kamoun, Aur{\'e}lie and Sefta, Meriem and Toldo, Sylvain and Zaslavskiy, Mikhail and others},
  journal={Nature Communications},
  volume={11},
  number={1},
  year={2020},
  publisher={Nature Publishing Group}
}

@article{mondol2023hist2rna,
	author = {Mondol, Raktim Kumar and Millar, Ewan K. A. and Graham, Peter H. and Browne, Lois and Sowmya, Arcot and Meijering, Erik},
	title = {{hist2RNA: An Efficient Deep Learning Architecture to Predict Gene Expression from Breast Cancer Histopathology Images}},
	journal = {Cancers},
	volume = {15},
	number = {9},
	pages = {2569},
	year = {2023},
	month = apr,
	issn = {2072-6694},
	publisher = {Multidisciplinary Digital Publishing Institute},
	doi = {10.3390/cancers15092569}
}

@article{monjo2022efficient,
	author = {Monjo, Taku and Koido, Masaru and Nagasawa, Satoi and Suzuki, Yutaka and Kamatani, Yoichiro},
	title = {{Efficient prediction of a spatial transcriptomics profile better characterizes breast cancer tissue sections without costly experimentation}},
	journal = {Sci. Rep.},
	volume = {12},
	number = {4133},
	pages = {1--12},
	year = {2022},
	month = mar,
	issn = {2045-2322},
	publisher = {Nature Publishing Group},
	doi = {10.1038/s41598-022-07685-4}
}

@article{rahaman2023breast,
	author = {Rahaman, Md Mamunur and Millar, Ewan K. A. and Meijering, Erik},
	title = {{Breast cancer histopathology image-based gene expression prediction using spatial transcriptomics data and deep learning}},
	journal = {Sci. Rep.},
	volume = {13},
	number = {13604},
	pages = {1--11},
	year = {2023},
	month = aug,
	issn = {2045-2322},
	publisher = {Nature Publishing Group},
	doi = {10.1038/s41598-023-40219-0}
}

@article{el2024regression,
  title={Regression-based Deep-Learning predicts molecular biomarkers from pathology slides},
  author={El Nahhas, Omar SM and Loeffler, Chiara ML and Carrero, Zunamys I and van Treeck, Marko and Kolbinger, Fiona R and Hewitt, Katherine J and Muti, Hannah S and Graziani, Mara and Zeng, Qinghe and Calderaro, Julien and others},
  journal={Nature communications},
  volume={15},
  number={1},
  pages={1253},
  year={2024},
  publisher={Nature Publishing Group UK London}
}

@article{he2020integrating,
  title={Integrating spatial gene expression and breast tumour morphology via deep learning},
  author={He, Bryan and Bergenstr{\aa}hle, Ludvig and Stenbeck, Linnea and Abid, Abubakar and Andersson, Alma and Borg, {\AA}ke and Maaskola, Jonas and Lundeberg, Joakim and Zou, James},
  journal={Nature biomedical engineering},
  volume={4},
  number={8},
  pages={827--834},
  year={2020},
  publisher={Nature Publishing Group UK London}
}

@article{alsaafin2023learning,
  title={Learning to predict RNA sequence expressions from whole slide images with applications for search and classification},
  author={Alsaafin, Areej and Safarpoor, Amir and Sikaroudi, Milad and Hipp, Jason D and Tizhoosh, HR},
  journal={Communications Biology},
  volume={6},
  number={1},
  pages={304},
  year={2023},
  publisher={Nature Publishing Group UK London}
}

@article{zhao2024hist2cell,
	author = {Zhao, Weiqin and Liang, Zhuo and Huang, Xianjie and Huang, Yuanhua and Yu, Lequan},
	title = {{Hist2Cell: Deciphering Fine-grained Cellular Architectures from Histology Images}},
	journal = {bioRxiv},
	pages = {2024.02.17.580852},
	year = {2024},
	month = feb,
	publisher = {Cold Spring Harbor Laboratory}
}

@article{asp2020spatially,
	author = {Asp, Michaela and Bergenstr{\aa}hle, Joseph and Lundeberg, Joakim},
	title = {{Spatially Resolved Transcriptomes{\ifmmode---\else\textemdash\fi}Next Generation Tools for Tissue Exploration}},
	journal = {BioEssays},
	volume = {42},
	number = {10},
	pages = {1900221},
	year = {2020},
	month = oct,
	issn = {0265-9247},
	publisher = {John Wiley {\&} Sons, Ltd},
	doi = {10.1002/bies.201900221}
}

@article{chen2023giotto,
	author = {Chen, Jiaji George and Ch{\ifmmode\acute{a}\else\'{a}\fi}vez-Fuentes, Joselyn Cristina and O{'}Brien, Matthew and Xu, Junxiang and Ruiz, Edward and Wang, Wen and Amin, Iqra and Sarfraz, Irzam and Guckhool, Pratishtha and Sistig, Adriana and Yuan, Guo-Cheng and Dries, Ruben},
	title = {{Giotto Suite: a multi-scale and technology-agnostic spatial multi-omics analysis ecosystem}},
	journal = {bioRxiv},
	pages = {2023.11.26.568752},
	year = {2023},
	month = nov,
	publisher = {Cold Spring Harbor Laboratory},
	eprint = {2023.11.26.568752},
	url = {https://doi.org/10.1101/2023.11.26.568752}
}

@article{zhao2021spatial,
	author = {Zhao, Edward and Stone, Matthew R. and Ren, Xing and Guenthoer, Jamie and Smythe, Kimberly S. and Pulliam, Thomas and Williams, Stephen R. and Uytingco, Cedric R. and Taylor, Sarah E. B. and Nghiem, Paul and Bielas, Jason H. and Gottardo, Raphael},
	title = {{Spatial transcriptomics at subspot resolution with BayesSpace}},
	journal = {Nat. Biotechnol.},
	volume = {39},
	pages = {1375--1384},
	year = {2021},
	month = nov,
	issn = {1546-1696},
	publisher = {Nature Publishing Group}
}

@article{pham2023robust,
	author = {Pham, Duy and Tan, Xiao and Balderson, Brad and Xu, Jun and Grice, Laura F. and Yoon, Sohye and Willis, Emily F. and Tran, Minh and Lam, Pui Yeng and Raghubar, Arti and Kalita-de Croft, Priyakshi and Lakhani, Sunil and Vukovic, Jana and Ruitenberg, Marc J. and Nguyen, Quan H.},
	title = {{Robust mapping of spatiotemporal trajectories and cell{\textendash}cell interactions in healthy and diseased tissues}},
	journal = {Nat. Commun.},
	volume = {14},
	number = {7739},
	pages = {1--25},
	year = {2023},
	month = nov,
	issn = {2041-1723},
	publisher = {Nature Publishing Group},
	doi = {10.1038/s41467-023-43120-6}
}

@article{veeling2018rotation,
  title         = "Rotation Equivariant {CNNs} for Digital Pathology",
  author        = "Veeling, Bastiaan S and Linmans, Jasper and Winkens, Jim and
                   Cohen, Taco and Welling, Max",
  month         =  jun,
  year          =  2018,
  archivePrefix = "arXiv",
  primaryClass  = "cs.CV",
  eprint        = "1806.03962"
}

@article{spanhol2015dataset,
	author = {Spanhol, Fabio A. and Oliveira, Luiz S. and Petitjean, Caroline and Heutte, Laurent},
	title = {{A Dataset for Breast Cancer Histopathological Image Classification}},
	journal = {IEEE Trans. Biomed. Eng.},
	volume = {63},
	number = {7},
	pages = {1455--1462},
	year = {2015},
	month = oct,
	publisher = {IEEE},
	doi = {10.1109/TBME.2015.2496264}
}

@incollection{redmon2016you,
	author = {Redmon, Joseph and Divvala, Santosh and Girshick, Ross and Farhadi, Ali},
	title = {{You Only Look Once: Unified, Real-Time Object Detection}},
	booktitle = {{2016 IEEE Conference on Computer Vision and Pattern Recognition (CVPR)}},
	journal = {Published in: 2016 IEEE Conference on Computer Vision and Pattern Recognition (CVPR)},
	pages = {27--30},
	issn = {1063-6919},
	publisher = {IEEE},
    year = {2016},
	doi = {10.1109/CVPR.2016.91}
}

@article{palla2022squidpy,
	author = {Palla, Giovanni and Spitzer, Hannah and Klein, Michal and Fischer, David and Schaar, Anna Christina and Kuemmerle, Louis Benedikt and Rybakov, Sergei and Ibarra, Ignacio L. and Holmberg, Olle and Virshup, Isaac and Lotfollahi, Mohammad and Richter, Sabrina and Theis, Fabian J.},
	title = {{Squidpy: a scalable framework for spatial omics analysis}},
	journal = {Nat. Methods},
	volume = {19},
	pages = {171--178},
	year = {2022},
	month = feb,
	issn = {1548-7105},
	publisher = {Nature Publishing Group},
	doi = {10.1038/s41592-021-01358-2}
}

@article{butler2018integrating,
	author = {Butler, Andrew and Hoffman, Paul and Smibert, Peter and Papalexi, Efthymia and Satija, Rahul},
	title = {{Integrating single-cell transcriptomic data across different conditions, technologies, and species}},
	journal = {Nat. Biotechnol.},
	volume = {36},
	pages = {411--420},
	year = {2018},
	month = may,
	issn = {1546-1696},
	publisher = {Nature Publishing Group},
	doi = {10.1038/nbt.4096}
}

@article{wolf2018scanpy,
	author = {Wolf, F. Alexander and Angerer, Philipp and Theis, Fabian J.},
	title = {{SCANPY: large-scale single-cell gene expression data analysis}},
	journal = {Genome Biol.},
	volume = {19},
	number = {1},
	pages = {1--5},
	year = {2018},
	month = dec,
	issn = {1474-760X},
	publisher = {BioMed Central},
	doi = {10.1186/s13059-017-1382-0}
}

@article{marconato2024spatialdata,
	author = {Marconato, Luca and Palla, Giovanni and Yamauchi, Kevin A. and Virshup, Isaac and Heidari, Elyas and Treis, Tim and Vierdag, Wouter-Michiel and Toth, Marcella and Stockhaus, Sonja and Shrestha, Rahul B. and Rombaut, Benjamin and Pollaris, Lotte and Lehner, Laurens and V{\ifmmode\ddot{o}\else\"{o}\fi}hringer, Harald and Kats, Ilia and Saeys, Yvan and Saka, Sinem K. and Huber, Wolfgang and Gerstung, Moritz and Moore, Josh and Theis, Fabian J. and Stegle, Oliver},
	title = {{SpatialData: an open and universal data framework for spatial omics}},
	journal = {Nat. Methods},
	pages = {1--5},
	year = {2024},
	month = mar,
	issn = {1548-7105},
	publisher = {Nature Publishing Group},
	doi = {10.1038/s41592-024-02212-x}
}

@article{maynard2021transcriptome,
  title={Transcriptome-scale spatial gene expression in the human dorsolateral prefrontal cortex},
  author={Maynard, Kristen R and Collado-Torres, Leonardo and Weber, Lukas M and Uytingco, Cedric and Barry, Brianna K and Williams, Stephen R and Catallini, Joseph L and Tran, Matthew N and Besich, Zachary and Tippani, Madhavi and others},
  journal={Nature neuroscience},
  volume={24},
  number={3},
  pages={425--436},
  year={2021},
  publisher={Nature Publishing Group US New York}
}

@article{madissoon2023spatially,
  title={A spatially resolved atlas of the human lung characterizes a gland-associated immune niche},
  author={Madissoon, Elo and Oliver, Amanda J and Kleshchevnikov, Vitalii and Wilbrey-Clark, Anna and Polanski, Krzysztof and Richoz, Nathan and Ribeiro Orsi, Ana and Mamanova, Lira and Bolt, Liam and Elmentaite, Rasa and others},
  journal={Nature Genetics},
  volume={55},
  number={1},
  pages={66--77},
  year={2023},
  publisher={Nature Publishing Group US New York}
}

@article{bandaru2014targeting,
  title={Targeting filamin B induces tumor growth and metastasis via enhanced activity of matrix metalloproteinase-9 and secretion of VEGF-A},
  author={Bandaru, S and Zhou, AX and Rouhi, P and Zhang, Y and Bergo, MO and Cao, Yihai and Aky{\"u}rek, LM},
  journal={Oncogenesis},
  volume={3},
  number={9},
  pages={e119--e119},
  year={2014},
  publisher={Nature Publishing Group}
}

@article{ren2021tumor,
  title={Tumor protein D52 promotes breast cancer proliferation and migration via the long non-coding RNA NEAT1/microRNA-218-5p axis},
  author={Ren, Jing and Chen, Yunzi and Kong, Weishu and Li, Ye and Lu, Feng},
  journal={Annals of Translational Medicine},
  volume={9},
  number={12},
  year={2021},
  publisher={AME Publications}
}

@article{badve2007foxa1,
  title={FOXA1 expression in breast cancer—correlation with luminal subtype A and survival},
  author={Badve, Sunil and Turbin, Dmitry and Thorat, Mangesh A and Morimiya, Akira and Nielsen, Torsten O and Perou, Charles M and Dunn, Sandi and Huntsman, David G and Nakshatri, Harikrishna},
  journal={Clinical cancer research},
  volume={13},
  number={15},
  pages={4415--4421},
  year={2007},
  publisher={AACR}
}

@article{wolf2007foxa1,
  title={FOXA1: Growth inhibitor and a favorable prognostic factor in human breast cancer},
  author={Wolf, Ido and Bose, Shikha and Williamson, Elizabeth A and Miller, Carl W and Karlan, Beth Y and Koeffler, H Phillip},
  journal={International journal of cancer},
  volume={120},
  number={5},
  pages={1013--1022},
  year={2007},
  publisher={Wiley Online Library}
}

\end{document}